\documentclass[11pt,a4paper]{article}
\usepackage[a4paper, total={7in, 10in}]{geometry}
\usepackage{marvosym} 
\usepackage[superscript]{cite}
\usepackage[dvipsnames]{xcolor}              
\usepackage[colorlinks=true, citecolor=blue, linkcolor=black, urlcolor=blue]{hyperref}  
\usepackage[bottom,hang,flushmargin]{footmisc}  

\usepackage[left]{lineno}
\usepackage{setspace}

\newcommand\blfootnote[1]{%
  \begingroup
  \renewcommand\thefootnote{}\footnote{#1}%
  \addtocounter{footnote}{-1}%
  \endgroup
}

\usepackage{graphicx}
\usepackage{lineno,hyperref}
\usepackage{tabularx} 
\usepackage{booktabs}
\usepackage{amsmath,amssymb,amsfonts}
\usepackage{stackengine}
\usepackage{graphicx}
\usepackage{amssymb}
\usepackage{eqnarray} 
\usepackage{amsmath}
\usepackage{mathalfa} 
\usepackage[dvips]{epsfig}   
\usepackage[ruled,vlined,linesnumbered]{algorithm2e}
\usepackage{psfrag}
\usepackage{amsmath}
\usepackage{stfloats}
\usepackage{amssymb}
\usepackage{dsfont}
\usepackage{xcolor}
\usepackage{bbm}
\usepackage{float}
\usepackage[prependcaption,colorinlistoftodos]{todonotes}
\usepackage{hyperref}
\usepackage{MnSymbol}
\usepackage{mathrsfs}
\usepackage{lscape}
\usepackage{longtable}
\usepackage{rotating}
\usepackage{multirow}
\usepackage{amsthm}
\usepackage{color}
\usepackage{xcolor}
\usepackage{url}
\usepackage{subfigure}
\usepackage{rotating}
\usepackage{hyperref}
\usepackage{tikz}
\usetikzlibrary{shapes.geometric, arrows}
\hypersetup{
    colorlinks=true, 
    linktoc=all,     
    linkcolor=blue,  
}

\usepackage{fancyhdr}
\usepackage{chngcntr}

\usepackage{totcount}
\usepackage{atbegshi}

\usepackage[labelfont=bf]{caption} 
\usepackage{caption} 

\usepackage{pdflscape}
\usepackage{tabularx}

\usepackage{mwe}

\DeclareMathOperator{\support}{supp}
\DeclareMathOperator{\st}{s.t.}

\newcommand{\sB}{\mathcal{B}}

\newcommand{\sD}{\mathcal{D}}

\newcommand{\sX}{\mathcal{X}}
\newcommand{\sU}{\mathcal{U}}

\newcommand{\R}{\mathbb{R}}
\newcommand{\E}{\mathbb{E}}

\newcommand{\bv}[1]{\mathbf{#1}}
\newcommand{\statecost}[1]{c_{#1}^{(x)}\left(\bv{x}_{#1}\right)}
\newcommand{\actioncost}[1]{c_{#1}^{(u)}\left(\bv{u}_{#1}\right)}

\newcommand{\statecostexpectation}[1]{c_{#1}^{(x)}\left(\bv{X}_{#1}\right)}
\newcommand{\actioncostexpectation}[1]{c_{#1}^{(u)}\left(\bv{U}_{#1}\right)}

\newcommand{\costtogo}[2]{\hat{c}_{#1}\left(\bv{x}_{#2}\right)}

\newcommand{\costtot}[1]{\bar{c}_{#1}\left(\bv{x}_{#1}\right)}

\newcommand{\costtotexpectation}[1]{\bar{c}_{#1}\left(\bv{X}_{#1}\right)}

\newcommand{\costuncertainty}[2]{\tilde{c}\left(\bv{x}_{#2},\bv{u}_{#1}\right)}

\newcommand{\Vx}[2]{V_{\alpha}\left(\bv{x}_{k-1},\bv{u}_k\right)}
\newcommand{\Wx}[2]{W_{\alpha}\left(\bv{x}_{k-1},\bv{u}_k\right)}
\newcommand{\Vxtilde}[2]{\tilde{V}_{\alpha}\left(\bv{x}_{k-1},\bv{u}_k\right)}
\newcommand{\Mx}[2]{M\left(\bv{x}_{k-1},\bv{u}_k\right)}

\newcommand{\plant}[2]{p_{{#1}} \left(\bv{x}_{{#1}}\mid \bv{x}_{{#2}}, \bv{u}_{{#1}} \right)}

\newcommand{\shortplant}[2]{p_{{#1}\mid{#2}}^{(x)}}
\newcommand{\refplant}[2]{q_{{#1}} \left(\bv{x}_{{#1}}\mid \bv{x}_{{#2}}, \bv{u}_{{#1}} \right)}

\newcommand{\shortrefplant}[2]{q_{{#1}\mid{#2}}^{(x)}}
\newcommand{\nominalplant}[2]{\bar{p}_{{#1}} \left(\bv{x}_{{#1}}\mid \bv{x}_{{#2}}, \bv{u}_{{#1}} \right)}

\newcommand{\shortnominalplant}[2]{\bar{p}_{{#1}\mid{#2}}^{(x)}}

\newcommand{\policy}[2]{{\pi}_{{#1}}\left(\bv{u}_{{#1}}\mid \bv{x}_{{#2}} \right)}
\newcommand{\shortpolicy}[2]{{\pi}_{{#1}\mid{#2}}^{(u)}}
\newcommand{\refpolicy}[2]{q_{{#1}}\left(\bv{u}_{{#1}}\mid \bv{x}_{{#2}} \right)}
\newcommand{\refpolicyobs}[2]{q_{{#1}}\left(\bv{u}_{{#1}}=\hat{\bv{u}}_{{#1}}\mid \bv{x}_{{#2}} = \hat{\bv{x}}_{{#2}} \right)}
\newcommand{\refpolicygiven}[2]{q_{{#1}}\left(\bv{u}_{{#1}}\mid \bv{x}_{{#2}} = \hat{\bv{x}}_{{#2}} \right)}
\newcommand{\shortrefpolicy}[2]{q_{{#1}\mid{#2}}^{(u)}}

\newcommand{\optimalpolicy}[2]{{\pi}^{\star}_{{#1}} \left(\bv{u}_{{#1}}\mid \bv{x}_{{#2}} \right)}

\newcommand{\ball}[2]{\sB_{\eta}\left(\nominalplant{#1}{#2}\right)}

\newcommand{\KL}{\text{KL}}

\newcommand{\radius}[2]{\eta_{#1}\left(\bv{x}_{#2},\bv{u}_{#1}\right)}

\newcommand{\DKL}[2]{{D}_{\KL}\left(#1\mid \mid #2 \right)}
\newcommand{\AlgoDM}{DR-FREE}

\newcommand{\SI}{Supplementary Information}

\usepackage{dsfont}

\DeclareMathOperator{\argmin}{\operatorname{arg min}}

\fancypagestyle{SI}{
  \fancyhf{}

  \fancyfoot[C]{SI-\thepage}
}

\begin{document}
\onehalfspacing
\title{Distributionally Robust Free Energy Principle for Decision-Making}

\author{Allahkaram Shafiei~\textsuperscript{1,$\ast$} \and Hozefa Jesawada~\textsuperscript{{2,$\ast$}} \and Karl Friston~\textsuperscript{{3}} \and Giovanni Russo~\textsuperscript{4} \Letter}

\maketitle~\blfootnote{\textsuperscript{1} Czech Technical University, Prague, Czech Republic. {\textsuperscript{2} New York University Abu Dhabi,  Abu Dhabi Emirate. } \textsuperscript{{3}} Wellcome Centre for Human Neuroimaging, Institute of Neurology, University College London,  United Kingdom.  \textsuperscript{4} Department of Information and Electrical Engineering and~Applied Mathematics, University of Salerno,  Italy. 

\textsuperscript{$\ast$} These authors contributed equally.   \Letter~e-mail: \href{mailto:giovarusso@unisa.it}{giovarusso@unisa.it}}

\begin{abstract}
\textbf{Abstract.} Despite their groundbreaking performance, autonomous agents can misbehave when training and environmental conditions become inconsistent, with minor mismatches leading to undesirable behaviors or even catastrophic failures. Robustness towards these training-environment ambiguities is a core requirement for intelligent agents and its fulfillment is a long-standing challenge towards their real-world deployments.  Here, we introduce a Distributionally Robust Free Energy model (\AlgoDM) that instills this core property by design. Combining a robust extension of the free energy principle with a resolution engine, \AlgoDM~wires robustness into the agent decision-making mechanisms. Across benchmark experiments,  \AlgoDM~enables the agents to complete the task even when, in contrast, state-of-the-art models fail. This milestone may inspire both deployments in multi-agent settings and, at a perhaps deeper level, the quest for an explanation of how natural agents -- with little or no training  -- survive in capricious environments.
\end{abstract}

\section*{Introduction}\label{sec:introduction}

\begin{figure*}[!h]
	\centering
\noindent\includegraphics[width=0.85\columnwidth]{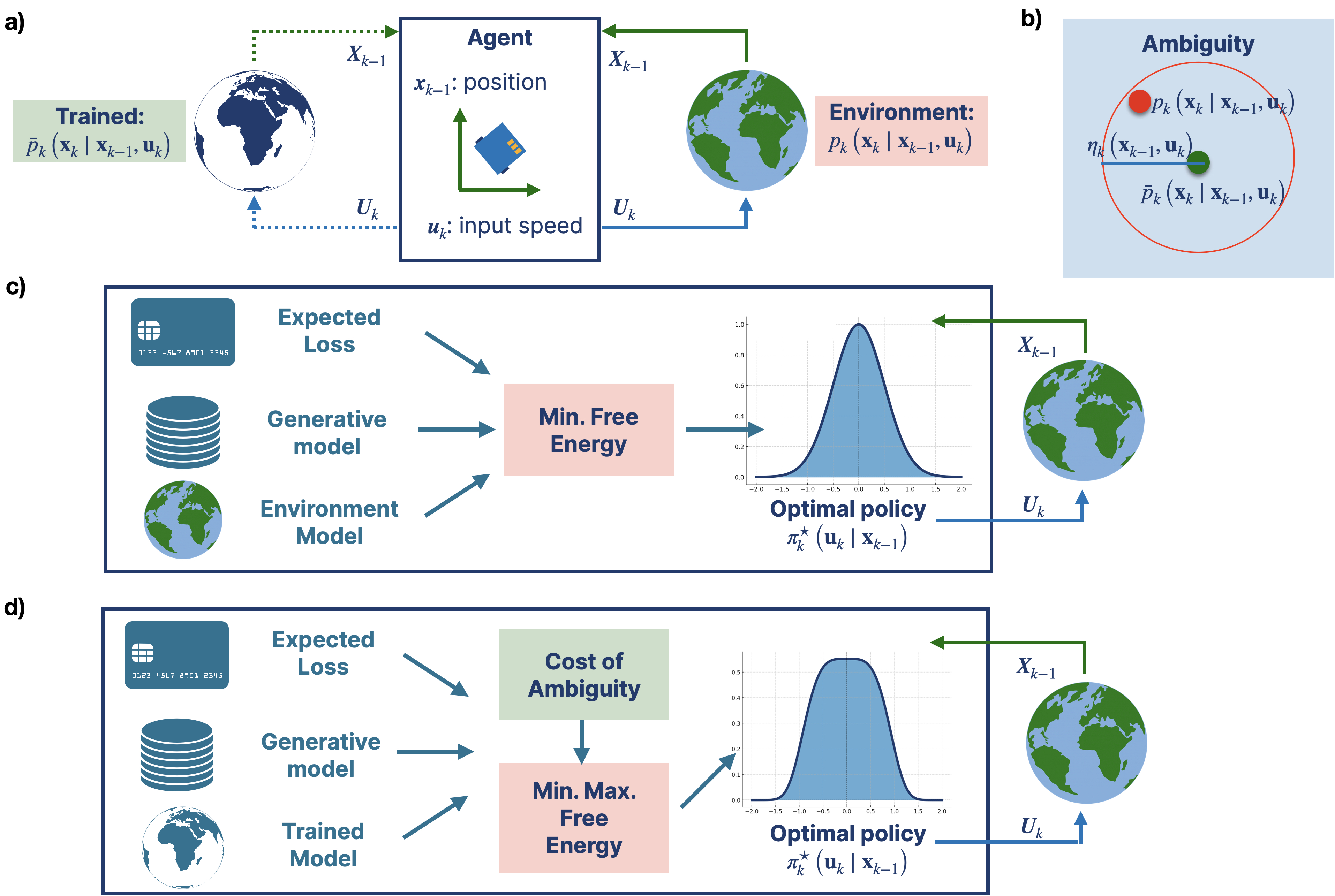}  
\caption{{\bf Comparison between free energy and robust free energy  for policy computation. a.} A robotic agent navigating a stochastic environment to reach a  destination while avoiding obstacles. At a given time-step, $k-1$, the agent determines an action $\bv{U}_k$ from a policy using a model of the environment (e.g., available at training via a simulator possibly updated via real world data) and observations/beliefs (grouped in the state $\bv{X}_{k-1}$). The environment and model can change over time. Capital letters are random variables, lower-case letters are realizations. {\bf b.} The trained model and the agent environment differ. This mismatch is a training/environment {(model)} ambiguity: for a state/action pair, the ambiguity set is the set of all possible environments that have statistical complexity from the trained model of at most $\radius{k}{k-1}$. We use the  wording trained model in a very broad sense. A trained model is any model available to the agent offline: for example, this could be a model obtained  from a simulator or, for natural agents, this could be hardwired into evolutionary processes or even determined by prior beliefs. {\bf c.} A free energy minimizing agent in an environment matching its own model. The agent determines an action by sampling from the policy $\optimalpolicy{k}{k-1}$. Given the model, the policy is obtained by minimizing the variational free energy: the sum of a statistical complexity (with respect to a generative model, $q_{0:N}$) and expected loss (state/action costs, $\statecost{k}$ and $\actioncost{k}$) terms. {\bf d.} \AlgoDM~extends the free energy principle to account for model ambiguities. According to \AlgoDM, the maximum free energy across all environments -- in an ambiguity set -- is  minimized to identify a robust policy. This amounts to variational policy optimization under the epistemic uncertainty engendered by  ambiguous environment.}\label{fig:intro}
\end{figure*}

A popular approach to designing autonomous agents is to feed them with data, using Reinforcement Learning (RL) and simulators to train a policy (Fig. \ref{fig:intro}a). Deep RL agents designed on this paradigm have demonstrated remarkable abilities, including outracing human champions in Gran Turismo\cite{PW_etal_22}, playing  Atari games\cite{VM_et_al:15}, controlling plasmas\cite{JD_etal_22} and achieving champion-level performance in drone races\cite{EK_etal_23}. However, despite their groundbreaking performance, state-of-the-art agents cannot yet compete with natural intelligence in terms of policy robustness: natural agents have, perhaps through evolution, {acquired} decision-making abilities so that they can function in challenging environments despite little or no training\cite{BL_JT_SG:17,VV_etal:22,HM_etal:24}. In contrast, for artificial agents, even when {they have access to a} high fidelity simulator, the learned policies can be  brittle to mismatches, or ambiguities, between {the model available during learning} and {the real} environment (Fig. \ref{fig:intro}b).  For example, drone-champions and Atari-playing agents assume consistent environmental conditions from training, and if this assumption fails, because, e.g.,  the environment illumination or color of the objects changes,  or the drone has a malfunctioning -- so that its dynamics becomes different from the one available during training --  learned policies can fail. More generally,  {model} ambiguities {-- even if minor --} can lead to non-robust behaviors and failures in open-world environments\cite{MK_etal_24}. Achieving robustness towards {these} training/environment ambiguities is a long-standing challenge{\cite{RM_PE:25,JM_KH_HA_SS_DC_JP:22,BT_DI_CK_DK:23}} for the design of intelligent machines\cite{BL_JT_SG:17,LR_etal_23,MW_etal_23} that can operate in the real world.

Here we present \AlgoDM, a free energy\cite{GH_RZ:93,PD_GH:95} computational model that addresses this challenge: \AlgoDM~instills this core property of intelligence directly into the agent decision-making mechanisms. This is achieved by grounding \AlgoDM~in the minimization of the free energy, a unifying account across information theory, machine learning\cite{SJ_OS:21,MHA_EI_RW_RM_JC:21,TP_GP_KF:22,PJ_HH_VR:24,TS:14,TB_AP_TM:24,MR_etal:21}, neuroscience,  computational and cognitive sciences\cite{KF:09,CH_etal:24,TP_SW:23,JH:13,KF_LDC_NS_CH_KU_GP_AG_TP:23,SG_DB:20,AI_JW_JP:20}. The principle postulates that adaptive behaviors in natural and artificial agents arise from the minimization of variational free energy (Fig. \ref{fig:intro}c). \AlgoDM~consists of two components. The first component is an  extension of the free energy principle: the distributionally robust (DR) free energy (FREE) principle, which fundamentally reformulates how free energy minimizing agents handle ambiguity. While classic free energy models (Fig. \ref{fig:intro}c) obtain a policy by minimizing the free energy based on a model of the environment available to the agent,  under our robust principle, free energy is instead minimized across all possible environments within an ambiguity set around a trained model. The set is defined in terms of statistical complexity around the trained model. This means the actions of the agent are sampled from a policy that minimizes the maximum free energy across ambiguities. The robust principle yields {the problem statement for policy computation. This is a distributionally robust problem having a free energy functional as objective and ambiguity constraints formalized in terms of statistical complexity.  The problem has} not only a nonlinear cost functional with nonlinear constraints but also probability densities over decision variables that equip agents with explicit estimates of uncertainty and confidence. The product of this  framework is a policy that {minimizes free energy and is} robust across model ambiguities. {T}he second key component of \AlgoDM~-- its resolution engine -- is {the method} to compute this  policy.  In contrast to conventional approaches for policy computation based on free energy models, our {method} shows that the  policy can be conveniently found  by first maximizing the free energy across model ambiguities -- furnishing a cost under ambiguity -- and then minimizing the free energy in the policy space (Fig. \ref{fig:intro}d).  Put simply, policies are selected under the best worst-case scenario, where the worst cases accommodate ambiguity.  {Our robust free energy principle yields -- when there is no ambiguity  -- a problem statement for policy computation naturally arising across learning\cite{TB:24,TB_AP_TM:24,BE_SL:22} -- in the context of maximum diffusion (MaxDiff) and maximum entropy (MaxEnt) -- and control\cite{EG_HJ_CDV_GR:24}.  This means that \AlgoDM~can yield policies that not only inherit all the desirable properties of these approaches but ensures them across an ambiguity set, which is explicitly defined in the formulation.  In MaxEnt -- and MaxDiff --  robustness depends on the entropy of the optimal policy, with explicit bounds on the ambiguity set over which the policy is robust available  in discrete settings\cite{BE_SL:22}.  To compute a policy that robustly maximizes a reward, MaxEnt needs to be used with a different, pessimistic, reward\cite{BE_SL:22}  – this is not required in \AlgoDM. These desirable features of our free energy computational model are enabled by its resolution}  engine. {This is -- to the best of our knowledge -- the only available method  tackling the full distributionally robust, nonlinear and infinite-dimensional policy computation problem arising from our robust principle{;} see Results and Sec.  S2 in \SI~for details{. In \SI~we also highlight a connection with the  Markov Decision Processes (MDPs) formalism}.  \AlgoDM~}yields a policy with a well-defined structure{: this is a soft-max with its exponent depending on ambiguity. This structure} elucidates the crucial role of  ambiguity {on optimal decisions, i.e., how it modulates  the probability of selecting a given action}.  
 
\AlgoDM~not only returns the policy arising from our free energy model, but also establishes its performance limits.  In doing so, \AlgoDM~harbors two  implications. First,  \AlgoDM~policy is interpretable and supports (Bayesian) belief updating. The second implication is that it is impossible for an agent faced with  ambiguity to outperform an ambiguity-free agent. As ambiguity vanishes, \AlgoDM~recovers the policy of an agent that has perfect knowledge of its environment, and no agent can obtain better performance. Vice-versa,  as ambiguity increases,  \AlgoDM~shows that {the policy down-weights the model available to the agent over ambiguity}. 

We evaluate \AlgoDM~on an experimental testbed involving real rovers, which are given the task of reaching a desired destination while avoiding obstacles. The trained model available to \AlgoDM~is learned from biased experimental data, which does not adequately capture the real environment and  introduces ambiguity.  In th{e experiments --} even despite the ambiguity arising from having learned a model from biased data {--} \AlgoDM~successfully enables the rovers to complete their task{, even in settings where both a state-of-the-art} free energy minimizing agent {and other methods} struggle{,} unable to complete the task. The experiments  {results -- confirmed by evaluating \AlgoDM~in a popular higher dimensional  simulated environment -- } suggest that, to operate in open environments,  agents require built-in {robustness} mechanisms and these are crucial to compensate for poor training. \AlgoDM, providing a mechanism {that defines robustness in the problem formulation},  delivers this capability. 

Our free energy computational model, \AlgoDM, reveals how free energy minimizing agents can compute optimal actions that {are} robust {over an} ambiguity {set defined in the problem formulation}.  It establishes a normative framework to both empower the design of artificial agents built upon free energy models with robust decision-making abilities, and to understand natural behaviors beyond current free energy explanations\cite{KF_FR_DO_CM_TF_GP:15,TP_KF:17,YK_HN:23,KK_SP_SM_RP_MS_JCD_RR:19,AM_PL_GP:22}. {D}espite its success, there is no theory currently explaining if and how these free energy agents can compute actions in ambiguous settings.  \AlgoDM~provides these explanations. 

\begin{figure*}[h!]
	\centering
\noindent\includegraphics[width=0.83\columnwidth]{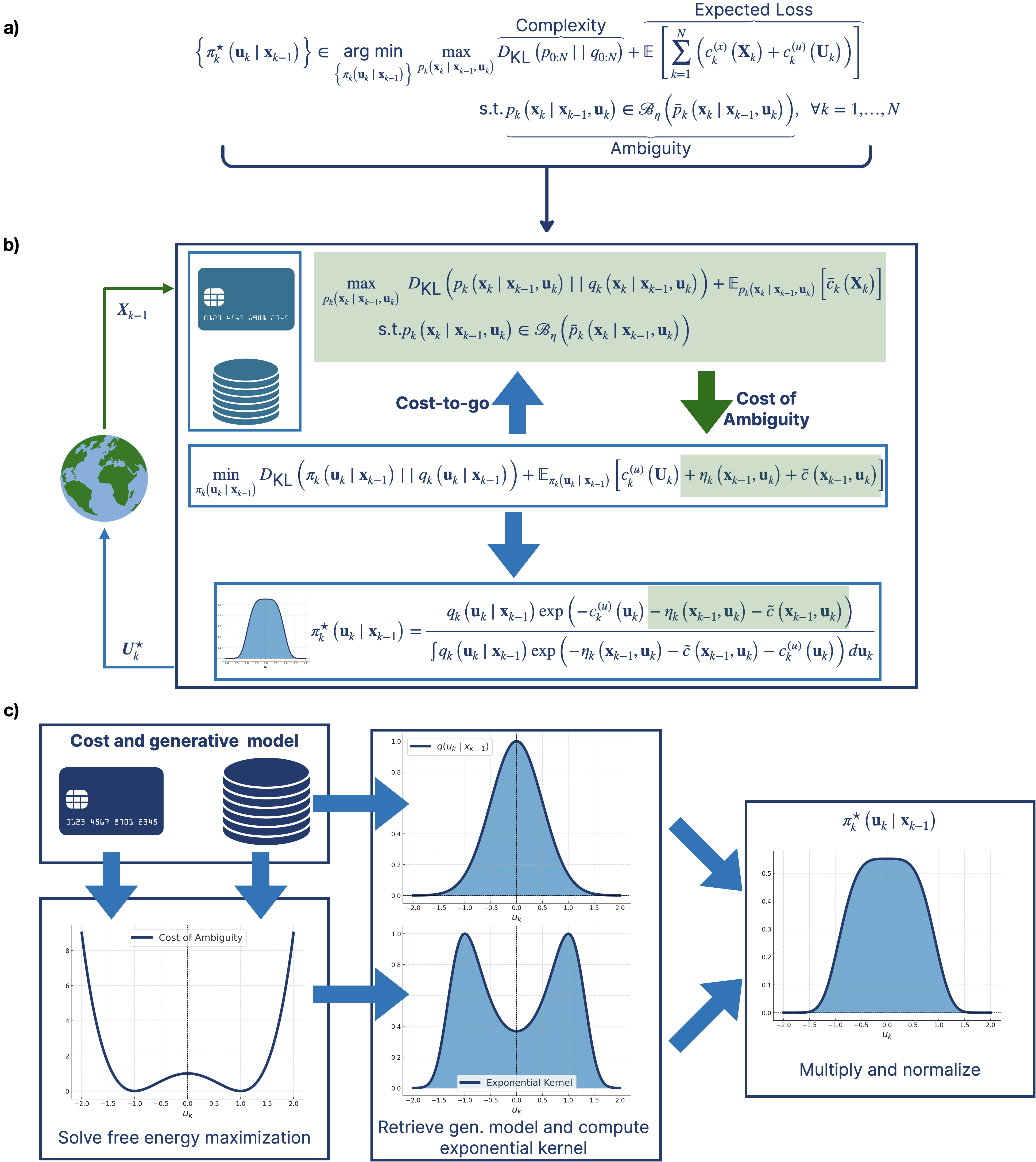}
 \caption{{\bf \AlgoDM~a.} Summarizing the distributionally robust free energy principle {-- the problem statement for policy computation}.  Our generalization of active inference yields an optimization framework where policies emerge by minimizing the maximum free energy over all possible environments in the ambiguity set{, which formalizes the constraints in the problem formulation.}  {\bf b.} The resolution engine to find the policy. Given the current state, the engine uses the generative model and the loss to find the maximum free energy $\DKL{\plant{k}{k-1}}{\refplant{k}{k-1}} + \E_{\plant{k}{k-1}}\left[\costtotexpectation{k}\right]$ across all the environments in the ambiguity set. This yields the cost of ambiguity $\radius{k}{k-1}+\costuncertainty{k}{k-1}$ that builds up the expected loss for the subsequent minimization problem. In this second problem, the variational free energy is minimized in the space of polices providing: (i) 
 $\optimalpolicy{k}{k-1}$, the \AlgoDM~policy from which actions are sampled.  Elements that guarantee robustness
in green -- {these terms depend on ambiguity}; (ii) the smallest free energy that the agent can achieve, i.e., the cost-to-go $\costtot{k}$ fed back to the maximization problem at the next time-step. For {reactive} actions, where $N=1$, the cost-to-go equals the state cost given by the agent loss. {\bf c.} Using the generative model and the state-cost, \AlgoDM~first computes the cost of ambiguity, which is non-negative. This, together with the action cost is then used to obtain the exponential kernel in the policy, i.e. $\exp\left(-\actioncost{k}-\radius{k}{k-1}-\costuncertainty{k}{k-1}\right)$. After multiplication of the kernel with $\refpolicy{k}{k-1}$ and normalization, this returns $\optimalpolicy{k}{k-1}$.}\label{fig:DRFREE}
\end{figure*}

\section*{Results}

\noindent{\bf \AlgoDM.}  \AlgoDM~comprises a distributionally robust free energy principle and the accompanying resolution engine { -- the principle (Fig. \ref{fig:DRFREE}a) is the problem statement for policy computation; the resolution engine is the  method for policy computation.} The principle establishes a sequential policy optimization framework, where randomized policies arise from the minimization of the maximum free energy over ambiguity. The resolution engine finds the solution in the space of policies.  This is done by computing -- via the maximum free energy over all possible environments in the ambiguity set -- a cost associated to ambiguity. Then, the ensuing maximum free energy  is minimized in policy space (Fig. \ref{fig:intro}d). 

In Fig. \ref{fig:intro}a,  {random variables $\bv{X}_{k-1}$ and $\bv{U}_k$ are state at time $k-1$ and action at $k$ (see Methods and Sec.  S3 of  \SI).  T}he agent infers the state of a stochastic environment $\plant{k}{k-1}$ and the action is sampled from  $\policy{k}{k-1}$. In the decision horizon (e.g., from $1$ to $N$) the agent-environment interactions are captured by $p_{0:N}$, defined as $p_0(\bv{x}_0)\prod_{k=1}^N\plant{k}{k-1}\policy{k}{k-1}$, where $p_0(\bv{x}_0)$ is an initial prior. The trained model available to the agent, $\nominalplant{k}{k-1}$, does not (necessarily) match its environment and the goal of \AlgoDM~is to compute a policy that, while optimal, is robust against the training/environment ambiguities. For example, if -- as in our experiments --  the robot model of  Fig. \ref{fig:intro}a is learned from corrupted/biased data, then it differs from the real robot and this gives rise to ambiguities. These ambiguities are captured via the ambiguity set of Fig. \ref{fig:intro}b. For a given state/action pair, this is the set of all possible environments with statistical complexity within $\radius{k}{k-1}$ from the trained model. For this reason, the ambiguity set $\ball{k}{k-1}$ is captured via the Kullback-Leibler (KL) divergence: the ambiguity set is the set of all {possible models, say} $\plant{k}{k-1}${,} such that $\DKL{\plant{k}{k-1}}{\nominalplant{k}{k-1}} \le \radius{k}{k-1}$. The radius of ambiguity {available to \AlgoDM}, $\radius{k}{k-1}$, is positive and bounded. For a state/action pair, a small radius indicates low ambiguity, meaning that the agent is confident in its trained model. Vice-versa, high values indicate larger ambiguity and hence low confidence in the model. See Methods for the detailed definitions.

\AlgoDM~computes the optimal policy via a distributionally robust generalization of the free energy
principle that accounts for ambiguities.  {This first component of \AlgoDM, providing the problem statement for policy computation,} generalizes the conventional free energy principle to ensure policies remain robust even when the environment deviates from the trained model. Our principle is formulated as follows: over the decision horizon, the optimal policy sequence $\{\optimalpolicy{k}{k-1}\}_{1:N}$ is obtained by minimizing over the policies the maximum free energy across all possible environments in the ambiguity set expected under the policy in question. The expected free energy combines two terms: (i) the statistical complexity, $\DKL{p_{0:N}}{q_{0:N}}$, of the agent-environment behavior from $q_{0:N}$; (ii) the expected cumulative loss  $\E_{p_{0:N}}\left[\sum_{k=1}^N {\left(\right.}\statecostexpectation{k} + \actioncostexpectation{k}{\left.\right)} \right]$.  {The principle is formalized in Fig. \ref{fig:DRFREE}a (see Methods for details) as a distributionally robust policy optimization problem in the probability space.   \AlgoDM~computes the policy via bi-level optimization,  without requiring the environment $\plant{k}{k-1}$ and without performing stochastic sampling for the ambiguity set. The ambiguity set defines the problem constraints -- these are nonlinear in the decision variable -- and the min-max objective is the free energy -- also nonlinear in the decision variables (see Sec.  S2 in the \SI~for connections with other frameworks).  As also shown in Fig. \ref{fig:DRFREE}a, t}he complexity term {in the min-max objective} regularizes the {optimal} policy to prevent environment-agent interactions that are overly complex with respect to $q_{0:N}$. This is specified as $ q_0(\bv{x}_0)\prod_{k=1}^N\refplant{k}{k-1}\refpolicy{k}{k-1}$, with $q_0(\bv{x}_0)$ being a prior.  {Hence,  the first term in the min-max objective biases the optimal solution of our robust principle -- \AlgoDM~policy --  towards  $q_{0:N}$.  The second term in the objective minimizes the worst case expected loss across ambiguity.  We refer to  $q_{0:N}$ as generative model,  although -- for the application of our principle in Fig \ref{fig:DRFREE}a -- this is not necessarily required  to be a time-series model.  For example,  in some of our navigation experiments,  $q_{0:N}$ only encodes the agent goal destination -- in this case the complexity term in Fig. \ref{fig:DRFREE}a encodes an error from the goal.  In contrast,  in a second set of experiments -- where we relate \AlgoDM~to a state-of-the-art approach from the literature\cite{TB:24,TB_AP_TM:24} --  $q_{0:N}$ encodes a time-series model.   Our formulation also allows for the generative model $\refplant{k}{k-1}$ --  provided to \AlgoDM~--  to be different from the true environment  $\plant{k}{k-1}$ -- not available to the agent -- and the trained model $\nominalplant{k}{k-1}$,  see also Fig. \ref{fig:intro}a-b.  This feature can be useful in applications where, by construction, the model available to the agent and the generative model differ\cite{CH_etal:24}. } Both {reactive} and planned behaviors can be seen through the lenses of our formulation: a width of the decision horizon $N$ greater than $1$ means that the policy at time-step $k$ is computed as part of a plan. If $N=1$, the agent action is {reactive (or greedy)}. {These} actions frequently emerge as reflexive responses in biologically inspired motor control\cite{JV_etal:06,GP_FR_KF:15}.  In brief, this generalization of active inference can be regarded as robust Bayesian model averaging to accommodate epistemic uncertainty about the environment, where the prior over models is supplied by the KL divergence between each model and the trained model.  When there is no ambiguity,  our robust principle in Fig. \ref{fig:DRFREE}a  {connects with}  free energy minimization in active inference based upon expected free energy\cite{TP_KF:19}  {--}  which itself generalizes schemes such as KL control and control as inference\cite{JB_WW_HK:10,AH:03,MB_MT:12,AM_PL_GP:22}.  See Methods and Sec.  S2 of \SI.

The {policy} optimization {problem in Fig. \ref{fig:DRFREE}a} is infinite-dimensional as both  minimization and maximization are in the space of probability densities.  This enables a Bayes-optimal handling of uncertainty and ambiguity that characterizes {control and} planning as (active) inference.  \AlgoDM~{resolution engine -- the method to compute the policy --} not only finds the policy but also, perhaps counterintuitively, returns a solution with a well-defined and explicit functional form. {The analytical results behind the resolution engine are in Sec.  S3 and Sec.  S6 of the \SI. In summary,  the analytical results show that, at each $k$,  the optimal policy can be found via a bi-level optimization approach,  first maximizing free energy across the ambiguity constraint and then minimizing over the policies.  While the maximization problem is still infinite dimensional,  its optimal value -- yielding a cost of ambiguity -- can be obtained by solving a scalar optimization problem.  This scalar optimization problem is convex and has a global minimum.  Therefore,  once the cost of ambiguity is obtained, the resulting free energy can be minimized in the policy space and the optimal policy is unique. These theoretical findings are summarized in} Fig. \ref{fig:DRFREE}b. Specifically, the policy at time-step $k$ is a soft-max (Fig. \ref{fig:DRFREE}b) obtained by equipping the generative model $\refpolicy{k}{k-1}$ with an exponential kernel. The kernel contains two costs: the action cost $\actioncost{k}$ and the cost of ambiguity,  $\radius{k}{k-1} + \costuncertainty{k}{k-1}$. Intuitively, given the cost-to-go $\costtot{k}$, the latter cost is the maximum free energy across all possible environments in the ambiguity set. The infinite dimensional free energy maximization step, which can be reduced to scalar convex optimization, yields a cost of ambiguity that is always bounded and non-negative (Fig. \ref{fig:DRFREE}c and {Methods}). This implies that an agent always incurs a positive cost for ambiguity and the higher this cost is for a given state/action pair, the lower the probability of sampling that action is. The result is that \AlgoDM~policy balances between the cost of ambiguity and the agent beliefs encoded in the generative model (Fig. \ref{fig:DRFREE}c).\\

\begin{figure*}[!h]
	\centering
\noindent\includegraphics[width=0.79\columnwidth]{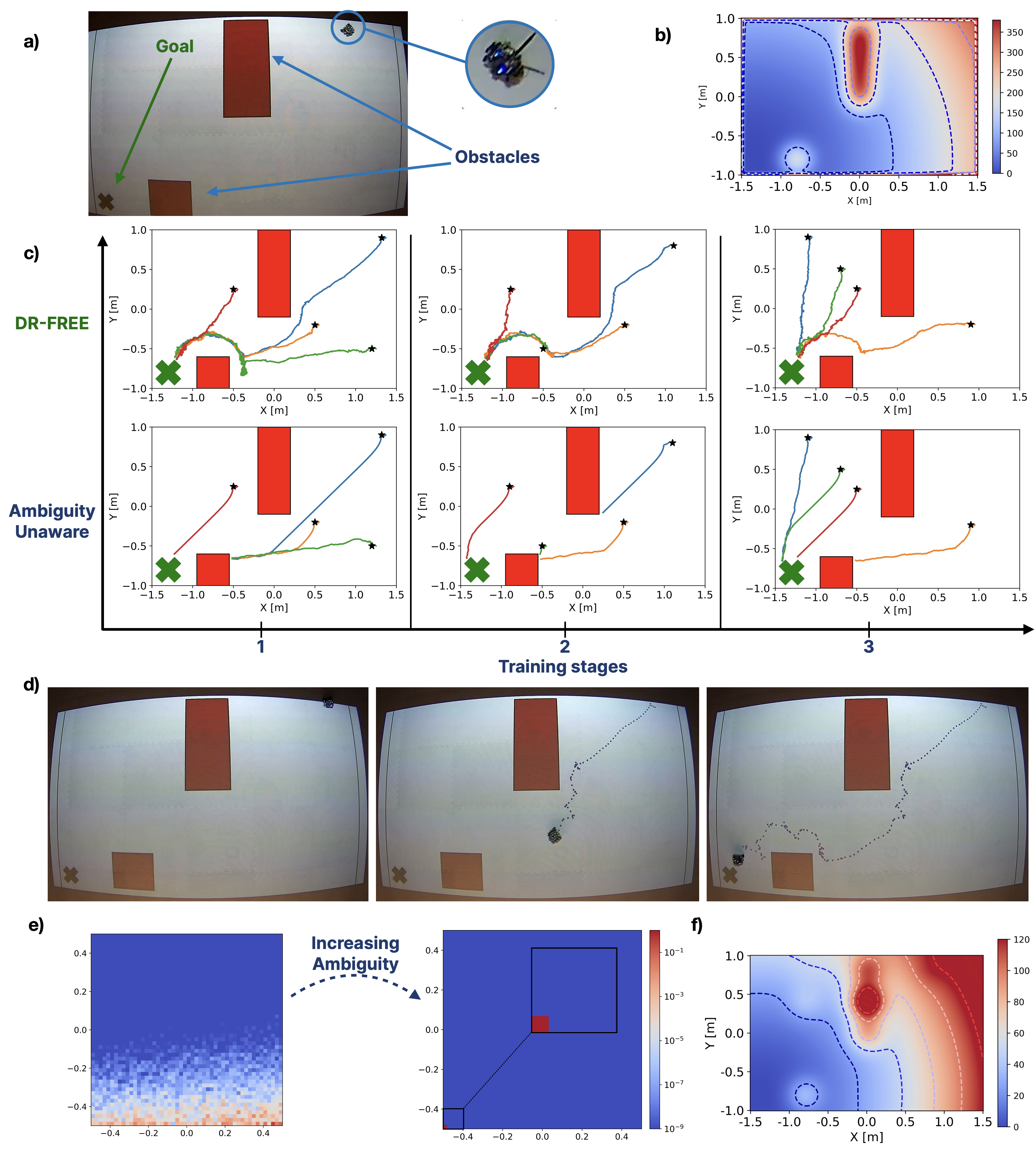}  
\caption{{\bf \AlgoDM~evaluation. a.} Unicycle robots  of \(11 \text{cm} \times 8.5\text{cm} \times 7.5\text{cm}\) (width, length, height) that need to achieve the goal destination, $\bv{x}_d$, avoiding  obstacles. The work area is \(3\text{m} \times 2\text{m}\), the robot position is the state, and actions are vertical/horizontal speeds; $\refplant{k}{k-1}$ is a Gaussian centered in $\bv{x}_d$ and $\refpolicy{k}{k-1}$ is uniform. See {Methods} for the settings. {\bf b.} The non-convex state cost for the navigation task. See {Methods} for the expression. {\bf c.}  Comparison  between \AlgoDM~and a free-energy minimizing agent that makes optimal decisions but is unaware of the ambiguity. \AlgoDM~enables the robot to successfully complete the task at each training stage.  The ambiguity-unaware agent fails, except when the shortest path is obstacle-free.  Training details are in {Methods}. {\bf d.} Screenshots from the Robotarium platform recording of one experiment.   \AlgoDM~allows the robot (starting top-right) to  complete the task (trained model from stage $3$ used).  {\bf e.} How \AlgoDM~policy changes as a function of ambiguity.  By increasing  the radius of ambiguity by $50\%$,  \AlgoDM~policy (left) becomes a policy dominated by ambiguity (right).  As a result, actions with low ambiguity are assigned higher probability.  Screenshot of the robot policy when this is in position $[0.2, 0.9]$,  i.e., near the middle obstacle.  The ambiguity increase deterministically drives the robot bottom-left (note the higher probability) regardless of the presence of the obstacle.  {\bf f.} Belief update.  Speeds/positions from the top-right experiments in panel c) are used together with $F=16$ state/action features,  $\varphi_i(\bv{x}_{k-1},\bv{u}_k) = \E_{\nominalplant{k}{k-1}}\left[\phi_i(\bv{X}_{k})\right]$ in Supplementary Fig.  1b. Once the optimal weights, $w_i^{\star}$, are obtained, the reconstructed cost is $-E_{\nominalplant{k}{k-1}}\left[\sum_{i=1}^{16}w_i^\star\phi_i(\bv{X}_{k})\right]$. Since this lives in a $4$-dimensional space,   we show $-\sum_{i=1}^{16}w_i^\star\phi_i(\bv{x}_{k})$, which can be conveniently plotted.} 
 \label{fig:results}
\end{figure*}

\noindent{\bf \AlgoDM~succeeds when ambiguity-unaware free energy minimizing agents fail.}  To evaluate \AlgoDM~we specifically considered an experiment where simplicity was a deliberate feature, ensuring that the effects of model ambiguity on decision-making could be identified, benchmarked against the literature\cite{EG_HJ_CDV_GR:24}, and measured quantitatively. The experimentation platform (Fig. \ref{fig:results}a) is the Robotarium\cite{SW_etal:20}, providing both hardware and a high-fidelity simulator.  The task is robot navigation: a rover needs to reach a goal destination while avoiding obstacles (Fig. \ref{fig:results}b). In this set-up we demonstrate that an ambiguity-unaware free energy minimizing agent   -- even if it makes optimal actions -- does not reliably complete the task, while \AlgoDM~succeeds.  {The ambiguity-unaware agent from the literature\cite{EG_HJ_CDV_GR:24} computes the optimal policy by solving a relaxation of the problem in Fig. \ref{fig:DRFREE}a without ambiguity.   This agent solves a policy computation problem -- relevant across learning and control\cite{GARRABE202281} -- having \AlgoDM~objective but without constraints. } We performed several experiments: in each experiment, \AlgoDM, used to compute {reactive} actions, only had access to a trained model $\nominalplant{k}{k-1}$ and did not know $\plant{k}{k-1}$. We trained {off-line a Gaussian Process} model{,} learned in stages.  {At each stage, data were obtained by applying randomly sampled actions to the robot}  and a bias was purposely added to the {robot positions} (see Experiments settings in {Methods} {for training details and Sec.  S5 in \SI~for the data}) thus introducing ambiguity. The corrupted data from each stage were then used to learn a trained model via Gaussian Processes. Fig. \ref{fig:results}c shows the performance of \AlgoDM~at each stage of the training, compared to the performance of a free energy minimizing agent that makes optimal decisions but is ambiguity unaware. In the first set of experiments, when equipped with \AlgoDM, the robot is always able to successfully complete the task (top panels in Fig. \ref{fig:results}c): in all the experiments, the robot was able to reach the goal while avoiding the obstacles. In contrast, in the second set of experiments,  when the robot {computes reactive actions by minimizing the} free energy -- without using \AlgoDM~-- it fails the task, crashing in the obstacles, except in trivial cases where the shortest path is obstacle-free ({see} Fig. \ref{fig:results}c, bottom; {d}etails in Methods). {The conclusion is confirmed when this ambiguity-unaware agent is equipped with planning capabilities.  As shown in Supplementary Fig.  3 for different widths of the planning horizon, the ambiguity-unaware agent still only fulfills the task when the shortest path is obstacle-free,  confirming the findings shown in Fig. \ref{fig:results}c, bottom.} The experiments provide two key highlights. First, ambiguity alone can have a catastrophic impact on the agent and its surroundings. Second, \AlgoDM~enables agents to succeed in their task despite the very same ambiguity.  This conclusion is also supported by experiments {where \AlgoDM~is deployed on the Robotarium hardware}. As shown in Fig. \ref{fig:results}d, \AlgoDM~in fact enabled the robot provided by the Robotarium to navigate to the destination, effectively completing the task despite model ambiguity.  {The computation time measured on the Robotarium hardware experiments was of approximately $0.22$ seconds (Methods for details).} See Data Availability for a recording; code also provided (see Code Availability).   {Supplementary Fig.  4  presents results from a complementary set of experiments in the same domain but featuring different goal positions and obstacle configurations.  The experiments confirm that, despite ambiguity,  \AlgoDM~consistently enables the robot to complete the task across all tested environments (code also available).} \\

\noindent{\bf \AlgoDM~elucidates the mechanistic role of ambiguity on optimal decision making.} \AlgoDM~policy (Fig. \ref{fig:DRFREE}b) assigns lower probabilities to states and actions associated with higher ambiguity. In simpler terms, an agent that follows \AlgoDM~policy is more likely to select actions and states associated with lower ambiguity.   \AlgoDM~yields a characterization of the agent behavior in regimes of small and large ambiguity.   Intuitively, as ambiguity increases,  \AlgoDM~yields a policy dominated by the agent's generative model and {radius of ambiguity}. In essence, as ambiguity increases, \AlgoDM~implies that the agent grounds  decisions on priors and ambiguity, reflecting its lack of confidence. Conversely, when the agent is confident about its trained model, \AlgoDM~returns the policy of a free energy minimizing agent making optimal decisions in a well-understood, ambiguity-free, environment. 

Characterizing optimal decisions in the regime of large ambiguity amounts at studying  \AlgoDM~policy (Fig. \ref{fig:DRFREE}b) as $\radius{k}{k-1}$ increases.  More precisely,  this means studying what happens when $\eta_{\min} = \min\radius{k}{k-1}$ increases. Since $\costuncertainty{k}{k-1}$ and $\actioncost{k}$ are non-negative,  and since $\costuncertainty{k}{k-1}$ does not depend on $\radius{k}{k-1}$ for sufficiently large $\eta_{\min}$,  this means that,  when $\radius{k}{k-1}$ is large enough,  $\exp\left(-\radius{k}{k-1} - \costuncertainty{k}{k-1} - \actioncost{k}\right) \approx \exp\left(-\radius{k}{k-1}\right)$ in \AlgoDM~policy (derivations in \SI). Therefore, as ambiguity increases, $\optimalpolicy{k}{k-1}$ only depends on the generative model and the ambiguity radius.   Essentially, \AlgoDM~shows that, when an agent is very unsure about its environment,  {optimal decisions are dominated by} the generative model and {the radius of} ambiguity.  {In our experiments, an implication of this is that with larger $\radius{k}{k-1}$ the robot may sacrifice physical risk (obstacle avoidance) in favor of epistemic risk (avoiding model mismatch). Hence, an interesting but not usual outcome is that {risk-averse is not equal to obstacle-avoidance especially when model uncertainty is not evenly distributed}. } This behavior is clearly evidenced in our experiments.   As shown in Fig. \ref{fig:results}e, as ambiguity increases the agent's policy,  becoming dominated by ambiguity,  deterministically directs the robot towards the goal position -- associated to the lowest ambiguity -- disregarding the presence of obstacles.  At the time-step captured in the figure, the robot was to the right of the middle obstacle. Consequently, while  \AlgoDM~policy with the original ambiguity radius would assign higher probabilities to speeds that drive the robot towards the bottom of the work-area,  when ambiguity increases the robot is instead directed bottom-left and this engenders a behavior that makes the robot crash in the obstacle.  

Conversely, to characterize the \AlgoDM~policy in the regimes of low ambiguity, we need to study how the policy changes as $\radius{k}{k-1}$ shrinks. As $\radius{k}{k-1}\to 0$, the ambiguity constraint is relaxed  {and $\costuncertainty{k}{k-1}$ simply becomes  $\DKL{\nominalplant{k}{k-1}}{\refplant{k}{k-1}} + \E_{\nominalplant{k}{k-1}}\left[\costtotexpectation{k}\right]$.  See  \SI~for the  derivations.} This yields the optimal policy   {from} the literature\cite{EG_HJ_CDV_GR:24}.  The minimum free energy attained by this policy  {when there is no ambiguity} is always smaller than the free energy achieved by an agent affected by ambiguity.  Hence, \AlgoDM~shows that ambiguity cannot be exploited by a free energy minimizing agent to obtain a better cost.  See {Methods} for further discussion.   {Supplementary Fig.  5 show experiments results for different ambiguity radii.  Experiments confirm that,  due to the presence of ambiguity, when the radius is set to zero the agent -- now ambiguity unaware -- does not always complete the task.   Additionally,   we conduct a second set of experiments in which the trained model $\nominalplant{k}{k-1}$ coincides with $\plant{k}{k-1}$.  This scenario  remains stochastic but no longer features ambiguity.  The results (Supplementary Fig.  6) show that -- unlike in the previous experiments -- \AlgoDM~enables the robot to complete the task even when the ambiguity radius is set to $0$.  This outcome is consistent with our analysis: when there is no ambiguity, the KL divergence between $\plant{k}{k-1}$ and $\nominalplant{k}{k-1}$  is identically zero and \AlgoDM~therefore succeeds even for a zero ambiguity radius.}  We provide the code to replicate the results (see Code Availability). \\

\noindent{\bf \AlgoDM~supports Bayesian belief updating.} \AlgoDM~policy associates higher probabilities to actions for which the combined action and ambiguity cost $\actioncost{k} + \radius{k}{k-1} + \costuncertainty{k}{k-1} $ is higher.  This means that the reason why an action is observed can be understood by estimating this combined cost:  \AlgoDM~supports a systematic framework to achieve this. Given a sequence of observed states/actions, ($\hat{\bv{x}}_{k-1}$, $\hat{\bv{u}}_k$) and the generative policy $\refpolicy{k}{k-1}$, the combined cost can be estimated by minimizing the negative log-likelihood.  The resulting optimization problem is convex if a widely adopted (see Methods) linear parametrization of the cost in terms of known (and arbitrary) features is available. Given $F$ state/action features and $G$ action features, this parametrization is $\sum_{i=1}^Fw_i\varphi_i(\bv{x}_{k-1},\bv{u}_k) + \sum_{i=1}^Gv_i\gamma_i(\bv{u}_k)$. Reconstructing the cost then amounts at finding the optimal weights that minimize the negative log-likelihood, with likelihood function being $\prod_{k=1}^M p^{\star}_k\left(\hat{\bv{u}}_k\mid\hat{\bv{x}}_{k-1};\bv{v},\bv{w}\right)$.  Here,  $M$ is the number of observed state/inputs pairs and  $p^{\star}_k\left(\hat{\bv{u}}_k\mid\hat{\bv{x}}_{k-1};\bv{v},\bv{w}\right)$ is{,  following the literature\cite{EG_HJ_CDV_GR:24},} the \AlgoDM~policy itself but with $-\actioncost{k} - \radius{k}{k-1} - \costuncertainty{k}{k-1} $ replaced by the linear parametrization ($\bv{v}$ and $\bv{w}$ are the stacks of the parametrization weights). We unpack the resulting optimization problem in the Methods. This convenient implication of \AlgoDM~policy allows one to reconstruct the cost driving the actions of the rovers using \AlgoDM~in our experiments and Fig. \ref{fig:results}f shows the outcome of this process. The similarity with Fig. \ref{fig:results}b is striking and to further assess the effectiveness of the reconstructed cost we carried out a number of additional experiments. In these experiments, the robots are equipped with \AlgoDM~policy but, crucially, $-\actioncost{k} - \radius{k}{k-1} - \costuncertainty{k}{k-1} $ is replaced with the reconstructed cost. The outcome from these experiments confirms the effectiveness of the results: as shown in Supplementary Fig.  1{d}, the robots are again able to fulfill their goal, despite ambiguity. Additionally, we also benchmarked our reconstruction result with other state-of-the-art approaches. Specifically, we {use} an algorithm from the literature that, building on maximum entropy\cite{BZ_AM_AB_AD:08}, is most related to our approach\cite{ZZ_etal:18}.  {This algorithm makes use of Monte Carlo sampling and soft-value iteration.} When {using} this algorithm -- after benchmarking it on simpler problems -- we observed that it would not converge to a reasonable estimate of the robot cost (see Supplementary Fig.  1e for the reconstructed cost using this {approach} and the Methods for details).  Since our proposed approach leads to a convex optimization problem, our cost reconstruction results could be implemented via off-the-shelf software tools.  The code for the implementation is provided (see Code Availability). \\

\noindent{\noindent{\bf Relaxing ambiguity yields maximum diffusion.} Maximum diffusion (MaxDiff)  is a policy computation framework that generalizes maximum entropy (MaxEnt) and inherits its  robustness properties.  It outperforms other state-of-the-art methods across popular benchmarks\cite{TB_AP_TM:24,TB:24}.  We  show that the distributionally robust free energy principle (Fig. \ref{fig:DRFREE}a) can recover, with a proper choice of $q_{0:N}$, the MaxDiff objective when ambiguity is relaxed.  This explicitly connects \AlgoDM~to MaxDiff and -- through it -- to a broader literature on robust decision-making (Sec. S2 of \SI).  In MaxEnt -- and   MaxDiff -- robustness guarantees stem from the entropy of the optimal policy \cite{BE_SL:22}, with explicit a-posteriori bounds on the ambiguity set over which the policy guarantees robustness available for discrete settings and featuring a constant radius of ambiguity \cite[Lemma 4.3]{BE_SL:22}.  To compute policies that robustly maximize a reward,  MaxEnt must be used with an auxiliary, pessimistic, reward\cite{BE_SL:22}.   In contrast,  by tackling the problem in Fig. \ref{fig:DRFREE}a,  \AlgoDM~defines robustness guarantees directly in the problem formulation,  explicitly via the ambiguity set.  As a result,  \AlgoDM~policy is guaranteed to be robust across this  ambiguity set.  As detailed in Sec.  S2 of \SI, \AlgoDM~is, to our knowledge,  the full min-max problem in Fig. \ref{fig:DRFREE}a -- featuring at the same time a free energy objective and distributionally robust constraints -- remains a challenge for many methods\cite{BE_SL:22,TB_AP_TM:24,TB:24,JM_KH_HA_SS_DC_JP:22}. This is not just a theoretical achievement uniquely positioning \AlgoDM~in the literature -- we explore its implications by revisiting our robot navigation task:  we equip \AlgoDM~with a generative model that recovers MaxDiff objective and compare their performance. The experiments reveal that \AlgoDM~succeeds in settings where MaxDiff fails.  This is because  \AlgoDM~not only retains the desirable properties of MaxDiff, but also guarantees them in the worst case over the ambiguity set.}

{In MaxDiff, given some initial state $\bv{x}_0$, policy computation is framed as minimizing in the policy space $\DKL{p_{0:N}}{p_{\max}(\bv{x}_{0:N},\bv{u}_{1:N})}$. This is the KL divergence between (using the time-indexing and notation in Fig. \ref{fig:DRFREE}a for consistency) $p_{0:N}$ and $p_{\max}(\bv{x}_{0:N},\bv{u}_{1:N})  = \frac{1}{Z} \prod_{k=1}^N p_{\max}(\bv{x}_{k}\mid\bv{x}_{k-1})\exp\left(r(\bv{x}_{k},\bv{u}_k)\right)$.  In this last expression,   $r(\bv{x}_{k},\bv{u}_k)$ is the state/action reward when the agent transitions in state $\bv{x}_k$ under action $\bv{u}_k$,  $Z$ is the normalizer and $p_{\max}(\bv{x}_{k}\mid\bv{x}_{k-1})$ is the maximum entropy sample path probability\cite{TB_AP_TM:24}. On the other hand,  the distributionally robust free energy principle (Fig. \ref{fig:DRFREE}a) is equivalent  to 
\begin{equation}\label{eqn:KL_div_DRFREE}
\underset{\left\{\policy{k}{k-1}\right\}_{1:N}}{\min}\ \ \underset{\plant{k}{k-1}\in\ball{k}{k-1}}{\max}
\DKL{p_{0:N}}{\tilde q_{0:N}}
\end{equation}
in the sense that the optimal solution of~\eqref{eqn:KL_div_DRFREE} is the same as the optimal solution of the problem in Fig. \ref{fig:DRFREE}a (see Methods).  In the above expression,  $\tilde q_{0:N} = \frac{1}{Z} q_{0:N}\exp(-\sum_{k=1}^N(\statecost{k}+\actioncost{k}))$ and $Z$ is again a normalizer.  When there is no ambiguity,  the optimization problem in~\eqref{eqn:KL_div_DRFREE} is relaxed and it becomes 
\begin{equation}\label{eqn:KL_div_DRFREE_no_ambiguity}
\underset{\left\{\policy{k}{k-1}\right\}_{1:N}}{\min} \DKL{p_{0:N}}{\tilde q_{0:N}}
\end{equation}
Given an initial state $\bv{x}_0$,  as we unpack in the Methods, this problem has the same optimal solution as the MaxDiff objective when the reward is $- \statecost{k} - \actioncost{k}$ and in the \AlgoDM~generative model -- which we recall is defined as $ q_0(\bv{x}_0)\prod_{k=1}^N\refplant{k}{k-1}\refpolicy{k}{k-1}$ --  we set $\refplant{k}{k-1}$  to be the maximum entropy sample path probability and $\refpolicy{k}{k-1}$ to be uniform. }

\begin{figure*}[!h]
	\centering
\noindent\includegraphics[width=\columnwidth]{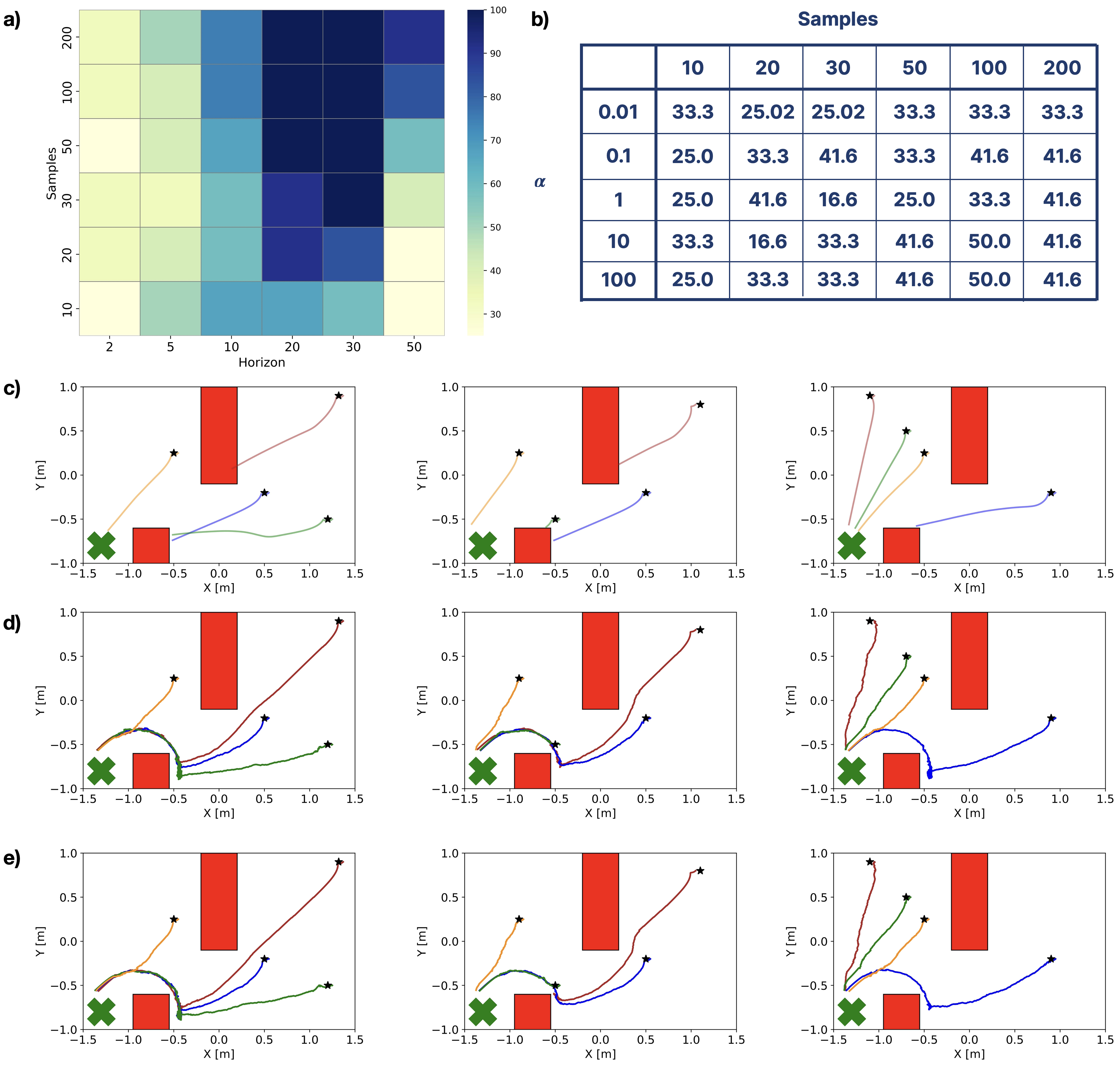}  
\caption{{{\bf \AlgoDM~and MaxDiff. a.} MaxDiff success rates for different values of the sampling size and planning horizon.  Experiments highlight a sweetspot in the hyperparameters with $100\%$ success rate. Worst rates are obtained for low horizons, where the success rate is between $25\%$ and approximately $40\%$.  All experiments are performed with the temperature-like hyperparameter $\alpha$ set to $0.1$.  Data for each cells obtained from $12$ experiments corresponding to the initial conditions in Fig. \ref{fig:results}c.  {\bf b.} Success rates  for different values of $\alpha$ and samples when horizon is set to $2$.  Success rates are consistent with the previous panel --  for the best combination of parameters, MaxDiff agent completes the task half of the times.  See Supplementary Fig.   {7} for a complementary set of MaxDiff experiments.  {\bf c.} Robot trajectories  using the MaxDiff policy when the horizon is equal to $2$ and samples is set to $50$.  MaxDiff  fulfills the task when the shortest path is obstacle-free.  {\bf d.} \AlgoDM~allows the robot to complete the task when it is equipped with a generative model from MaxDiff computed using the same set of hyperparameters from the previous panel.   {\bf e.} This desirable behavior is confirmed even when samples is decreased to $10$. See Methods  and \SI~for details. }} 
 \label{fig:MaxDiff}
\end{figure*}

{The above derivations show that relaxing ambiguity in \AlgoDM~yields, with a properly defined $q_{0:N}$, the MaxDiff objective -- provided that the rewards are the same.  This means that, in this setting, \AlgoDM~guarantees the desirable MaxDiff properties in the worst case,  over  $\ball{k}{k-1}$. To evaluate the implications of this finding,  we revisit the robot navigation task.  We equip \AlgoDM~with $q_{0:N}$ defined as described above, in accordance with MaxDiff, and compare \AlgoDM~with MaxDiff itself.  In the experiments,  the cost is again the one of Fig. \ref{fig:results}b and again the agents have access to  $\nominalplant{k}{k-1}$. Initial conditions are the same as in Fig. \ref{fig:results}c.  Given this setting,  Fig. \ref{fig:MaxDiff}a  summarizes the success rates of MaxDiff -- i.e.,  the number of times the agent successfully reaches the goal -- for different values  of key hyperparameters\cite{TB_AP_TM:24}: {samples} and {horizon}, used to compute the maximum entropy sample path probability.  The figure shows a {sweetspot} where $100\%$ success rate is achieved,  yielding two key insights.  First,  increasing  samples  increases the success rate.  Second -- and rather counter-intuitively -- planning horizons that are {too} long yield a decrease in the success rate and this might be an effect of planning under a {wrong} model.  Worst performance are obtained with  horizon set to $2$ and Fig. \ref{fig:MaxDiff}b confirms that the  success rates remain consistent when an additional temperature-like\cite{TB_AP_TM:24} hyperparameter is changed.   Given this analysis, to compare \AlgoDM~with MaxDiff, we equip \AlgoDM~with the generative model computed in accordance with MaxDiff (with horizon set to $2$ and samples to $50$).  In this setting,  MaxDiff  successfully completes the task when the shortest path between the robot initial position and the goal is obstacle free (Fig.  \ref{fig:MaxDiff}c).  In contrast,  in this very same setting,  the experiments show that \AlgoDM~--  computing reactive actions -- allows the robot to consistently complete its task (Fig. \ref{fig:MaxDiff}d).  This desirable behavior is confirmed when samples is decreased to $10$  (Fig. \ref{fig:MaxDiff}e).  In summary,  the experiments confirm that \AlgoDM~succeeds in settings where MaxDiff fails.  Experiments details are reported in  Methods ({Experiments settings}) and \SI~(Sec.  S6 and Tab.  S-1). See also code availability.}

\begin{figure}[ht]
  \centering
  \includegraphics[width=0.6\textwidth]{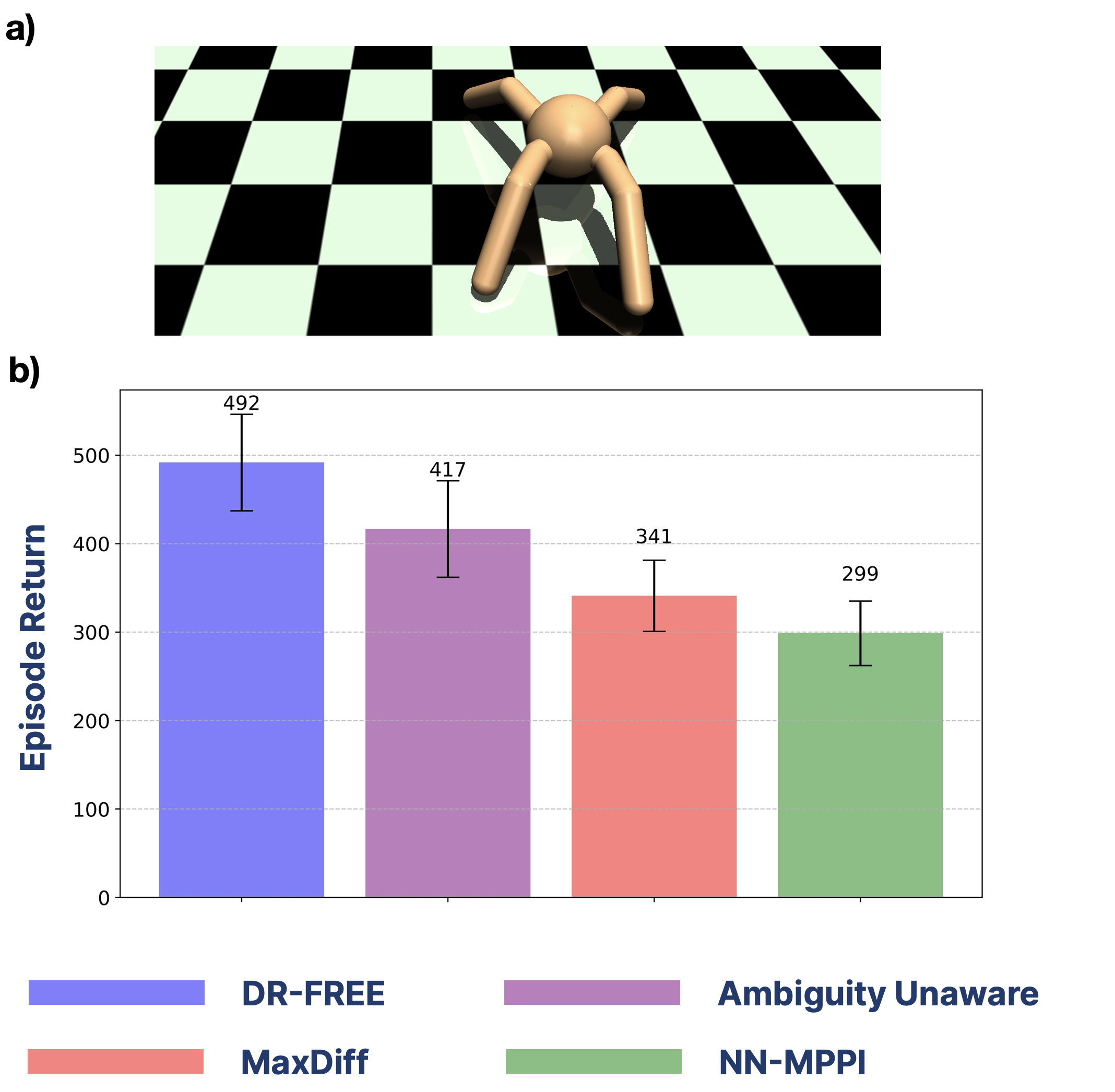}
  \caption{{{\bf Ant experiments.  a.} Screenshot from the MuJoCo environment ({Ant v-3}). The state space is $29$-dimensional and the action space is $8$-dimensional.  {\bf b.} Performance comparison.  Charts show means and bars standard deviations from the means across $30$ experiments.   In some episodes,  the ambiguity unaware, MaxDiff and NN-MPPI agents terminate prematurely due to the Ant becoming unhealthy; rewards were set to zero from that point to the end of the episode.  The Ant becomes unhealthy in $6\%$ of the episodes for the ambiguity unaware agent, $20\%$ for MaxDiff and $23\%$ for NN-MPPI.  In contrast, the Ant remains healthy in all \AlgoDM~experiments. As in previous experiments,  \AlgoDM~is used to compute reactive actions.  The ambiguity-unaware policy\cite{EG_HJ_CDV_GR:24} corresponds to \AlgoDM~with the ambiguity radius set to zero.
  }}
  \label{fig:Ant}
\end{figure}

 {Finally, we evaluate \AlgoDM~in the {MuJoCo}\cite{ET_TE_YT:12}   {Ant} environment (Fig.  \ref{fig:Ant}a).  The goal is for the quadruped agent to move forward along the $x$-axis while maintaining an  upright posture. Each episode lasts $1000$ steps, unless the Ant becomes {unhealthy} -- a terminal condition defined in the standard environment that  indicates failure.  We compare \AlgoDM~with all previously considered methods, as well as with model-predictive path integral control\cite{GW_NW_BG_PD_JR_BB_ET:17} (NN-MPPI). Across all experiments, the agents have access to the trained model $\nominalplant{k}{k-1}$ and not to $\plant{k}{k-1}$. The trained model was obtained using the same neural network architecture as in the original MaxDiff paper\cite{TB_AP_TM:24}, which also included benchmarks with NN-MPPI.  The cost provided to the agents is the same across all the experiments and corresponds to the negative reward defined by the standard environment.  Fig.  \ref{fig:Ant}b shows the experimental results for this setting.  The experiments yield two main observations.  First, \AlgoDM~outperforms all comparison methods on average, and even the highest error bars (standard deviations from the mean) of the other methods do not surpass \AlgoDM~average return. Second,  in some of the trials, the other methods would terminate prematurely the episode due to the Ant becoming unhealthy.  In contrast, across all \AlgoDM~experiments the Ant is always kept healthy and therefore episodes do not terminate prematurely.  See {Experiments Settings} in Methods and \SI~for details;  code also provided.}

\section*{Discussion}

\noindent Robustness is a core requirement for intelligent agents that need to operate in the real world. Rather than leaving its fulfillment to -- quoting the literature\cite{KC_etal:24} -- an {emergent and potentially brittle} property from training, \AlgoDM~ensures  this core requirement by design, building on the minimization of the free energy and installing sequential policy optimization into a rigorous (variational or Bayesian) framework.  \AlgoDM~provides not only a free energy principle that accounts for environmental ambiguity, but also the resolution engine to address the resulting sequential policy optimization framework.  This milestone is important because addresses a {challenge} for intelligent machines operating in open-worlds.  In doing so, \AlgoDM~elucidates the mechanistic role of ambiguity on optimal decisions and its policy supports (Bayesian) belief-based updates.   \AlgoDM~establishes what are the limits of performance in the face of ambiguity, showing that,  at a very fundamental level, it is impossible for an agent affected by ambiguity to outperform an ambiguity-free free energy minimizing agent. These analytic results are confirmed by our experiments.

In the  {navigation} experiments, we compared the behaviors of an ambiguity-unaware free energy minimizing agent\cite{EG_HJ_CDV_GR:24} with the behavior of an agent equipped with \AlgoDM. All the experiments  {show} that \AlgoDM~is essential for the robot to successfully  {complete the task} amid ambiguity{, and this is confirmed when we consider additional benchmarks and different environments.}   \AlgoDM~enables to reconstruct the cost functions that underwrote superior performance over related methods.  Our experimental setting is exemplary not only for intelligent machines, underscoring the severe consequences of ambiguity, but also for natural intelligence. For example, through evolutionary adaptation, bacteria can navigate unknown environments and this crucial ability for survival is achieved with little or no training. \AlgoDM~suggests that this may be possible if bacteria follow a decision-making strategy that, while simple, foresees a robustness promoting step. Run-and-tumble motions\cite{ZA_etal:20,AT_AS_CB:20} might be an astute way to achieve this: interpreted through \AlgoDM, tumbles might be driven by free energy maximization, needed to quantify across the environment a cost of ambiguity,  and runs would be sampled from a free-energy minimizing policy that considers this cost. 

{\AlgoDM~offers a model for robust decision making via free energy minimization, with robustness guarantees defined in the problem formulation -- it  also opens a number of interdisciplinary research questions.  First,  our results suggest that a promising research direction originating from this work is to integrate \AlgoDM~with perception and learning{, coupling training with policy computation}.  This framework would embed distributional constraints in the formulation of the policy computation problem, as in \AlgoDM, while retaining perception and learning mechanisms inspired by, for example,  MaxDiff and/or evidence free energy minimization.   {The framework would motivate analytical studies to quantify the benefits of integrated learning over an offline pipeline.} Along these lines,   analytical studies should be developed to extend our framework so that it can explicitly account for ambiguities in the agent cost/reward.  Second,  \AlgoDM~takes as input the ambiguity radius and this motivates the derivation of a radius estimation mechanism within our model.  Through our analytic results we know that reducing ambiguity improves performance; hence,  integrating in our framework a method to learn ambiguity would be a promising step towards agents that are not only robust, but also {antifragile}\cite{NT:12}.  Finally, our experiments  prompt a broader question: {what} makes for a {good} generative model/planning horizon in the presence of ambiguity? The answer   remains elusive -- \AlgoDM~guarantees robustness against ambiguity and experiments suggest that it compensates for poor planning/models; however,  with e.g.,} more task-oriented model{/planning},  ambiguity-unaware agents could succeed. {This yields a follow-up question.  I}n challenging environments,  is a specialized model better than a  multi-purpose one for survival?

If, quoting the popular aphorism, {all models are wrong, but some are useful}, then relaxing the requirements on training, \AlgoDM~makes more models useful. This is achieved by departing from  views that emphasize the role, and the importance, of training: in \AlgoDM~the emphasis is instead on rigorously installing robustness into decision-making mechanisms. With its robust free energy minimization principle and resolution engine, \AlgoDM~suggests that, following this path, intelligent machines can recover robust policies from largely imperfect, or even poor,  models. We hope that this work may inspire both the deployment of our free energy  model in multi-agent settings (with heterogeneous agents such as drones, autonomous vessels and humans) across a broad range of application domains and, combining \AlgoDM~with Deep RL,  lead to  learning schemes that -- learning ambiguity --  succeed when classic methods fail.  At a perhaps deeper level { -- as ambiguity is a key theme (see also Sec.  S7 in \SI) across, e.g., psychology, economics and neuroscience\cite{Barto2013,Hsu2005,Zak2004} -- we hope that} this work may provide the foundation for a biologically plausible neural explanation of how natural agents -- with little or no training -- can operate robustly in challenging environments.

\section*{Methods}
    
\begingroup
\fontsize{8}{8}
The agent has access to: (i) the generative model, $q_{0:N}$; (ii) the loss, specified via state/action costs $c_{k}^{(x)}:\sX\to\R$, $c_{k}^{(u)}:\sU\to\R$, with $\sX$ and $\sU$ being the state and action spaces {(see Sec.  S1 and Sec.  S3 of the \SI~for notation and details)}; (iii) the trained model $\nominalplant{k}{k-1}$.  \\

\noindent{\bf Distributionally Robust Free Energy Principle.} Model ambiguities are specified via the  ambiguity set around $\nominalplant{k}{k-1}$, i.e., $\ball{k}{k-1}$. {This is the set of all models with statistical complexity of at most $\radius{k}{k-1}$ from $\nominalplant{k}{k-1}$. Formally,  {$\ball{k}{k-1}$} is defined as the set}
\begin{align*}
{\left\{\right.} \plant{k}{k-1}\in\sD : \DKL{\plant{k}{k-1}}{\nominalplant{k}{k-1}} \le \radius{k}{k-1},    \support{\plant{k}{k-1}}\subseteq\support{\refplant{k}{k-1}}{\left.\right\}}.
\end{align*}
The symbol $\sD$ stands for the space of densities and $\support$ for the support. The ambiguity set captures all the models that have statistical complexity of at most $\radius{k}{k-1}$ from the trained model and that have a support included in the generative model. This second property explicitly built in the ambiguity set makes the optimization meaningful as violation of the property would make the optimal free energy infinite. The radius $\radius{k}{k-1}$ is positive and bounded. As summarized in Fig. \ref{fig:DRFREE}a,  the principle    in the main text yields the following sequential policy optimization framework:
\begin{equation*}
    \begin{aligned}
& \left\{\optimalpolicy{k}{k-1}\right\}_{1:N}\in & \underset{\left\{ \policy{k}{k-1}\right\}_{1:N}}{\argmin}  \underset{\left\{\plant{k}{k-1}\right\}_{1:N}}{\max} & \overbrace{\DKL{p_{0:N}}{q_{0:N}}}^{\text{Complexity}} + \overbrace{\E_{p_{0:N}}\left[\sum_{k=1}^N \statecostexpectation{k} + \actioncostexpectation{k} \right]}^{\text{Expected Loss}} \\
&   &  \st  &    \underbrace{\plant{k}{k-1} \in\ball{k}{k-1}}_{\text{Ambiguity}}, \ \ \forall k =1,\ldots,N.
    \end{aligned}
\end{equation*}
This is an extension of the free energy principle\cite{KF_LDC_NS_CH_KU_GP_AG_TP:23} accounting for policy robustness against model ambiguities. We are not aware of any other free energy account that considers this setting and the corresponding infinite-dimensional optimization framework cannot be solved with excellent methods. When the ambiguity constraint is removed and the loss is the negative log-likelihood, our formulation reduces to the {expected free energy} minimization in active inference.  In this special case, the expected complexity (i.e., ambiguity cost) becomes {risk}; namely, the KL divergence between inferred and preferred (i.e., trained) outcomes. The expected free energy can be expressed as {risk} plus {ambiguity}; however, the {ambiguity} in the expected free energy pertains to the ambiguity of likelihood mappings in the generative model (i.e., conditional entropy), not ambiguity {about} the generative model considered in our free energy model. In both robust and conventional active inference, the complexity term establishes a close relationship between optimal control and Jaynes’ maximum caliber (a.k.a., path entropy) or minimum entropy production principle\cite{EJ:80,HH_JP:21}.  It is useful to note that, offering a generalization to free energy minimization in active inference, our robust formulation yields as special cases other popular computational models such as KL control\cite{ET:09},control as inference\cite{JB_WW_HK:10}, and the Linear Quadratic Gaussian Regulator. Additionally,  when the loss is the negative log-likelihood,  the negative of the variational free energy in the cost functional is the evidence lower bound\cite{KM:23} a key concept in machine learning and inverse reinforcement learning\cite{BZ_AM_AB_AD:08}. With its resolution engine,  \AlgoDM~shows that in this very broad set-up the optimal policy can still be computed. \\

\noindent{\noindent{\bf Drawing the connection between MaxDiff and \AlgoDM.} We start with showing that the robust free energy principle formulation in Fig. \ref{fig:DRFREE}a has the same optimal solution as~\eqref{eqn:KL_div_DRFREE}.  We have the following identity:
$$
\DKL{p_{0:N}}{q_{0:N}} + \E_{p_{0:N}}\left[\sum_{k=1}^N \statecostexpectation{k} + \actioncostexpectation{k} \right] = \DKL{p_{0:N}}{\tilde q_{0:N}} - \ln\E_{q_{0:N}}\left[\exp\left(-\sum_{k=1}^N \left(\statecostexpectation{k} + \actioncostexpectation{k}\right)\right)\right]
$$
The left hand-side is the objective of Fig. \ref{fig:DRFREE}a. In the right-hand side, $\tilde q_{0:N}$ is given in the main text and  $\E_{q_{0:N}}\left[\exp\left(-\sum_{k=1}^N \left(\statecostexpectation{k} + \actioncostexpectation{k}\right)\right)\right]$ is the normalizing constant denoted by $Z$ in the main text.  The last term in the right-hand side does not depend on the decision variables and this yields that the min-max problem in Fig. \ref{fig:DRFREE}a has the same optimal solution as~\eqref{eqn:KL_div_DRFREE}.  Next, we show why -- with the choice of $q_{0:N}$ described in the main text -- \AlgoDM~yields the MaxDiff objective when there is no ambiguity.  In this case,  the ambiguity constraint is relaxed  and \AlgoDM~min-max problem becomes 
\begin{equation}\label{eqn:KL_div_DRFREE_no_ambiguity_methods}
\underset{\left\{\policy{k}{k-1}\right\}_{1:N}}{\min} \DKL{p_{0:N}}{\frac{1}{Z} q_{0:N}\exp\left(-\sum_{k=1}^N\left(\statecost{k}+\actioncost{k}\right)\right)}
\end{equation}
This is the problem given in~\eqref{eqn:KL_div_DRFREE_no_ambiguity} but with $\tilde{q}_{0:N}$ explicitly included in the objective functional. To establish the connection between MaxDiff and \AlgoDM~we recall that the MaxDiff objective consists in minimizing in the policy space the KL divergence between $p_{0:N}$ and $p_{\max}(\bv{x}_{0:N},\bv{u}_{1:N})$, which can be conveniently written as  $ \frac{1}{Z} \prod_{k=1}^N \left(p_{\max}(\bv{x}_{k}\mid\bv{x}_{k-1})\right)\exp\left(\sum_{k=1}^N r(\bv{x}_{k},\bv{u}_k)\right)$.  Now, this minimization problem has the same optimal solution of
\begin{equation}\label{eqn:KL_div_MaxDiff_methods}
\underset{\left\{\policy{k}{k-1}\right\}_{1:N}}{\min}\DKL{p_{0:N}}{\frac{1}{Z} \prod_{k=1}^N \left(p_{\max}(\bv{x}_{k}\mid\bv{x}_{k-1})\bar{p}(\bv{u_k}\mid\bv{x}_{k-1})\right)\exp\left(\sum_{k=1}^N r(\bv{x}_{k},\bv{u}_k)\right)}
\end{equation}
when $\bar{p}(\bv{u_k}\mid\bv{x}_{k-1})$ is uniform.  The equivalence can be shown by noticing that this reformulation has been obtained by adding and subtracting the constant quantity $\ln\prod_{k=1}^N\bar{p}(\bv{u_k}\mid\bv{x}_{k-1})$ to the MaxDiff cost functional.  The similarity between~\eqref{eqn:KL_div_MaxDiff_methods} and ~\eqref{eqn:KL_div_DRFREE_no_ambiguity_methods} is striking.  In particular, the two problems are the same when: (i) $\bv{x}_0$ is given, (ii) in $q_{0:N}$ -- which we recall is defined as $p(\bv{x}_0)\prod_{k=1}^N\refplant{k}{k-1}\refpolicy{k}{k-1}$ -- we have $\refplant{k}{k-1}$ set to $p_{\max}(\bv{x}_{k}\mid\bv{x}_{k-1})$ and $\refpolicy{k}{k-1}$ uniform; (iii) $r(\bv{x}_k,\bv{u}_k) = -\statecost{k}-\actioncost{k}$.  This establishes the connection between \AlgoDM~and MaxDiff objective from the Results.}\\

\noindent{\bf Resolution engine.} Both the variational free energy and the ambiguity constraint are nonlinear in the infinite-dimensional decision variables and this poses a number of challenges that are addressed with our resolution engine.  The resolution engine allows to tackle the sequential policy optimization framework arising from our robust free energy principle. We detail here the resolution engine and refer to \SI~for the formal treatment. Our starting point is the robust free energy principle formulated via the above sequential policy optimization framework. This can be solved via a backward recursion where, starting from $k=N$, at each $k$ the following optimization problem needs to be solved:
\begin{equation*}
 \begin{aligned}
&  & \underset{\shortpolicy{k}{k-1}}{\min} \DKL{\shortpolicy{k}{k-1}}{\shortrefpolicy{k}{k-1}} +\E_{\shortpolicy{k}{k-1}}\left[\actioncostexpectation{k}\right] +  \underset{\shortplant{k}{k-1}}{\max} & \E_{\shortpolicy{k}{k-1}}\left[\DKL{\shortplant{k}{k-1}}{\shortrefplant{k}{k-1}}+\E_{\shortplant{k}{k-1}}\left[\costtotexpectation{k}\right]\right]\\
&   &  \st  &    \ \shortplant{k}{k-1} \in\ball{k}{k-1}.
    \end{aligned}
 \end{equation*}
In the above expression, for compactness we used the shorthand notations $\shortpolicy{k}{k-1}$, $\shortrefpolicy{k}{k-1}$, $\shortplant{k}{k-1}$ and $\shortrefplant{k}{k-1}$ for $\policy{k}{k-1}$, $\refpolicy{k}{k-1}$, $\plant{k}{k-1}$ and $\refplant{k}{k-1}$, respectively. The term $\costtot{k}$ is the cost-to-go. This is given by $\costtot{k} = \statecost{k} + \costtogo{k+1}{k}$, where $\costtogo{k+1}{k}$ is the smallest free energy that can be achieved by the agent at $k+1$. That is, $\costtogo{k+1}{k}$ is the optimal solution of the above optimization problem evaluated at $k+1$. When $k=N$, $\costtogo{N+1}{N}$ is initialized at $0$. This means that, for {reactive} actions, e.g., reflexes, $\costtot{k} = \statecost{k}$. The above reformulation is convenient because it reveals that, at each $k$, $\optimalpolicy{k}{k-1}$ can be computed via a bi-level optimization approach, consisting in first maximizing over $\plant{k}{k-1}$, obtaining the maximum expected variational free energy across all possible environments in the ambiguity set, to finally minimize over the policies. Crucially, this means that to make optimal decisions, the agent does not need to know the environment that maximizes the free energy but rather it only needs to know what the actual maximum free energy is. In turn, this can be found by first tackling the problem in green in Fig. \ref{fig:DRFREE}b, i.e., finding the cost of ambiguity, and then taking the expectation $\E_{\policy{k}{k-1}}\left[\cdot\right]$.  While the problem in green in Fig. \ref{fig:DRFREE}b is  infinite-dimensional, \AlgoDM~finds the cost of uncertainty by solving a convex and scalar optimization problem. This is possible because the optimal value of the problem in green in Fig. \ref{fig:DRFREE}b equals $\radius{k}{k-1}+\min_{\alpha \ge 0}\Vxtilde{k}{k-1}$. In this expression, $\alpha$ is a scalar decision variable and $\Vxtilde{k}{k-1}$, detailed in the \SI, is a scalar function of $\alpha$, convex for all $\alpha \ge 0$. The  global non-negative minimum of $\Vxtilde{k}{k-1}$ is  $\costuncertainty{k}{k-1}$. In summary, the free energy maximization step can be conveniently solved with off-the-shelf software tools. In \AlgoDM, the cost of ambiguity promotes  robustness and contributes to the expected loss for the subsequent minimization problem in Fig. \ref{fig:DRFREE}b. The optimal solution of this class of problems has an explicit expression ($\optimalpolicy{k}{k-1}$ in  Fig. \ref{fig:DRFREE}b) and  the optimal value is $\costtogo{k}{k-1}$ used at the next step in the recursion. Derivations in \SI.\\

\noindent {\bf Why is it always better to be ambiguity-aware.} As ambiguity vanishes,  $\costuncertainty{k}{k-1}$ becomes  $\DKL{\shortnominalplant{k}{k-1}}{\shortrefplant{k}{k-1}} + \E_{\shortnominalplant{k}{k-1}}\left[\costtotexpectation{k}\right]$. Thus, the \AlgoDM~policy becomes
 $$
\optimalpolicy{k}{k-1}  = \frac{\shortrefpolicy{k}{k-1}\exp\left(-\DKL{\nominalplant{k}{k-1}}{\refplant{k}{k-1}} - \E_{\plant{k}{k-1}}\left[\costtotexpectation{k}\right]
{-\actioncost{k}}\right)}{\int \refpolicy{k}{k-1}\exp\left(-\DKL{\nominalplant{k}{k-1}}{\refplant{k}{k-1}}-\E_{\plant{k}{k-1}}\left[\costtotexpectation{k}\right]{-\actioncost{k}}\right)d\bv{u}_{k}}.
    $$
This is the optimal policy of an ambiguity-free agent\cite{EG_HJ_CDV_GR:24} (with $\plant{k}{k-1} = \nominalplant{k}{k-1}$).  Given the current state $\bv{x}_{k-1}$, the optimal cost  is 
$$
-\ln\int_{\sU} \refpolicy{k}{k-1}\exp\left(-\DKL{\nominalplant{k}{k-1}}{\refplant{k}{k-1}}-\E_{\nominalplant{k}{k-1}}[\costtot{k}]-\actioncost{k}\right)d\bv{u}_k.
$$
This is smaller than the cost achieved by the agent affected by ambiguity. In fact, when there is ambiguity, the \AlgoDM~policy achieves the optimal cost
$ - \ln \int \refpolicy{k}{k-1}\exp\left(-\radius{k}{k-1} - \costuncertainty{k}{k-1}{-\actioncost{k}}\right)d\bv{u}_k$ and $\DKL{\shortnominalplant{k}{k-1}}{\shortrefplant{k}{k-1}}+\E_{\shortnominalplant{k}{k-1}}[\statecostexpectation{k}]<\radius{k}{k-1}+\costuncertainty{k}{k-1}$. See Sec. S4 in the \SI~for the formal details. \\

\noindent {\bf Why \AlgoDM~Supports Bayesian belief updating.}  The approach adopts a widely used parametrization\cite{IG_JB_AC:16,BZ_AM_AB_AD:08,KD_ET:10,EG_HJ_CDV_GR:24} of the cost in terms of $F$ state-action features, $\varphi_{i}(\bv{x}_{k-1},\bv{u}_k)$,  and $G$ action features, $\gamma_i(\bv{u}_k)$. No assumptions are made on the features, which can be e.g., nonlinear.  With this parametrization,  given $M$ observed actions/state pairs, the likelihood function {-- inspired by the literature\cite{KD_ET:10,EG_HJ_CDV_GR:24} --} is
\begin{align*}
 \prod_{k=1}^M{\frac{\refpolicyobs{k}{k-1}\exp\left(\sum_{i=1}^{F} w_i\varphi_{i}\left(\hat{\bv{x}}_{k-1},\hat{\bv{u}}_k\right) + \sum_{i=1}^Gv_{i}\gamma_i\left(\hat{\bv{u}}_k\right)\right)}{\int \refpolicygiven{k}{k-1}\exp\left(\sum_{i=1}^{F} w_i\varphi_{i}\left(\hat{\bv{x}}_{k-1},{\bv{u}}_k\right) + \sum_{i=1}^Gv_{i}\gamma_i\left({\bv{u}}_k\right)\right)d\bv{u}_{k}}},
\end{align*}
where $\bv{w}$ and $\bv{v}$ are the stacks of the weights $w_i$ and $v_i$, respectively.  The negative log-likelihood\cite{KD_ET:10,EG_HJ_CDV_GR:24} is then given by
\begin{align*}
- L (\bv{w},\bv{v}) & = - \sum_{k=1}^M\left(\ln\refpolicyobs{k}{k-1} + \sum_{i=1}^{F} w_i\varphi_{i}\left(\hat{\bv{x}}_{k-1},\hat{\bv{u}}_k\right) + \sum_{i=1}^Gv_{i}\gamma_i\left(\hat{\bv{u}}_k\right) \right. \\
& \left. - \ln\int \refpolicygiven{k}{k-1}\exp\left(\sum_{i=1}^{F} w_i\varphi_{i}\left(\hat{\bv{x}}_{k-1},{\bv{u}}_k\right) + \sum_{i=1}^Gv_{i}\gamma_i\left({\bv{u}}_k\right)\right)d\bv{u}_{k} \right).
\end{align*}
The cost reconstruction in the main paper is then obtained by finding the weights that are optimal for the problem $\min_{\bv{w},\bv{v}}- L (\bv{w},\bv{v})$, after dropping the first term from the cost because it does not depend on the weights.  Convexity of the problem follows because\cite{KD_ET:10,EG_HJ_CDV_GR:24} the cost functional is a conical combination of convex functions.   See \SI.\\

\noindent{\bf Experiments settings.} \AlgoDM~was turned into Algorithm 1 shown in the \SI~and engineered to be deployed on the agents (see Code Availability). In the robot experiments $\plant{k}{k-1}$ is $\mathcal{N}\left(\bv{x}_{k-1}+\bv{u}_{k}dt,\Sigma\right)$ with \(\bv{\Sigma} = \begin{bmatrix} 0.001 & 0.0002 \\ 0.0002 & 0.001 \end{bmatrix}\), and where \(\bv{x}_{k}=[p_{x,k},p_{y,k}]^{T}\) is the position of the robot at time-step \(k\), \(\bv{u}_k=[v_{x,k},v_{y,k}]^{T}\) is the input velocity vector, and \(dt=0.033\text{s}\) is the Robotarium time-step.  In these experiments, the state space is $[-1.5,1.5]\times[-1,1]$ {m}, matching the work area, and the action space is $[-0.5,0.5]\times[-0.5,0.5]$ {m/s}.  In accordance with the maximum allowed speed in the platform,  the inputs to the robot were automatically clipped by the Robotarium when the  speed was higher than $0.2$ {m/s}.  \AlgoDM~does not have access to $\plant{k}{k-1}$. Its trained model, $\nominalplant{k}{k-1}$, is learned via Gaussian Processes (GPs) with covariance function being an exponential kernel. The trained model was learned in stages and the model learned in a given stage would also use the data from the previous stages. The data for the training had a bias: the input $\bv{u}_k$ {was sampled from a uniform policy and the next observed robot position was corrupted by adding}  a quantity proportional to the current  position ($0.1\bv{x}_{k-1}$). See \SI~for the details and {the data used for training}. This means that the models learned at each stage of  the training data were necessarily wrong: the parameters of the trained model at each stage of the training are in Supplementary Fig.  1a. For the generative model, $\refplant{k}{k-1}=\mathcal{N}\left(\bv{x}_{d},\Sigma_{\bv{x}}\right)$, with $\Sigma_{\bv{x}}=0.0001\bv{I}_{2}$, and $\refpolicy{k}{k-1}$ being the uniform distribution. Also, the ambiguity radius, $\radius{k}{k-1} = \DKL{\refplant{k}{k-1}}{\nominalplant{k}{k-1}}$  and clipped at $100$, is higher the farther the robot is from the goal position {(we recall  that the agent has access to both $\refplant{k}{k-1}$ and $\nominalplant{k}{k-1}$)}. This captures higher agent confidence as it gets closer to its goal destination (see  Supplementary Fig.  2 {, also reporting the number of steps recorded across the experiments}). The state cost in Fig. \ref{fig:results}b,  {adapted} from the literature\cite{EG_HJ_CDV_GR:24}, is 
$c(\bv{x}_{k}) = 50(\bv{x}_{k} - \bv{x}_{d})^{2}+20\sum_{i=1}^{n}g_{i}(\bv{x}_{k}) + 5b(\bv{x}_{k})$, where:
(i) \(\bv{x}_{d}\) is the goal destination and thus this term promotes goal-reaching;
(ii) $n = 6$, $g_i$ are Gaussians, $\mathcal{N}\left(\bv{o}_i,\Sigma_o\right)$, with $\boldsymbol{\Sigma}_o = 0.025\bv{I}_2$ ($\bv{I}_2$ is the identity matrix of dimension $2$). The $\bv{o}_i$'s are  in Supplementary Fig.  1c) and capture the presence of the obstacles. Hence, the second term penalizes proximity to obstacles;
(iii) \(b(\bv{x}_{k})\) is the boundary penalty term given by $b(\bv{x}_{k}):=\sum_{j=1}^{2}\left(\exp\left(-0.5\left(\frac{p_{x,k}-b_{x_{j}}}{\sigma}\right)^{2}\right)+\exp\left(-0.5\left(\frac{p_{y,k}-b_{y_{j}}}{\sigma}\right)^{2}\right)\right)/\sigma\sqrt{2\pi}$,
with $b_{x_j}$, $b_{y_j}$ representing the $j_{th}$ component of the boundary coordinates ($b_{x}=[-1.5,1.5]$, $b_{y}=[-1,1]$) and $\sigma=0.02$.  {In summary,  the cost embeds obstacle avoidance  in the formulation.} The optimal policy of the ambiguity-unaware free energy minimizing agent is available in  the literature\cite{EG_HJ_CDV_GR:24}. This is again exponential but, contrary to \AlgoDM~policy in Fig. \ref{fig:results}b, the exponent is now $-\DKL{\nominalplant{k}{k-1}}{\refplant{k}{k-1}}-\E_{\nominalplant{k}{k-1}}\left[\statecost{k}\right]$.  Across experiments,  the first term dominated the second, which accounted for obstacles (see \SI~for details). As a result,  the policy consistently directs the robot along the shortest path to the goal,  disregarding the presence of the obstacles.  This explains the behavior observed in Fig. \ref{fig:results}c. {The computation time reported in the main text were obtained from \AlgoDM~deployment on the Robotarium hardware -- measurements obtained using the {time()} function in  {Python} and averaging across the experiment.}
 
The benchmark for our belief updating is with respect to the Infinite Horizon Maximum Causal Entropy Inverse RL algorithm with Monte-Carlo policy evaluation {and soft-value iteration (not required in \AlgoDM~belief update)} from the literature\cite{ZZ_etal:18}.  In order to use this algorithm, we discretized the state space in a $50\times50$ grid (this step is not required within our results) and we needed to redefine the action space as \([-1.0,1.0]\times[-1.0,1.0]\). This was then discretized into a $5\times5$ grid. The corresponding reconstructed cost in Supplementary Fig.  1e was obtained with the same dataset and features used to obtain Fig. \ref{fig:results}f. After multiple trials and trying different settings, we were not able to obtain a reconstructed cost that was better than the one in Supplementary Fig.  1e. The settings used to obtain  Supplementary Fig.  1e are: (i) initial learning rate of $1$, with an exponential decay function to update the learning rate after each iteration; (ii) discount factor for soft-value iteration, $0.9$; (iii) initial feature weights randomly selected from a uniform distribution with support $[-100,100]$; (iv)  gradient descent stopping threshold,  $0.01$.  The code to replicate all the results is provided (see Code Availability). {For the comparison with MaxDiff\cite{TB_AP_TM:24} we used the  code available on the paper repository.  In the experiments reported in Fig. \ref{fig:MaxDiff},  consistently with all the other experiments, MaxDiff had access to $\nominalplant{k}{k-1}$ and cost/reward from Fig. \ref{fig:results}b.   In the experiments of the Supplementary Fig.   {7}b, MaxDiff learns both the reward and the model.  In this case, the data used by MaxDiff is corrupted by the same bias as the data used by \AlgoDM~($100000$ data points are used for learning, consistently with the original MaxDiff repository). Across all MaxDiff experiments,  again consistently with the original code,  the additional MaxDiff hyperparameters are $\gamma = 0.95$, $\lambda = 0.5$ and learning rate $0.0005$ (this last parameter is used only in the experiments where MaxDiff does not have access to the reward/model).  No changes to the original MaxDiff code were made to obtain the results.  The available MaxDiff implementation natively allows users to configure if the framework uses reward/model supplied by the environment. These details, together with a table summarizing all the parameter settings used in the experiments, are given in the \SI~(see at the end of Sec.  S5 and  Tab.  S-1).}  {For the experiments of Fig. \ref{fig:Ant},  the Ant becomes unhealthy when either the value of any of the states is no longer finite or the height of the torso is  too low/high (we refer to the environment documentation at \url{https://gymnasium.farama.org/environments/mujoco/ant/} for the standard definition of the unhealthy condition). The cost is the negative reward provided by the environment and the trained model $\nominalplant{k}{k-1}$ is learned  from $10000$ data points collected through random policy rollouts, using the same neural network as in the Ant experiments in the MaxDiff paper\cite{TB_AP_TM:24}. The same network -- providing as output mean and variance -- is used across all the experiments.   For the \AlgoDM~generative model,  $\refpolicy{k}{k-1}$ is uniform and $\refplant{k}{k-1}$ is set as an anisotropic Gaussian having the same mean provided by the neural network but with a different variance.  The ambiguity radius captures higher agent confidence as this is moving in the right direction.   See \SI~for details.  MaxDiff and NN-MPPI implementations are from the MaxDiff repository. }\\

\noindent{\bf Schematic figure generation.} The schematic figures were generated in Apple Keynote (v.  13.2).  Panels were assembled using the same software. 

\section*{Data Availability}
All (other) data needed to evaluate the conclusions in the paper and to replicate the experiments are available in the paper,  \SI~and accompanying code (see the  {Assets} folder in Code Availability).  A recording from the robot experiments, together with the figures of this paper, is available  at the folder  {Assets} of our repository\cite{repo}.

\section*{Code Availability}
Pseudocode for \AlgoDM~is provided in the \SI. The full code for \AlgoDM~to replicate all the experiments is provided at our repository\cite{repo}.  The folder  {Experiments} contains our \AlgoDM~implementation for the Robotarium experiments. The folder also contains: (i) the code for the ambiguity-unaware free energy minimizing agent; (ii) the data shown in Fig.  SI-8,  together with the GP models and the code to train the models; (iii) the code to replicate the results in Fig. \ref{fig:results}f and Fig. \ref{fig:results}e.   {The folder also contains the code for the experiments in Supplementary Fig.  4 and provides the instructions to replicate the experiments of Supplementary Fig.  5 and Supplementary Fig.  6.} The folder  {Belief Update Benchmark} contains the code to replicate our benchmarks for the belief updating results. The folder  {Assets} contains all the figures of this paper, the data from the experiments used to generate these figures,  and the movie from which the screen-shots of Fig.  \ref{fig:results}d were taken.  {The folder  {MaxDiff Benchmark} contains the code to replicate our MaxDiff benchmark experiments. We build upon the original code-base from the MaxDiff paper\cite{TB_AP_TM:24}, integrating it in the Robotarium Python environment.   {The sub-folder  {Ant Benchmark} contains the code for the  {Ant} experiments. MaxDiff and NN-MPPI implementations are from the literature\cite{TB_AP_TM:24}. } } 

{As highlighted in the Discussion, extending our analytical results to consider ambiguity inherently in the reward is an open theoretical research direction, interesting {per se}. Nevertheless, we now discuss how \AlgoDM~can be adapted to this setting.  To achieve this,  {quoting form} the literature\cite{BE_SL:22},  one could define a modified problem formulation where the {reward is appended to the observations available to the agent}. In this setting,  the reward becomes -- quoting the literature\cite{BE_SL:22} -- the last {coordinate of the observation}, so that it can be embedded into model ambiguity.}

\noindent{\bf{Acknowledgments.}} AH and HJ did this work while at University of Salerno.  HJ and GR  supported by the European Union-Next Generation EU Mission 4 Component 1 CUP E53D23014640001.  KF  supported by funding from the Wellcome Trust (Ref: 226793/Z/22/Z).  AS supported by MOST - Sustainable Mobility National Research Center and received funding from the European Union Next-GenerationEU (PIANO NAZIONALE DI RIPRESA E RESILIENZA (PNRR)–MISSIONE 4 COMPONENTE 2, INVESTIMENTO 1.4–D.D. 1033 17/06/2022) under Grant CN00000023. This document reflects only the authors’ views and opinions.  We acknowledge the use of ChatGPT for assistance in improving the wording and grammar of this document.  GR and HJ wish to thank Prof. Del Vecchio (Sannio University) who allowed HJ to perform early preliminary experiments before him joining University of Salerno.   GR thanks Prof. Francesco Bullo (University College Santa Barbara, USA),  Prof. Michael Richardson (Macquarie University, Australia) and Prof. Mirco Musolesi (University College London, UK) for the insightful discussions and comments on an early version of this paper. \\

\noindent{\bf{Author contributions.}}  KF and GR conceptualized, designed and formulated the research. AS and GR conceptualized, designed and formulated  resolution engine concepts.   {AS developed all the proofs with inputs from GR.} AS, GR and HJ revised the proofs.  GR and HJ designed the experiments. HJ, with inputs from GR and AS, implemented \AlgoDM, performed the experiments and obtained the corresponding figures and data.  \AlgoDM~code was revised by AS.  {All authors contributed to the interpretation of the results.} GR wrote the manuscript with inputs from all the authors.  All authors contributed to and edited the manuscript.  \\

\noindent\noindent{\bf Competing interests.} The authors declare no competing interests.

\endgroup


\begin{thebibliography}{1}

\bibitem{PW_etal_22}
Peter~R. Wurman, Samuel Barrett, Kenta Kawamoto, James MacGlashan, Kaushik Subramanian, Thomas~J. Walsh, Roberto Capobianco, Alisa Devlic, Franziska Eckert, Florian Fuchs, Leilani Gilpin, Piyush Khandelwal, Varun Kompella, HaoChih Lin, Patrick MacAlpine, Declan Oller, Takuma Seno, Craig Sherstan, Michael~D. Thomure, Houmehr Aghabozorgi, Leon Barrett, Rory Douglas, Dion Whitehead, Peter D\"{u}rr, Peter Stone, Michael Spranger, and Hiroaki Kitano.
\newblock Outracing champion {G}ran {T}urismo drivers with deep reinforcement learning.
\newblock {\em Nature}, 602(7896):223–228, February 2022.

\bibitem{VM_et_al:15}
Volodymyr Mnih, Koray Kavukcuoglu, David Silver, Andrei~A. Rusu, Joel Veness, Marc~G. Bellemare, Alex Graves, Martin Riedmiller, Andreas~K. Fidjeland, Georg Ostrovski, Stig Petersen, Charles Beattie, Amir Sadik, Ioannis Antonoglou, Helen King, Dharshan Kumaran, Daan Wierstra, Shane Legg, and Demis Hassabis.
\newblock Human-level control through deep reinforcement learning.
\newblock {\em Nature}, 518(7540):529–533, February 2015.

\bibitem{JD_etal_22}
Jonas Degrave, Federico Felici, Jonas Buchli, Michael Neunert, Brendan Tracey, Francesco Carpanese, Timo Ewalds, Roland Hafner, Abbas Abdolmaleki, Diego de~las Casas, Craig Donner, Leslie Fritz, Cristian Galperti, Andrea Huber, James Keeling, Maria Tsimpoukelli, Jackie Kay, Antoine Merle, Jean-Marc Moret, Seb Noury, Federico Pesamosca, David Pfau, Olivier Sauter, Cristian Sommariva, Stefano Coda, Basil Duval, Ambrogio Fasoli, Pushmeet Kohli, Koray Kavukcuoglu, Demis Hassabis, and Martin Riedmiller.
\newblock Magnetic control of {T}okamak plasmas through deep reinforcement learning.
\newblock {\em Nature}, 602(7897):414–419, February 2022.

\bibitem{EK_etal_23}
Elia Kaufmann, Leonard Bauersfeld, Antonio Loquercio, Matthias M\"{u}ller, Vladlen Koltun, and Davide Scaramuzza.
\newblock Champion-level drone racing using deep reinforcement learning.
\newblock {\em Nature}, 620(7976):982–987, August 2023.

\bibitem{KC_etal:24}
Katherine~M. Collins, Ilia Sucholutsky, Umang Bhatt, Kartik Chandra, Lionel Wong, Mina Lee, Cedegao~E. Zhang, Tan Zhi-Xuan, Mark Ho, Vikash Mansinghka, Adrian Weller, Joshua~B. Tenenbaum, and Thomas~L. Griffiths.
\newblock Building machines that learn and think with people.
\newblock {\em Nature Human Behaviour}, 8(10):1851–1863, October 2024.

\bibitem{BL_JT_SG:17}
Brenden~M. Lake, Tomer~D. Ullman, Joshua~B. Tenenbaum, and Samuel~J. Gershman.
\newblock Building machines that learn and think like people.
\newblock {\em Behavioral and Brain Sciences}, 40:e253, {November} 201{6}.

\bibitem{VV_etal:22}
Vitaly Vanchurin, Yuri~I. Wolf, Mikhail~I. Katsnelson, and Eugene~V. Koonin.
\newblock Toward a theory of evolution as multilevel learning.
\newblock {\em Proceedings of the National Academy of Sciences}, 119(6), February 2022.

\bibitem{HM_etal:24}
Héctor~Marín Manrique, Karl~John Friston, and Michael~John Walker.
\newblock ‘snakes and ladders’ in paleoanthropology: From cognitive surprise to skillfulness a million years ago.
\newblock {\em Physics of Life Reviews}, 49:40–70, July 2024.

\bibitem{MK_etal_24}
Mayank Kejriwal, Eric Kildebeck, Robert Steininger, and Abhinav Shrivastava.
\newblock Challenges, evaluation and opportunities for open-world learning.
\newblock {\em Nature Machine Intelligence}, 6(6):580–588, June 2024.

\bibitem{RM_PE:25}
{Robert~D. McAllister and Peyman~M. Esfahani.
\newblock Distributionally robust model predictive control: Closed-loop
  guarantees and scalable algorithms.
\newblock {\em IEEE Transactions on Automatic Control}, 70(5):2963--2978, May 2025.}

\bibitem{JM_KH_HA_SS_DC_JP:22}
{Janosch Moos, Kay Hansel, Hany Abdulsamad, Svenja Stark, Debora Clever, and Jan
  Peters.
\newblock Robust reinforcement learning: A review of foundations and recent
  advances.
\newblock {\em Machine Learning and Knowledge Extraction}, 4(1):276–315,
  March 2022.}

 \bibitem{BT_DI_CK_DK:23}
{Bahar Taskesen, Dan Iancu, \c{C}a\u{g}\i~l Ko\c{c}yi\u{g}it, and Daniel Kuhn.
\newblock Distributionally robust linear quadratic control.
\newblock In A.~Oh, T.~Naumann, A.~Globerson, K.~Saenko, M.~Hardt, and
  S.~Levine, editors, {\em Advances in Neural Information Processing Systems}, 36,  18613--18632. Curran Associates, Inc., 2023.}
  
\bibitem{LR_etal_23}
Luc Rocher, Arnaud~J. Tournier, and Yves-Alexandre de~Montjoye.
\newblock Adversarial competition and collusion in algorithmic markets.
\newblock {\em Nature Machine Intelligence}, 5(5):497–504, May 2023.

\bibitem{MW_etal_23}
Maxwell~T. West, Shu-Lok Tsang, Jia~S. Low, Charles~D. Hill, Christopher Leckie, Lloyd C.~L. Hollenberg, Sarah~M. Erfani, and Muhammad Usman.
\newblock Towards quantum enhanced adversarial robustness in machine learning.
\newblock {\em Nature Machine Intelligence}, 5(6):581–589, May 2023.

\bibitem{GH_RZ:93}
Geoffrey~E{.} Hinton and Richard S{.} Zemel.
\newblock Autoencoders, minimum description length and Helmholtz free energy.
\newblock In J.~Cowan, G.~Tesauro, and J.~Alspector, editors, {\em Advances in Neural Information Processing Systems}, volume~6. Morgan-Kaufmann, 1993.

\bibitem{PD_GH:95}
Geoffrey~E{.} Hinton, Peter Dayan, {Radford M. Neal}, and { Richard S.Zemel}.
\newblock The Helmholtz machine.
\newblock {\em Neural Computation}, 7:889–904,  {September} 1995.

\bibitem{SJ_OS:21}
Sharu~T {.} Jose and Osvaldo Simeone.
\newblock Free energy minimization: A unified framework for modeling, inference, learning, and optimization [lecture notes].
\newblock {\em IEEE Signal Processing Magazine}, 38(2):120–125, March 2021.

\bibitem{MHA_EI_RW_RM_JC:21}
Mohamed Hibat-Allah, Estelle~M. Inack, Roeland Wiersema, Roger~G. Melko, and Juan Carrasquilla.
\newblock Variational neural annealing.
\newblock {\em Nature Machine Intelligence}, 3(11):952–961, October 2021.

\bibitem{TP_GP_KF:22}
Thomas Parr, Giovanni Pezzulo, and Karl~J. Friston.
\newblock {\em Active Inference: The Free Energy Principle in Mind, Brain, and Behavior}.
\newblock The MIT Press, March 2022.

\bibitem{PJ_HH_VR:24}
Prateek Jaiswal, Harsha Honnappa, and Vinayak~A. Rao.
\newblock On the statistical consistency of risk-sensitive {B}ayesian decision-making.
\newblock In {\em Proceedings of the 37th International Conference on Neural Information Processing Systems}, NIPS '23, Red Hook, NY, USA, 2024. Curran Associates Inc.

\bibitem{TS:14}
Terence~D. Sanger.
\newblock Risk-aware control.
\newblock {\em Neural Computation}, 26(12):2669–2691, December 2014.

\bibitem{TB_AP_TM:24}
{Thomas~A. Berrueta, Allison Pinosky, and Todd~D. Murphey.
\newblock Maximum diffusion reinforcement learning.
\newblock {\em Nature Machine Intelligence}, 6(5):504–514, May 2024.}

\bibitem{MR_etal:21}
Pietro Mazzaglia, Tim Verbelen, and Bart Dhoedt.
\newblock Contrastive active inference.
\newblock In M.~Ranzato, A.~Beygelzimer, Y.~Dauphin, P.S. Liang, and J.~Wortman Vaughan, editors, {\em Advances in Neural Information Processing Systems}, volume~34, pages 13870--13882. Curran Associates, Inc., 2021.

\bibitem{KF:09}
Karl Friston.
\newblock The free-energy principle: a rough guide to the brain?
\newblock {\em Trends in Cognitive Sciences}, 13(7):293–301, July 2009.

\bibitem{CH_etal:24}
Conor Heins, Beren Millidge, Lancelot Da~Costa, Richard~P. Mann, Karl~J. Friston, and Iain~D. Couzin.
\newblock Collective behavior from surprise minimization.
\newblock {\em Proceedings of the National Academy of Sciences}, 121(17), April 2024.

\bibitem{TP_SW:23}
Tony~J. Prescott and Stuart~P. Wilson.
\newblock Understanding brain functional architecture through robotics.
\newblock {\em Science Robotics}, 8(78), May 2023.

\bibitem{JH:13}
Jakob Hohwy.
\newblock {\em The Predictive Mind}.
\newblock Oxford University Press, November 2013.

\bibitem{KF_LDC_NS_CH_KU_GP_AG_TP:23}
Karl Friston, Lancelot D{.} Costa, Noor Sajid, Conor Heins, Kai Ueltzh\"{o}ffer, Grigorios~A. Pavliotis, and Thomas Parr.
\newblock The free energy principle made simpler but not too simple.
\newblock {\em Physics Reports}, 1024:1–29, June 2023.

\bibitem{SG_DB:20}
Sebastian Gottwald and Daniel~A. Braun.
\newblock The two kinds of free energy and the {B}ayesian revolution.
\newblock {\em PLOS Computational Biology}, 16(12):e1008420, December 2020.

\bibitem{AI_JW_JP:20}
Abraham Imohiosen, Joe Watson, and Jan Peters.
\newblock {\em Active Inference or Control as Inference? A Unifying View}, page 12–19.
\newblock Springer International Publishing, {September} 2020.

\bibitem{TB:24}
{Thomas Berrueta.
\newblock {\em Robot Thermodynamics}.
\newblock Ph.d. dissertation, Nortwestern University, December 2024.
\newblock Available online at
  \url{https://www.proquest.com/openview/faffd739b9b7a1becbd5e99b0fbd83fe/}.}
  
\bibitem{BE_SL:22}
{Benjamin Eysenbach and Sergey Levine.
\newblock Maximum entropy {RL} (provably) solves some robust {RL} problems.
\newblock In {\em International Conference on Learning Representations}, 2022.}

 {\bibitem{ET_TE_YT:12}
{Emanuel Todorov, Tom Erez and Yuval Tassa. 
\newblock {MuJoCo}: A physics engine for model-based control. 
\newblock In {\em Proceedings of 2012 IEEE/RSJ International Conference on Intelligent Robots and Systems}, pp.  5026--5033, 2012}}

 {\bibitem{GW_NW_BG_PD_JR_BB_ET:17}
{Grady Williams, Nolan Wagener, Brian Goldfain,  Paul Drews,  James Rehg,  Byron Boots,  Evangelos Theodorou. 
\newblock Information theoretic {MPC} for model-based reinforcement learning. 
\newblock In {\em Proceedings of 2017 IEEE International Conference on Robotics and Automation}, pp.  1714--1721, 2017}}

\bibitem{KF_FR_DO_CM_TF_GP:15}
Karl Friston, Francesco Rigoli, Dimitri Ognibene, Christoph Mathys, Thomas Fitzgerald, and Giovanni Pezzulo.
\newblock Active inference and epistemic value.
\newblock {\em Cognitive Neuroscience}, 6(4):187--214,  {March} 2015.

\bibitem{TP_KF:17}
Thomas Parr and Karl~J. Friston.
\newblock Uncertainty, epistemics and active inference.
\newblock {\em Journal of The Royal Society Interface}, 14(136):20170376,  {November} 2017.

\bibitem{YK_HN:23}
Yuki Konaka and Honda Naoki.
\newblock Decoding reward–curiosity conflict in decision-making from irrational behaviors.
\newblock {\em Nature Computational Science}, 3(5):418–432, May 2023.

\bibitem{KK_SP_SM_RP_MS_JCD_RR:19}
Koosha Khalvati, Seongmin~A. Park, Saghar Mirbagheri, Remi Philippe, Mariateresa Sestito, Jean-Claude Dreher, and Rajesh P.~N. Rao.
\newblock Modeling other minds: Bayesian inference explains human choices in group decision-making.
\newblock {\em Science Advances}, 5(11), November 2019.

\bibitem{AM_PL_GP:22}
Antonella Maselli, Pablo Lanillos, and Giovanni Pezzulo.
\newblock Active inference unifies intentional and conflict-resolution imperatives of motor control.
\newblock {\em PLOS Computational Biology}, 18(6):e1010095, June 2022.

\bibitem{JV_etal:06}
Julian~ F {.}V {.} Vincent, Olga~A {.} Bogatyreva, Nikolaj~R {.} Bogatyrev, Adrian Bowyer, and Anja K{.} Pahl.
\newblock Biomimetics: its practice and theory.
\newblock {\em Journal of the Royal Society Interface}, 3(9):471--482,  {April} 2006.

\bibitem{GP_FR_KF:15}
Giovanni Pezzulo, Francesco Rigoli, and Karl Friston.
\newblock Active inference, homeostatic regulation and adaptive behavioural control.
\newblock {\em Progress in neurobiology}, 134:17--35, {November} 2015.

\bibitem{TP_KF:19}
Thomas Parr and Karl~J. Friston.
\newblock Generalised free energy and active inference.
\newblock {\em Biological Cybernetics}, 113(5–6):495–513, September 2019.

\bibitem{JB_WW_HK:10}
Bart V{.}D{.} Broek, Wim Wiegerinck, and Hilbert~J. Kappen.
\newblock Risk sensitive path integral control.
\newblock In {\em Conference on Uncertainty in Artificial Intelligence}, 2010.

\bibitem{AH:03}
Hagai Attias.
\newblock Planning by probabilistic inference.
\newblock In Christopher~M. Bishop and Brendan~J. Frey, editors, {\em Proceedings of the Ninth International Workshop on Artificial Intelligence and Statistics}, volume~R4 of {\em Proceedings of Machine Learning Research},  9--16.
\newblock Reissued by PMLR on 01 April 2021.

\bibitem{MB_MT:12}
Matthew Botvinick and Marc Toussaint.
\newblock Planning as inference.
\newblock {\em Trends in Cognitive Sciences}, 16(10):485–488, October 2012.

\bibitem{EG_HJ_CDV_GR:24}
Emiland Garrabe, Hozefa Jesawada, Carmen~Del Vecchio, and Giovanni Russo.
\newblock On convex data-driven inverse optimal control for nonlinear, non-stationary and stochastic systems.
\newblock {\em Automatica}, 173:112015, March 2025.

\bibitem{SW_etal:20}
Sean Wilson, Paul Glotfelter, Li~Wang, Siddharth Mayya, Gennaro Notomista, Mark Mote, and Magnus Egerstedt.
\newblock {The Robotarium: Globally Impactful Opportunities, Challenges, and Lessons Learned in Remote-Access, Distributed Control of Multirobot Systems}.
\newblock {\em IEEE Control Systems Magazine}, 40(1):26--44,  {February} 2020.

\bibitem{BZ_AM_AB_AD:08}
Brian~D. Ziebart, Andrew Maas, J.~Andrew Bagnell, and Anind~K. Dey.
\newblock {Maximum Entropy Inverse Reinforcement Learning}.
\newblock In {\em Proceedings of the 23rd National Conference on Artificial Intelligence - Volume 3}, AAAI'08, page 1433–1438. AAAI Press, 2008.

\bibitem{ZZ_etal:18}
Zhengyuan Zhou, Michael Bloem, and Nicholas Bambos.
\newblock {Infinite Time Horizon Maximum Causal Entropy Inverse Reinforcement Learning}.
\newblock {\em IEEE Transactions on Automatic Control}, 63(9):2787--2802,  {September} 2018.

\bibitem{ZA_etal:20}
Zahra Alirezaeizanjani, Robert Grossmann, Veronika Pfeifer, Marius Hintsche, and Carsten Beta.
\newblock Chemotaxis strategies of bacteria with multiple run modes.
\newblock {\em Science Advances}, 6(22), May 2020.

\bibitem{AT_AS_CB:20}
Alexander Tschantz, Anil~K. Seth, and Christopher~L. Buckley.
\newblock Learning action-oriented models through active inference.
\newblock {\em PLOS Computational Biology}, 16(4):e1007805, April 2020.

\bibitem{NT:12}
{Nassim N. Taleb.  
\newblock Antifragile: Things that gain from disorder. 
\newblock Random House,  November 2012.}

\bibitem{Barto2013}
{Andrew Barto, Marco Mirolli, and Gianluca Baldassarre.
\newblock Novelty or surprise?
\newblock {\em Frontiers in Psychology}, 4:907,  December 2013.}

\bibitem{Hsu2005}
{Ming Hsu, Meghana Bhatt, Ralph Adolphs, Daniel Tranel, and Colin F. Camerer.
\newblock Neural systems responding to degrees of uncertainty in human decision-making.
\newblock {\em Science}, 310:1680--1683,  December 2005.}

\bibitem{Zak2004}
{Paul J. Zak.
\newblock Neuroeconomics.
\newblock {\em Philosophical Transactions of the Royal Society B: Biological Sciences}, 359:1737--1748,  November 2004.}

\bibitem{GARRABE202281}
{Émiland Garrabé and Giovanni Russo.
\newblock Probabilistic design of optimal sequential decision-making algorithms in learning and control.
\newblock {\em Annual Reviews in Control}, 54:81--102,  November 2022.}

\bibitem{EJ:80}
Edwin~T {.} Jaynes.
\newblock The minimum entropy production principle.
\newblock {\em Annual Review of Physical Chemistry}, 31(1):579–601, October 1980.

\bibitem{HH_JP:21}
Hermann Haken and Juval Portugali.
\newblock {\em Relationships. Bayes, Friston, Jaynes and Synergetics 2nd Foundation},   85 -- 104.
\newblock Springer International Publishing,  {February} 2021.

\bibitem{ET:09}
Emanuel Todorov.
\newblock Efficient computation of optimal actions.
\newblock {\em Proceedings of the National Academy of Sciences}, 106(28):11478--11483, {April} 2009.

\bibitem{KM:23}
Kevin~P. Murphy.
\newblock {\em Probabilistic Machine Learning: Advanced Topics}.
\newblock MIT Press, 2023.

\bibitem{IG_JB_AC:16}
Ian Goodfellow, Yoshua Bengio, and Aaron Courville.
\newblock {\em Deep Learning}.
\newblock MIT Press, 2016.

\bibitem{KD_ET:10}
Krishnamurthy Dvijotham and Emanuel Todorov.
\newblock Inverse optimal control with {L}inearly-{S}olvable {MDPs}.
\newblock In {\em 27th International Conference on Machine Learning},  335–342, 2010.

\bibitem{repo}
Hozefa Jesawada, 
\textit{\AlgoDM~repository}, 
Zenodo, \url{https://doi.org/10.5281/zenodo.17638771}, 
2025

\end{thebibliography}
\end{document}


\onehalfspacing
\title{{\bf Supplementary Information:  
Distributionally Robust Free Energy Principle for Decision-Making}}

\author{Allahkaram Shafiei~\textsuperscript{1,$\ast$} \and Hozefa Jesawada~\textsuperscript{{2,$\ast$}} \and Karl Friston~\textsuperscript{{3}} \and Giovanni Russo~\textsuperscript{4} \Letter}

\date{}
\maketitle~\blfootnote{\textsuperscript{1} Czech Technical University, Prague, Czech Republic. {\textsuperscript{2} New York University Abu Dhabi,  Abu Dhabi Emirate. } \textsuperscript{{3}} Wellcome Centre for Human Neuroimaging, Institute of Neurology, University College London,  United Kingdom.  \textsuperscript{4} Department of Information and Electrical Engineering and~Applied Mathematics, University of Salerno,  Italy. 

\textsuperscript{$\ast$} These authors contributed equally.   \Letter~e-mail: \href{mailto:giovarusso@unisa.it}{giovarusso@unisa.it}}

\renewcommand{\thefigure}{SI-\arabic{figure}}
\setcounter{figure}{0} %

\section*{Supplementary Information}
\startSI

We provide supplementary figures and the formal details for the statements and results in the main text.  After providing some background (Sec. \ref{sec:background}) we {relate \AlgoDM~with other frameworks (Sec. \ref{sec:related_works}) and} give the formal statements behind the resolution engine in Fig.  2b (Sec. \ref{sec:resolution_engine}). In Sec. \ref{sec:ambiguity_to_zero} we show why \AlgoDM~shows that it is always better to be ambiguity aware. After reporting the supplementary details of the experiments (Sec. \ref{sec:experiments_details}) we provide the proofs of all the statements (Sec. \ref{sec:proofs}). {Finally, in Sec. \ref{sec:ambiguity},  we unpack some possible psychological and translational implications of \AlgoDM.}\newline

\newpage

\section*{Supplementary Figures}
\thispagestyle{empty}
\begin{figure*}[!h]
	\centering
\noindent\includegraphics[width=\columnwidth]{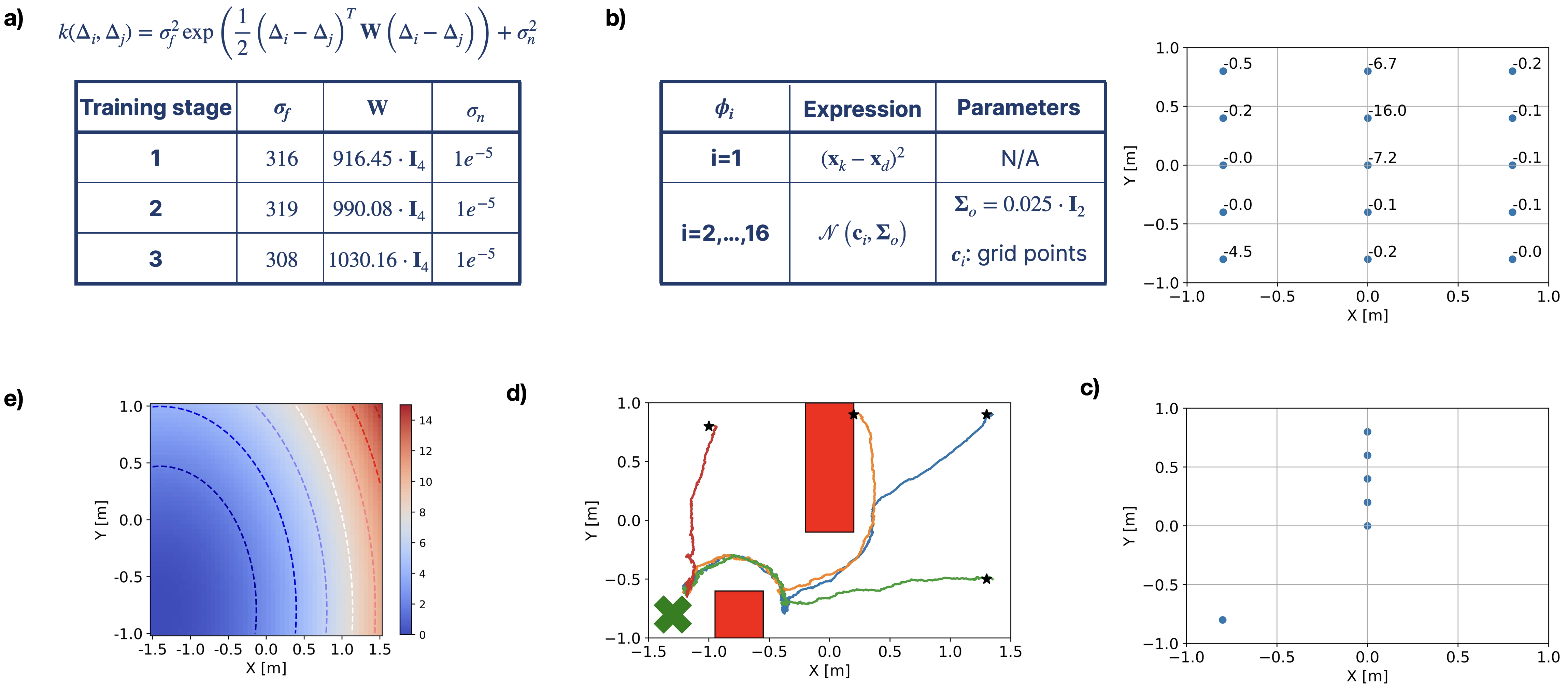}  
\captionsetup{labelformat=empty}
\caption{{\bf Supplementary Fig.  1. a.} The  kernel function for the Gaussian Process learning (from Methods) of the trained model together with the parameters learned at each stage of learning. {\bf b.} The features for cost reconstruction used to obtain Figure 3f. {\bf c.} Centers of the Gaussians promoting obstacles avoidance in Fig. 3 experiments (the parameter $\bv{o}_i$ in the agent cost given in Methods). {\bf d.} Recordings from the Robotarium simulator of experiments where the robots are equipped with \AlgoDM~policy but using the reconstructed cost. The initial positions are different from the ones used in the experiments in the main paper. {\bf e.} Cost reconstructed with related method from the literature\cite{ZZ_etal:18}.} 
\end{figure*}

\newpage
\thispagestyle{empty}
\begin{figure*}[!h]
	\centering
\noindent\includegraphics[width=\columnwidth]{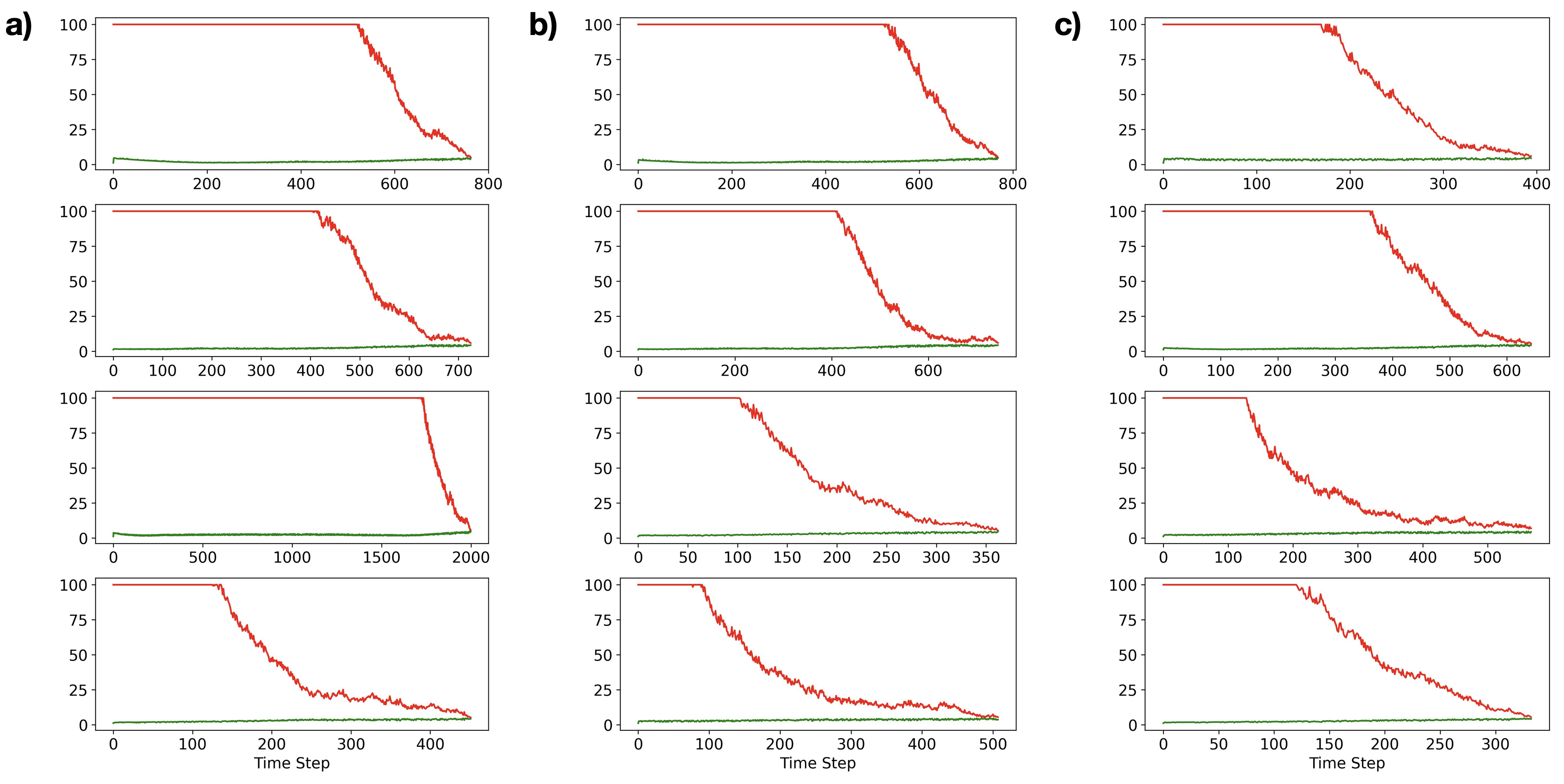}  
\captionsetup{labelformat=empty}
\caption{{\bf Supplementary Fig.  2.} In all the experiments from Fig.  3c, $\plant{k}{k-1}$ always belongs to the ambiguity set $\ball{k}{k-1}$. The plots confirm this by showing that $\radius{k}{k-1}$ {-- see {Experiments settings} in Methods for the radius used in the experiments -- } is always bigger than $\DKL{\plant{k}{k-1}}{\nominalplant{k}{k-1}}$ across all the experiments.  Column {\bf a.} is from the experiments in Fig.  3c top-left.  Column {\bf b.} is from the experiments in Fig.  3c top-middle.  Column {\bf c.} is from the experiments in Fig.  3c top-right. In all plots, $\radius{k}{k-1}$ is in red and $\DKL{\plant{k}{k-1}}{\nominalplant{k}{k-1}}$ in green. Time series stop when the episode ends {($x$-axes show the number of steps for each experiment)}.  } 
\end{figure*}

\newpage
\thispagestyle{empty}
\begin{figure*}[!h]
	\centering
\noindent\includegraphics[width=\columnwidth]{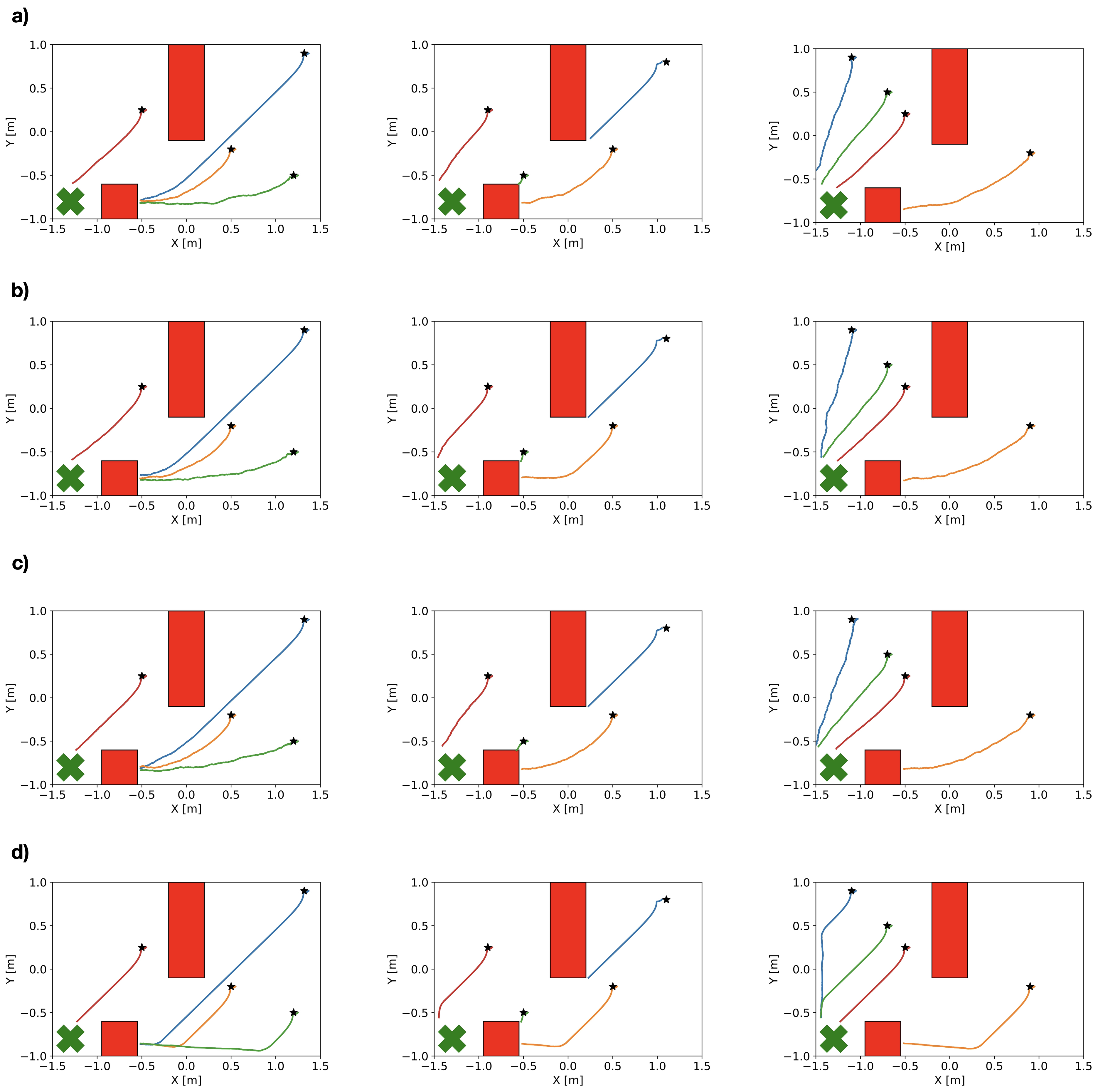}  
\captionsetup{labelformat=empty}
\caption{{\bf {Supplementary Fig.  3.}}  {Behavior of the ambiguity-unaware agent when equipped with planning.  Panel {\bf a.} shows the behavior of the agent when 
$N =3$.  Panels {\bf b--d.} show the behavior of the agent when $N =5$, $N=10$ and $N=50$, respectively.   In all panels the initial positions for the robot are the same as in Fig.  3c.}} 
\end{figure*}

\newpage
\thispagestyle{empty}
\begin{figure*}[!h]
	\centering
\noindent\includegraphics[width=\columnwidth]{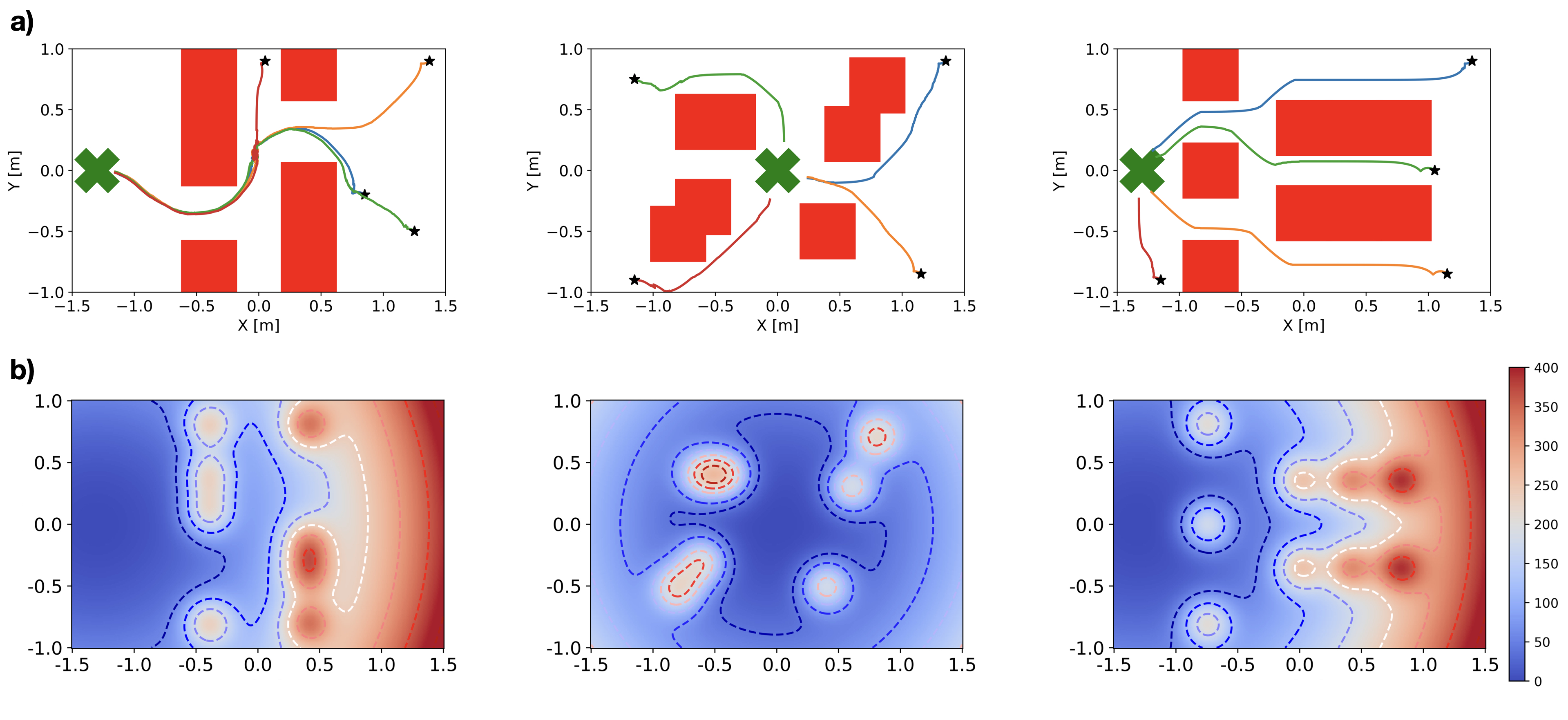}  
\captionsetup{labelformat=empty}
\caption{ {{\bf Supplementary Fig.  4.}   \AlgoDM~experiments for the robot navigation task in different environments.  {\bf a.} Experimental results for different obstacles and goal configurations.  Across all the experiments, \AlgoDM~enables the robot to successfully complete the task,  consistently with our main results.  {\bf b.} The corresponding costs (scale on the right).  The cost  function is the same to the one described in Methods, with updated coordinates for obstacles and  goal positions to match the different experimental set-ups.  }}
\end{figure*}

\newpage
\thispagestyle{empty}
\begin{figure*}[!h]
	\centering
\noindent\includegraphics[width=\columnwidth]{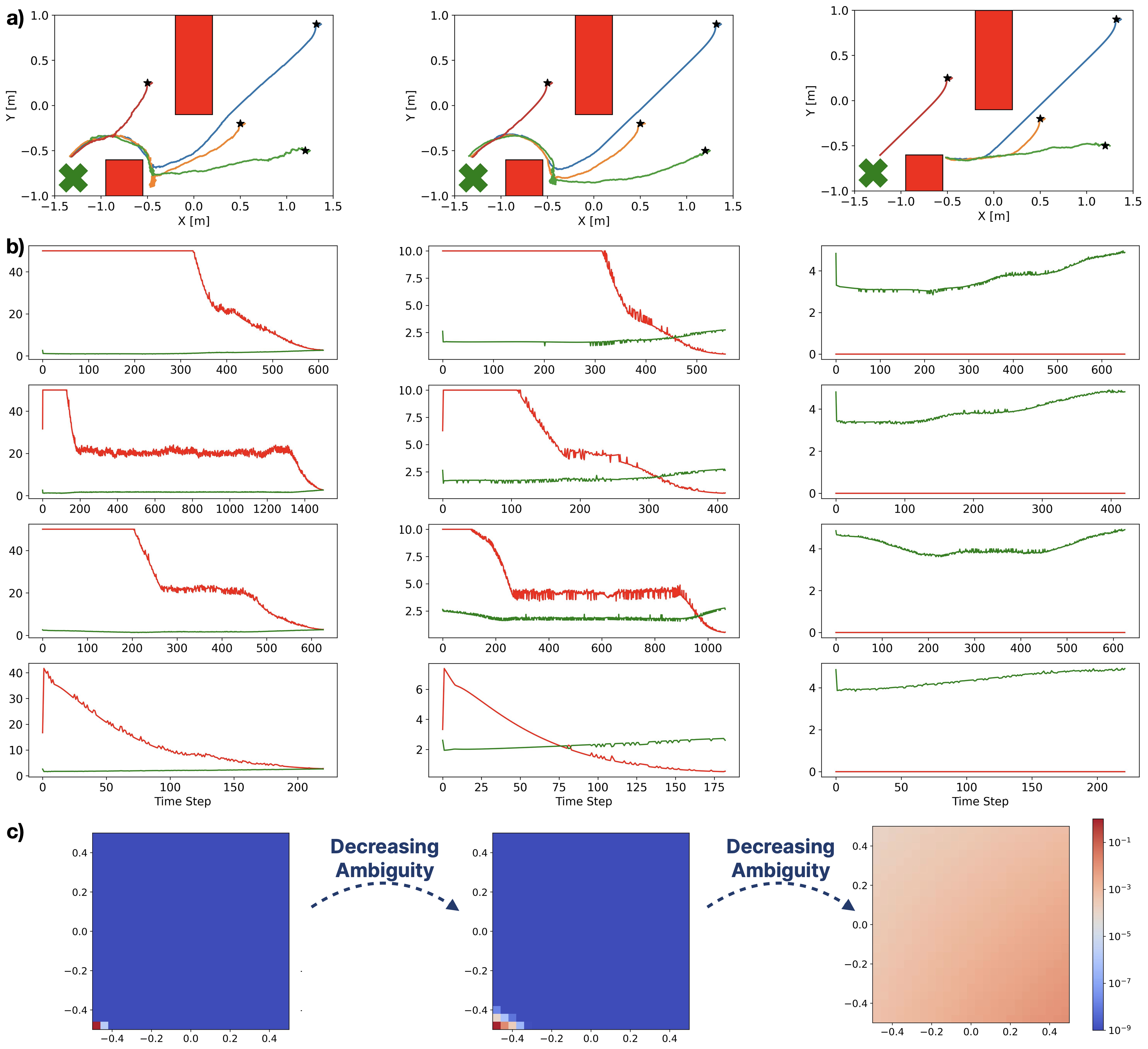}  
\captionsetup{labelformat=empty}
\caption{ {{\bf Supplementary Fig.  5.}   Experiments results for different ambiguity levels.  {\bf a.} With respect to the settings used in Fig.  3c, here the ambiguity radii are:  (i) decreased by $50\%$ (left); (ii) decreased by $90\%$ (middle); set to $0$ (right).  In this last case,  $\plant{k}{k-1}$ never belongs to the ambiguity constraint.  Consequently,  consistently with our analytical results,  the agent can  fail  in the presence of ambiguity.  {\bf b.} Following the format of Supplementary Fig.  2, each column shows -- for each of the experiments --  when $\plant{k}{k-1}$ belongs to the ambiguity set $\ball{k}{k-1}$.  Color coding is as in Expected Data 2.  This panel (middle column) reveals that \AlgoDM~can still complete the task even when  $\plant{k}{k-1}$ does not always lie within the ambiguity set.   {\bf c.} \AlgoDM~policy for different ambiguity levels.  The heatmaps -- computed for the same ambiguity levels as in the top panels -- show how the robot policy changes when this is in position $[-0.5, -0.5]$.  When the radius is set to $0$,  the ambiguity constraint is relaxed and \AlgoDM~policy coincides with the optimal policy for this relaxed problem.  The process is representative for all robot trajectory points.  See Sec. \ref{sec:experiments_details} in \SI. }}
\end{figure*}

\newpage
\thispagestyle{empty}
\begin{figure*}[!h]
	\centering
\noindent\includegraphics[width=\columnwidth]{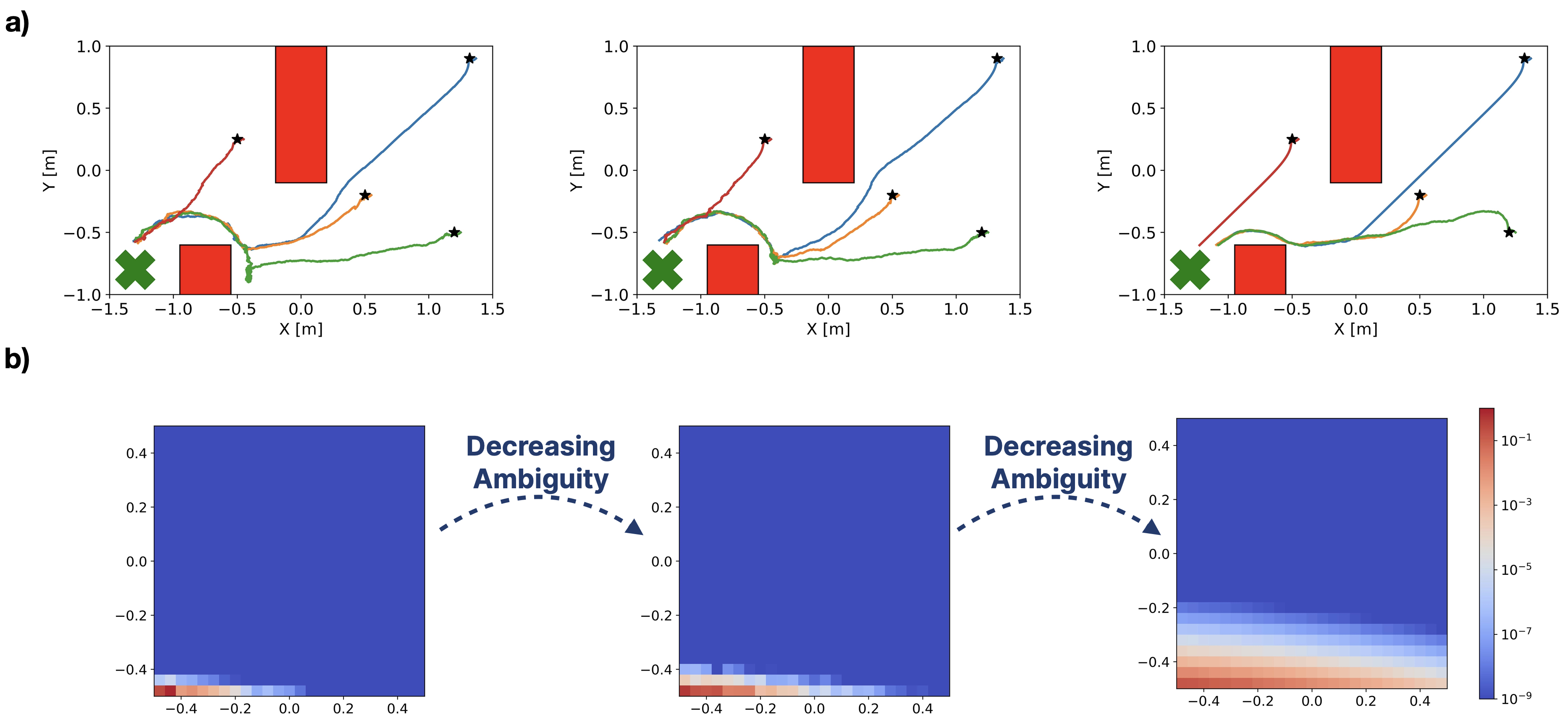}  
\captionsetup{labelformat=empty}
\caption{ {{\bf Supplementary Fig.  6.} Experiments results when there is no ambiguity.  {\bf a.}  The trained model $\nominalplant{k}{k-1}$ coincides with $\plant{k}{k-1}$.  Experiments use the same ambiguity radii as in Supplementary Fig.  5.  This time,  since the scenario does not  feature ambiguity,  \AlgoDM~enables the robot to  fulfill the task in all cases, even when the ambiguity radius is set to $0$ (right).  {\bf b.} As in Supplementary Fig.  5 and consistently with our analysis,  when the radius is set to $0$,  the ambiguity constraint is relaxed and \AlgoDM~policy again coincides with the optimal policy for this relaxed problem.   As in Supplementary Fig.  5, the  heatmap corresponds to the robot policy when this is in position $[-0.5, -0.5]$. See also Fig. \ref{fig:SI-ambiguity}. }}
\end{figure*}

\newpage
\thispagestyle{empty}
\begin{figure*}[!h]
	\centering
\noindent\includegraphics[width=\columnwidth]{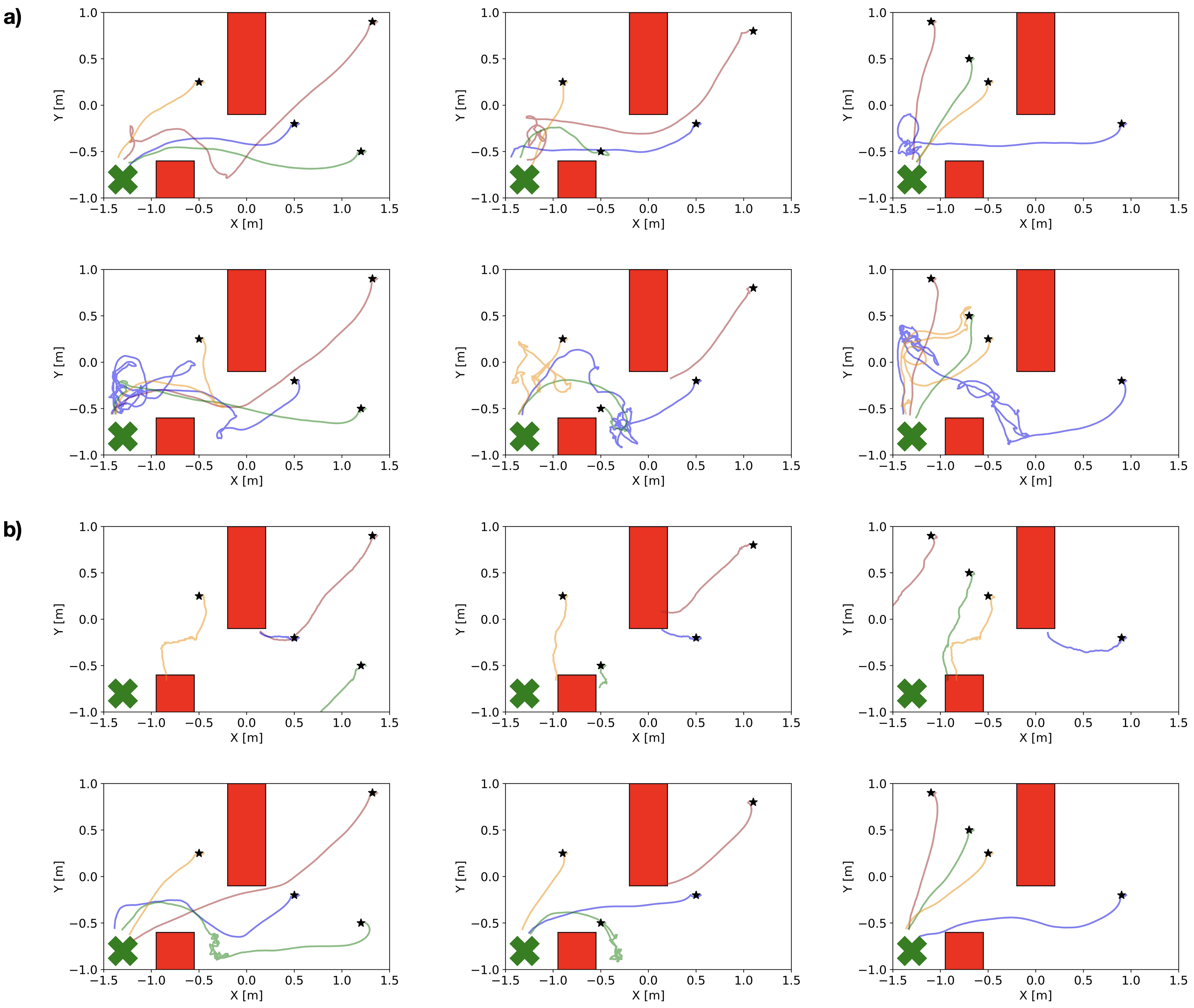}  
\captionsetup{labelformat=empty}
\caption{{{{\bf Supplementary Fig.   {7}.}  Additional MaxDiff experiments.}}  {Panel {\bf a.} shows the experiments with MaxDiff having access to both reward and model. The top row shows the experiments with hyperparameters taken from the sweetspot in Fig.  4a where MaxDiff achieves $100\%$ success rate --  horizon set to $20$ and samples to $50$.  The agent is in fact able to always complete the task.  However, near the goal position,  in some of the experiments, we observe an erratic behavior.  Upon experimenting with the environment,  we believe that this behavior might be due to planning under a model affected by ambiguity.  This phenomenon is more apparent in the bottom row.  Here,  the horizon set to $50$ and samples to $100$.  In panel {\bf b.}  MaxDiff does not have access to the reward/model and the policy is learned -- these experiments are developed to check consistency of success rates.  The learning rate is set to $0.0005$ and $100000$ data points are used to learn the policy, consistently with the MaxDiff code for the {point mass} example.  Data are corrupted by the same bias used in \AlgoDM~experiments. The top row shows the behavior of the agent using the learned policy when horizon is $2$ and samples $50$ (as in Fig.  4c).  The robot behavior is consistent with the one in Fig.  4c. The bottom row has horizon set to $20$ and samples to $50$. The learned policy achieves results that are consistent  with the top row (panel a).  In all panels, initial positions are the same as the ones in the main text.  See Methods (at the end of {Experiments settings}) and \SI~(Sec. \ref{sec:experiments_details} -- {MaxDiff settings}) for the  details.}} 
\end{figure*}

\section{Background}\label{sec:background}

Sets and operators are in {calligraphic} characters and vectors in {bold}. A random variable is denoted by $\bv{V}$ and its realization is $\bv{v}$.  The symbol $\coloneq$  denotes a definition.  We consider continuous (discrete) random variables and we denote the \textit{probability density function} (pdf) of $\mathbf{V}$  by $p(\mathbf{v})$ (for discrete variables, $p(\mathbf{v})$ is \textit{probability mass function}, pmf).  The convex subset of pdfs (pmfs) is denoted by $\sD$. The  expectation of a function $\mathbf{h}(\cdot)$ of a continuous $\mathbf{V}$ is denoted $\E_{{p}}[\mathbf{h}(\mathbf{V})]\coloneq\int_{\bv{v}}\mathbf{h}(\mathbf{v})p(\mathbf{v})d\bv{v}$, where the integral is over the (compact) support of $p(\mathbf{v})$,  which we denote by $\support p$; whenever it is clear from the context, we omit the subscript in the sum (for discrete variables, the integral is replaced with the sum). The joint pdf/pmf of $\mathbf{V}_1$ and $\mathbf{V}_2$ is denoted by  $p(\mathbf{v}_1,\mathbf{v}_2)$; the conditional pdf/pmf of $\mathbf{V}_1$ with respect to (w.r.t.) $\mathbf{V}_2$ is $p\left( \mathbf{v}_1\mid   \mathbf{v}_2 \right)$.  Given $p(\bv{v})$ and $q(\bv{v})$, we say that $p(\bv{v})$ is absolutely continuous with respect to (w.r.t.) $q(\bv{v})$ if $\support p \subseteq \support q$.  We denote this by writing  $p \ll q$. Countable sets are denoted by $\lbrace w_k \rbrace_{k_1:k_n}$, where $w_k$ is the generic set element, $k_1$ ($k_n$) is the index of the first (last) element and  $k_1:k_n$ is the set of consecutive integers between (including) $k_1$ and $k_n$.  Finally, 
we denote by $L^1(\sV)$ the space integrable functions on $\sV$ and we use the shorthand notation {a.s.} for {almost surely}.\\

\noindent {\bf The Kullback-Leibler Divergence.}
We recall the definition for the Kullback-Leibler (KL) divergence \cite{SK_RL:51}
\begin{definition}
The KL divergence of $p(\mathbf{v})$ w.r.t. $q(\mathbf{v})$ with $p\ll q$ is:
$$
\DKL{p}{q}:= \int_{\bv{v}} p(\bv{v}) \ln\left( \frac{p(\bv{v})}{q(\bv{v})}\right)d\bv{v}.
$$ 
\end{definition}
Intuitively, the KL divergence is a measure of the proximity of the pair of pdfs.  This is bounded only if $p\ll q$ \cite[Chapter $8$]{TC_JT:06}.  Also, $(p,q)\mapsto \DKL{p}{q}$  is a jointly convex function and hence  $p\mapsto \DKL{p}{q}$, $q\mapsto \DKL{p}{q}$ are convex. 
We also recall the following {chain rule} for the KL divergence: 
\begin{lemma}\label{lem:splitting_property}	
Let $\bv{V}$ and $\bv{Z}$ be two random variables  and let $p(\bv{v},\bv{z})$ and $q(\bv{v},\bv{z})$ be two joint pdfs. Then:
\begin{align*}
\DKL{p(\bv{v},\bv{z})}{q(\bv{v},\bv{z})} =  \DKL{p(\bv{v})}{q(\bv{v})} + \E_{p(\bv{v})}
\left[	
\DKL{p(\bv{z}\mid\bv{v})}{q(\bv{z}\mid\bv{v})}\right].
\end{align*}
\end{lemma}
Finally,  the following result is useful to characterize the feasibility domain of our control problem. In the statement,  adapted from \cite[Proposition 2.1]{pinski2015kullback}, $\mathcal{M}(\sV)$ is the subset of probability measures on $\sV \subseteq \R^k$
\begin{lemma}\label{lem:compactness} 
For any $\mu\in \mathcal{M}(\sV)$, $\sV\subseteq\R^k$ and any $M<\infty$, the set
\[\{\nu\in \mathcal{M}(\sV):~~~\DKL{\nu}{\mu}\le M\},\]
is compact with respect to weak convergence of probability  measures.
\end{lemma}

{\section{Relating \AlgoDM~with other frameworks}\label{sec:related_works}
We recall that \AlgoDM~computes the policy via the distributionally robust generalization of the free energy principle in Fig.  2a. This formulation -- offering the problem statement for policy computation -- explicitly embeds ambiguity constraints to account for (model) ambiguity.  The formulation yields a distributionally robust sequential policy optimization problem featuring the free energy as objective and ambiguity constraints formalized via the KL divergence.  Both the objective and constraints are nonlinear in the decision variables of the problem. The constraints explicitly define an ambiguity set with radius that can depend on both state and action.  In the main text we showed that the policy optimization problem in Fig. 2a tackled by \AlgoDM~has the same optimal solution as
\begin{equation}\label{eqn:DRFREE-SI}
\underset{\{\policy{k}{k-1}\}_{1:N}}{\min}\ \ \underset{\plant{k}{k-1}\in\ball{k}{k-1}}{\max}
\DKL{p_{0:N}}{\tilde q_{0:N}}
\end{equation}
We then showed that -- with a proper choice of  $\tilde q_{0:N}$ -- the MaxDiff objective is a relaxation of the above problem when there is no ambiguity.  Starting from this finding -- and building on the links highlighted in the main text and in the Methods -- we now connect \AlgoDM~with other related robust decision-making frameworks, placing it in the context of the broad literature on maximum entropy, variational inference, control-as-inference, distributionally robust learning and optimization.}

{As highlighted in the main text,  MaxDiff can generalize\cite{TB_AP_TM:24,TB:24} maximum entropy\cite{BZ_AB_AD:10,TH_HT_PA_SL:17} (MaxEnt) inheriting the desirable properties of MaxEnt policies.   A remarkable property is policy robustness to perturbations in the dynamics\cite{BE_SL:22}.  Formally,  MaxEnt policies maximize a lower bound of the distributionally robust objective\cite{BE_SL:22}
\begin{equation}\label{eqn:MaxEnt_Robust}
\underset{\{\policy{k}{k-1}\}_{1:N}}{\max}\ \ \underset{\plant{k}{k-1}\in\tilde{\mathcal{P}}}{\min} \E_{p_{0:N}}\left[\sum_{k=1}^N\tilde{r}(\bv{X}_{k},\bv{U}_k)\right]
\end{equation}
In~\eqref{eqn:MaxEnt_Robust},  $\tilde{r}$ is a pessimistic reward\cite{BE_SL:22},  this means that to compute policies that robustly maximize a reward  MaxtEnt must be used with a different (pessimistic) reward function.  Also,  in \eqref{eqn:MaxEnt_Robust} the ambiguity set $\tilde{\mathcal{P}}$ is not arbitrary; rather, its radius\cite{BE_SL:22} is constant -- it directly depends upon the entropy of the optimal policy, which is the solution of  MaxEnt itself and explicit bounds are available only in discrete settings.  The bounds come from the fact that, to obtain robustness estimates, the authors\cite{BE_SL:22}  do not solve the full problem in \eqref{eqn:MaxEnt_Robust} but rather obtain a lower bound for the inner minimization.  \AlgoDM~solves the problem in Fig.  2a -- and hence in \eqref{eqn:DRFREE-SI}. In doing so,  \AlgoDM~defines robustness against ambiguity in the problem formulation, via the ambiguity set.  This means not only that  \AlgoDM~policy is robust across this ambiguity set, but also that \AlgoDM~does not need to be applied on a pessimistic policy.  The key enabler to achieve these desirable properties is the ability to solve the robust free energy formulation in Fig.  2a.  As we detail next,  we are not aware of any other approach to solve this problem in the broad literature on robust decision-making. Namely,  our suvey highlights that, in this broader landscape,  there is no method that can solve the full $\min-\max$ problem in Fig.  2a. We take as a starting point MaxEnt -- which we discussed in the main text. While distributionally robust versions of imitation RL based on this principle have been proposed, results are obtained\cite{MB_BZ_XZ:21} by assuming that -- besides finite state spaces -- the underlying optimization features a cost functional and  constraints that are both linear in the decision variable of the inner optimization problem.  This  does not hold for the functional and constraints of our robust free energy principle.  More broadly,  as also surveyed in the literature\cite{HP_DZ_GH_TT:24,BVP_DK_PG_MM:16,SW_NS_JB_ZZ:24,CC_FR:07,RM_PE:25,JM_KH_HA_SS_DC_JP:22,BT_DI_CK_DK:23} the linearity assumption  on the objective function in one of the decision variables is crucial for other state-of-the-art policy computation methods across learning and control. This includes highly influential distributionally robust frameworks such as  distributionally robust Q-learning\cite{ZL_etal:22}. In this work, while ambiguity constraints are formalized via KL divergence -- as in \AlgoDM~-- the optimization objective benefits from being linear in the decision variable of the inner problem.   As a result, distributionally robust Q-learning cannot compute a policy that solves the full problem in Fig.  2a.  KL divergence is also widely used as a cost functional in the context of variational inference (and control-as-inference).  However, also in this literature,  when distributionally robust frameworks are proposed,  in line with the distributionally robust Q-learning setting, the KL divergence (and the entropy) is included in the problem in a way that guarantees linearity of the cost with respect to the inner decision variable\cite{BW_BZ_DB:25}.  Finally,  we also position~\AlgoDM~-- and its contributions -- in the context of the rich literature on distributionally robust optimization (DRO). Since Scarf's pioneering work\cite{HS:57}, DRO (and Bayesian DRO) has emerged as a central topic in robust optimization  -- gaining increasing attention in the context of machine learning\cite{JK_IB_SJ_AK:20,AS_EZ_YL:23,MRS_PEM:24,JD_PG_HN:21,BVP_PE_DK:21,ZH_JH:13}.  DRO concepts have developed significantly over the years and we refer to surveys\cite{FL_XF_ZG:22,HR_SM:22} for an overview of this vast literature.  In these works -- and surveys -- the optimization problem benefits from the cost being linear in at least one decision variable -- thus ruling out having free energy as objective. More recently\cite{MRS_PEM:24}, nonlinear DRO problems are considered and a Frank-Wolfe algorithm has been introduced to tackle a class of problems featuring cost functions that have a continuous Lipschitz G-derivative. This assumption does not hold for the free energy objective of the distributionally robust free energy principle in Fig.  2a. }

{In summary,  \AlgoDM~allows to solve distributionally robust, $\min-\max$, policy optimization problems having free energy as cost functional and, at the same time, distributionally robust constraints defined via the KL divergence.  This contribution -- interesting {per se} -- does not  just bring broad theoretical implications.  As discussed in the main text,  it is this key advancement that enables \AlgoDM~to retain the desirable theoretical properties of state-of-the-art -- provably robust -- methods such as MaxEnt and MaxDiff, guaranteeing these properties over the  ambiguity set defined in the problem formulation. }
 {\subsection{\AlgoDM~and the MDP formalism}
Following the literature on distributionally robust Markov Decision Processes\cite{HX_SM:10,ZL_etal:22},  we complement the above survey by relating the sequential policy computation problem tackled in \AlgoDM~(Fig.  2a) and the MDP formalism.  MDPs can be typically defined according to a $6$-tuple\cite{HX_SM:10} $(T, \gamma,\sX,\sU, \tilde p, r)$, where: (i) $T$ is the decision horizon; (ii) $\gamma \in (0,1]$ is the discount factor; (iii) $\sX$ and $\sU$ are the state/action spaces; (iv) $\tilde p(\bv{x}_k\mid\bv{x}_{k-1},\bv{u}_k)$ is the probability of reaching $\bv{x}_k$ from $\bv{x}_{k-1}$ after taking action $\bv{u}_k$; (v) $r$ is the expected reward.   These elements are directly mapped onto \AlgoDM~policy computation problem. Namely, in \AlgoDM: (i) $T$ is the decision horizon ($N$ in Fig.  2a); (ii) $\gamma = 1$; (iii) $\sX$ and $\sU$ are again the state and action spaces (see also at the beginning of Sec. \ref{sec:resolution_engine}); (iv) $\tilde p(\bv{x}_k\mid\bv{x}_{k-1},\bv{u}_k)$ is $\plant{k}{k-1}$; (v) the reward is the negative of the cost.  Following standard distributionally robust MDP frameworks\cite{ZL_etal:22},  an adversarial picks -- within an ambiguity set -- the worst-case transition model that minimizes the cumulative reward.   In \AlgoDM~it maximizes the cumulative cost in the ambiguity set. That is:
\begin{equation*}
    \begin{aligned}
&  \underset{\left\{\plant{k}{k-1}\right\}_{1:N}}{\max} & {\DKL{p_{0:N}}{q_{0:N}}} + {\E_{p_{0:N}}\left[\sum_{k=1}^N \statecostexpectation{k} + \actioncostexpectation{k} \right]} \\
&   \ \ \ \ \ \ \ \ \ \  \st  &    {\plant{k}{k-1} \in\ball{k}{k-1}}, \ \ \forall k =1,\ldots,N.
    \end{aligned}
\end{equation*}
This is the inner maximization problem in Fig.  2a with ambiguity set defined in Methods.}
 
\section{Resolution Engine Details}\label{sec:resolution_engine}
We now describe the formal details of the engine in Fig.  2b (proofs in Sec. \ref{sec:proofs}). The state-space is $\sX{\subseteq}\K^n$ and the action space is $\sU{\subseteq}\K^p$. The spaces can be both continuous and discrete {(we let $\K$ be either $\R$ or $\Z$)}. Also, we make the following two assumptions: 
\begin{description}
    \item[{\bf A1}] $\plant{k}{k-1}$ and $\refplant{k}{k-1}$ are bounded and  $\support{\nominalplant{k}{k-1}}\subseteq\support{\refplant{k}{k-1}}$;
    \item[{\bf A2}] the state cost is non-negative upper bounded in $\sX$, the action cost is non-negative lower bounded in $\sU$.\\
\end{description}
Our starting point is the optimization problem from the Methods,  also reported here for convenience:
\begin{equation}\label{eq:robustproblem 2}
 \begin{aligned}
&  & \underset{\shortpolicy{k}{k-1}\in\sD}{\min} \DKL{\shortpolicy{k}{k-1}}{\shortrefpolicy{k}{k-1}} +\E_{\shortpolicy{k}{k-1}}\left[\actioncostexpectation{k}\right] +  \underset{\shortplant{k}{k-1}}{\max} & \E_{\shortpolicy{k}{k-1}}\left[\DKL{\shortplant{k}{k-1}}{\shortrefplant{k}{k-1}}+\E_{\shortplant{k}{k-1}}\left[\costtotexpectation{k}\right]\right]\\
&   &  \st  &    \ \shortplant{k}{k-1} \in\ball{k}{k-1}.
    \end{aligned}
 \end{equation}
 We first show how \eqref{eq:robustproblem 2} is tackled in \AlgoDM.  Then we show that this problem is the right one that should be solved to find the optimal policy in accordance with the robust free energy principle. \\
 
 \noindent{\bf Tackling \ref{eq:robustproblem 2}.} At each $k$, according to \eqref{eq:robustproblem 2}  the optimal policy can be found by first finding the optimal value of the inner maximization problem and then minimizing over the policies. \AlgoDM~relies on finding the optimal value of the inner problem via a convenient scalar reformulation. To obtain the reformulation, we  first show that for the inner problem in \eqref{eq:robustproblem 2} the expectation and maximization can be swapped. Intuitively, this is possible because the feasibility domain is well-behaved (convex and compact, see Sec. \ref{sec:policy_computation}) and the decision variable of the optimization problem does not depend on the pdf over which the expectation is taken. This intuition is formalized with the following result:
\begin{lemma}\label{lem:minimax_expectation_swap}
The optimal solution, $\shortoptimalplant{k}{k-1}\coloneq\optimalplant{k}{k-1}$, for the inner maximization problem in \eqref{eq:robustproblem 2} exists and it holds that:
\begin{align*}
& \E_{\shortpolicy{k}{k-1}}\left[ \DKL{\shortoptimalplant{k}{k-1}}{\shortrefplant{k}{k-1}} +\E_{\shortoptimalplant{k}{k-1}}\left[\costtotexpectation{k}\right] \right]  \\
& =  \max_{ \shortplant{k}{k-1}\in\ball{k}{k-1}}\E_{\shortpolicy{k}{k-1}}\left[\DKL{\shortplant{k}{k-1}}{\shortrefplant{k}{k-1}}+\E_{\shortplant{k}{k-1}}\left[\costtotexpectation{k}\right]\right].
\end{align*}
\end{lemma}
Essentially, the above result establishes that the maximization step can be performed by first solving the problem in green in Fig.  2b and then taking the expectation. Still, this is an infinite dimensional optimization problem that we seek to conveniently recast as a scalar optimization. To do so, we leverage a change of variables technique\cite{ZH_JH:13} using the likelihood ratio (or Radon-Nikodym derivative)
\begin{align}\label{eqn:ratio_main_text}
\ratio{k}{k-1} & \coloneq\frac{\shortplant{k}{k-1}}{\shortnominalplant{k}{k-1}}.
\end{align} 
With this change of variables, and exploiting the fact that the frontier of the feasibility domain is the set of all $\shortplant{k}{k-1}\in\sD$ such  that $\DKL{\shortplant{k}{k-1}}{\shortnominalplant{k}{k-1}} = \radius{k}{k-1}$, we obtain the following:
\begin{lemma}\label{lem:reformulation}
 Consider the problem in green in Fig. \ref{eq:robustproblem 2}. Then:
\begin{align}\label{eq:robustproblem 5}
\underset{\shortplant{k}{k-1}\in\ball{k}{k-1}}{\max} \ \DKL{\shortplant{k}{k-1}}{\shortrefplant{k}{k-1}} +\E_{\shortplant{k}{k-1}}\left[\costtotexpectation{k}\right]
& =\radius{k}{k-1}
-\max_{\alpha\ge0}\ \min_{\ratio{k}{k-1}\in\sR}\sL(\ratio{k}{k-1},\alpha),
\end{align}
 where $\sL(\ratio{k}{k-1},\alpha)$ is given by
\begin{align}\label{eq:Lagrangian}
    \sL(\ratio{k}{k-1},\alpha)=\E_{\shortnominalplant{k}{k-1}} \left[\ratio{k}{k-1}\ln\frac{\shortrefplant{k}{k-1}}{\shortnominalplant{k}{k-1}}-\ratio{k}{k-1}\costtotexpectation{k}+\alpha\left(\ratio{k}{k-1}\ln\ratio{k}{k-1}\right)\right] -\alpha\radius{k}{k-1}.
\end{align}
\end{lemma}
The result establishes that the cost of ambiguity is $\radius{k}{k-1}
-\max_{\alpha\ge0}\ \min_{\ratio{k}{k-1}\in\sR}\sL(\ratio{k}{k-1},\alpha)$. Next, we need to find the optimal value of $-\max_{\alpha\ge0}\ \min_{\ratio{k}{k-1}\in\sR}\sL(\ratio{k}{k-1},\alpha)$. This is $\costuncertainty{k}{k-1}$ in Fig.  2b. To this aim, using \eqref{eq:Lagrangian} we have:
\begin{equation}\label{eq:robustproblem_for_proof}
\begin{aligned}
& {-\max_{\alpha\ge0}\min_{\ratio{k}{k-1}\in\sR}\sL(\ratio{k}{k-1},\alpha)=} \\
& -\max_{\alpha\ge0}\ -\radius{k}{k-1}\alpha + \min_{\ratio{k}{k-1}\in\sR}\E_{\shortnominalplant{k}{k-1}} \left[\ratio{k}{k-1}\ln\frac{\shortrefplant{k}{k-1}}{\shortnominalplant{k}{k-1}}-\ratio{k}{k-1}\costtotexpectation{k}+\alpha\ratio{k}{k-1}\ln\ratio{k}{k-1}\right].
\end{aligned}
\end{equation}
Then, we define
\begin{align} \label{eq:robustproblem 6}
   \Wx{k}{k-1} \coloneq
\underset{\ratio{k}{k-1}\in\sR}{\min}\E_{\shortnominalplant{k}{k-1}} \left[\ratio{k}{k-1}\ln \frac{\shortrefplant{k}{k-1}}{\shortnominalplant{k}{k-1}}-\ratio{k}{k-1}\costtotexpectation{k}+\alpha \ratio{k}{k-1}\ln\ratio{k}{k-1}\right],
\end{align}
so that the problem in \eqref{eq:robustproblem_for_proof} can be written as
\begin{equation}\label{eq:problem_reformulated}
\underset{\alpha\ge 0}{\min} \ \radius{k}{k-1}\alpha - \Wx{k}{k-1}.
\end{equation}
The $\costuncertainty{k}{k-1}$ in Fig.  2b is the optimal value of the above problem. The following theorem states that $\costuncertainty{k}{k-1}$ can be found by solving a scalar and convex optimization problem: 
\begin{theorem}\label{thm:optimal_value_max}
For each $\bv{x}_{k-1}$ and $\bv{u}_k$, the optimal value of the problem in \eqref{eq:problem_reformulated} is finite and given by:
\begin{equation}\label{eqn:v_optimization_statement}
\costuncertainty{k}{k-1} \coloneq \underset{\alpha \ge 0}{\min} \Vxtilde{k}{k-1},
\end{equation}
where 
\begin{equation}\label{eqn:tildeV}
      {\Vxtilde{k}{k-1}} = \left\{ \begin{aligned}
 &\alpha \ln\E_{\shortnominalplant{k}{k-1}}\left[\left(\frac{\shortnominalplant{k}{k-1}  {\exp{\costtot{k}}}}{\shortrefplant{k}{k-1}}\right)^{\frac{1}{\alpha}}\right]+\alpha{\radius{k}{k-1}},& \alpha> 0 \\ 
&\Mx{k}{k-1}, & \alpha= 0, \\ 
 \end{aligned} \right. 
  \end{equation}
with
\begin{align}\label{eq:M_x}
\Mx{k}{k-1}\coloneq\limsup\limits_{\bv{x}_k\in\support{\shortnominalplant{k}{k-1}}}\ln\left(\frac{\shortnominalplant{k}{k-1}\exp{\costtot{k}}}{\shortrefplant{k}{k-1}}\right) \ge 0.
\end{align}
\end{theorem}
The above results shows that, in order to determine its optimal decision amid environmental ambiguity, the agent does not need to know what is the {worst case} $\plant{k}{k-1}$. Rather, the agent only needs to know what is the cost of ambiguity, i.e., $\costuncertainty{k}{k-1}$ and $\radius{k}{k-1}$. 
In our experiments (see Results) we computed $\costuncertainty{k}{k-1}$, and hence built the cost of ambiguity in Fig.  2b, by solving the optimization problem in \eqref{eqn:v_optimization_statement}.  This problem is the one included in the algorithm deployed on the agents (reported in Section \ref{sec:experiments_details}). Moreover, we did not have to build any specialized software to solve this problem because it is convex and admits a global minimum. We can claim this because of the next result, which also establishes that $\costuncertainty{k}{k-1}$ is bounded. To introduce the result we define explicitly
\begin{equation}\label{eqn:Vx_explicit}
\Vx{k}{k-1} = \alpha \ln\E_{\shortnominalplant{k}{k-1}}\left[\left(\frac{\shortnominalplant{k}{k-1}}{\shortrefplant{k}{k-1}}\right)^{\frac{1}{\alpha}}\exp\left(\frac{\costtot{k}}{\alpha}\right)\right] + \alpha\radius{k}{k-1}.
\end{equation}
\begin{theorem}\label{thm:properties}
For each $\bv{x}_{k-1}$ and $\bv{u}_{k}$, the following statements hold.
\begin{enumerate}
\item $\lim\limits_{\alpha\to\infty}\Vx{k}{k-1}=+\infty$. Also, $\Vx{k}{k-1}>0$ for every $\alpha>0$;
\item $\Vx{k}{k-1}$ is: (i) {{linear and equal to $\ln\bar{c}+\radius{k}{k-1}\alpha$}}, if there exists some constant, say $\bar{c}$, such that $\frac{\shortnominalplant{k}{k-1}\exp{\costtot{k}}}{\shortrefplant{k}{k-1}} = \bar{c}$; (ii) strictly convex on $(0,+\infty)$ otherwise;
\item $\lim\limits_{\alpha\to0} \Vx{k}{k-1}=\Mx{k}{k-1}$, where $\Mx{k}{k-1}$ is defined by \eqref{eq:M_x};
\item $\Vxtilde{k}{k-1}$ has a global minimum on $[0,+\infty)$.
\end{enumerate}
\end{theorem}

In summary, the above results show that the optimal value for the inner maximization problem in \eqref{eq:robustproblem 2} is $\E_{\shortpolicy{k}{k-1}}\left[\radiusexpectation{k}{k-1} + \costuncertaintyexpectation{k}{k-1}\right]$ and $\costuncertainty{k}{k-1}$ can be effectively computed via convex optimization. This yields the problem in Fig.  2b where the variational free energy is minimized across the policies.  The next step is to show that \eqref{eq:robustproblem 2}  is the right problem to solve and, if so: (i) give the optimal solution of the minimization problem; (ii) establish how $\costtot{k}$ is built. This is done next.\\

\noindent{\bf The optimal policy.} The next result shows why, at each $k$,  problem in \eqref{eq:robustproblem 2} needs to be solved, gives the optimal policy and determines what is the lowest free energy that the agent can achieve.

 \begin{corollary}\label{cor:optimal_solution}
The distributionally robust free energy principle yields the optimal policy $\{\shortoptimalpolicy{k}{k-1}\}$, with 
\begin{equation}\label{eqn:optimal_policy_main_problem}
    \shortoptimalpolicy{k}{k-1} = \optimalpolicy{k}{k-1} = \shortoptimalpolicy{k}{k-1} = \frac{\shortrefpolicy{k}{k-1}\exp\left(-\radius{k}{k-1} - \costuncertainty{k}{k-1}{-\actioncost{k}}\right)}{\int \shortrefpolicy{k}{k-1}\exp\left(-\radius{k}{k-1}- \costuncertainty{k}{k-1}{-\actioncost{k}}\right)d\bv{u}_{k}},
    \end{equation}
     where \begin{equation}\label{eqn:costuncertainty_statement}
\begin{aligned}
\costuncertainty{k}{k-1} &= \min\left\{\limsup\limits_{\bv{x}_{k}\in\support{\shortnominalplant{k}{k-1}}}\ln\left(\frac{\shortnominalplant{k}{k-1}\exp\left(\costtot{k}\right)}{\shortrefplant{k}{k-1}}\right),\right.\\
&\left. \underset{\alpha>0}{\min} \ \radius{k}{k-1}\alpha + \alpha \ln\E_{\shortnominalplant{k}{k-1}}\left[\left(\frac{\shortnominalplant{k}{k-1}}{\shortrefplant{k}{k-1}}\right)^{\frac{1}{\alpha}}{\exp\left(\frac{\costtot{k}}{\alpha}\right)}\right]\right\};
\end{aligned}
\end{equation}
and
\begin{equation}\label{eqn:total_cost_statement_1}
\begin{aligned}
\costtot{k} & = \statecost{k} + \costtogo{k+1}{k},
\end{aligned}
\end{equation}
with 
\begin{equation}\label{eqn:total_cost_statement_2}
\begin{aligned}
   \costtogo{N+1}{N} & =0, \\
\costtogo{k}{k-1} & = - \ln \int \shortrefpolicy{k}{k-1}\exp\left(-\radius{k}{k-1} - \costuncertainty{k}{k-1}{-\actioncost{k}}\right)d\bv{u}_k, \ \ k\in 1:N.
\end{aligned}
\end{equation}
Moreover, the  optimal cost, i.e., the smallest free energy that can be achieved,  is bounded and given by
\begin{equation}
     \sum_{k\in 1:N}\E_{p(\bv{x}_{k-1})}\left[\costtogoexpectation{k}{k-1}\right].
\end{equation}
\end{corollary}

\section{Why It Is Better to Be Ambiguity-Free}\label{sec:ambiguity_to_zero}
We give the formal details explaining why an agent interacting with an ambiguous environment cannot outperform an agent that has no environmental ambiguity. First,  we characterize what happens to the problem in the left hand-side of \eqref{eq:inner_problem_4} as the ambiguity set shrinks. Then, we discuss what this implies for the optimal policy.  We recall that the radius of ambiguity is bounded and therefore there is some $\eta<+\infty$ such that $\radius{k}{k-1} \le \eta$. Also, the optimal value of the problem in \eqref{eq:inner_problem_4} is $\costuncertainty{k}{k-1}$, which is given in Theorem \ref{thm:optimal_value_max}, and we use the notation $\costuncertaintyeta{k}{k-1}$ to stress the dependency of $\costuncertainty{k}{k-1}$ on $\eta$.  We also discuss why the optimal policy {down-weights the model over the ambiguity radius} (see main text) when ambiguity increases.

\begin{theorem}\label{thm:small_radius}
Let the set-up of item 2(ii) in Theorem \ref{thm:properties} hold. Then:
\begin{equation}\label{eqn:small_radius}
\lim_{\eta \to 0} \costuncertaintyeta{k}{k-1} = \DKL{\shortnominalplant{k}{k-1}}{\shortrefplant{k}{k-1}} + \E_{\shortnominalplant{k}{k-1}}\left[\costtotexpectation{k}\right].
\end{equation}
\end{theorem}
We characterize what happens to $\costuncertainty{k}{k-1}$ as ambiguity vanishes. Next, we establish that an ambiguity-aware agent making optimal decisions necessarily outperforms an agent affected by ambiguity. In essence, \AlgoDM~forbids that ambiguity is exploited to achieve better performance.

\begin{lemma}\label{lem:worst}
The cost of an agent affected by ambiguity cannot be better than the optimal cost of an ambiguity-free agent. That is, 
\begin{align}\label{eq:worst}
  &-\ln\int_{\sU} \shortrefpolicy{k}{k-1}\exp\left(-\DKL{\shortnominalplant{k}{k-1}}{\shortrefplant{k}{k-1}}-\E_{\shortnominalplant{k}{k-1}}[\costtot{k}]-\actioncost{k}\right)d\bv{u}_k
   \\&
   < -\ln\int_{\sU}\shortrefpolicy{k}{k-1}\exp\left(-\radius{k}{k-1}-\costuncertainty{k}{k-1}-\actioncost{k}\right)d\bv{u}_k
\end{align}
\end{lemma}
We recall from the main text that characterizing the policy when ambiguity increases amounts at studying what happens when $\eta_{\min} = \min\radius{k}{k-1}$ increases.  Note that, from Theorem \ref{thm:optimal_value_max},  $\costuncertainty{k}{k-1}$ is independent on $\radius{k}{k-1}$ when $\eta_{\min}$ is sufficiently large.  In fact,  from the proof of Theorem \ref{thm:properties}(i) the left hand side in the first line of \eqref{eqn:tildeV} is greater than $\alpha\radius{k}{k-1}$.  Therefore, for $\eta_{\min}$ sufficiently large, the exponents in \AlgoDM~policy become $-\radius{k}{k-1}$, thus yielding the results observed in the main text.

\section{Supplementary Details For The Experiments}\label{sec:experiments_details}

The hardware used for the in-silico experiment results was a laptop with an 12th Gen Intel Core i5-12500H 2.50 GHz processor and 18 GB of RAM.  {A comparative table of the time taken,  at each step, to compute the control action for each of the policy computation methods from the main text is reported at the end of the \SI~(Tab. \ref{tab:avg_time_per_action}). } \\

\noindent{\bf \AlgoDM~deployment.} The pseudocode implementing the resolution engine deployed on the agents is given in Algorithm \ref{alg:algoDM}.  We first discuss the setting for the robot experiments. {The {Inputs} line specifies what information, given the current position/state,  \AlgoDM~has available at each $k$.} The algorithm outputs (line $4$) the policy $\optimalpolicy{k}{k-1}$, which we recall being a soft-max. The exponential in the soft-max contains the radius of ambiguity, $\radius{k}{k-1}$, and $\costuncertainty{k}{k-1}$. This latter term is computed in lines $1-3$, which directly follow from the statement of Theorem \ref{thm:optimal_value_max}. In the experiments, the input space is discretized in a $5\times 5$ grid. In our experimental deployment, $\radius{k}{k-1}$ is {defined as described in the Methods ({Experiments settings}); as  highlighted in the Discussion and in the Methods,  in the current implementation, the radius of ambiguity is provided to \AlgoDM~rather than learned.  As also outlined in the Methods,  the $\radius{k}{k-1}$ used in the experiments is set to capture higher agent confidence (i.e., lower ambiguity) as the agent gets closer to its goal destination.  Supplementary Fig.  2 shows that, effectively, $\plant{k}{k-1}$ is always inside the ambiguity set.  A}ll other inputs to the algorithm defined within the Methods.  In Fig.  3{,  the generative model $\refplant{k}{k-1}$ for the experiments  -- a Gaussian centered in the goal position -- is goal-encoding.  Instead,  for the \AlgoDM~experiments reported in Fig.  4, $\refplant{k}{k-1}$ is set to $p_{\max}\left(\bv{x}_{k}\mid\bv{x}_{k-1}\right)$ from the MaxDiff framework as described in Results.} See Code Availability.  In the accompanying code,  we used \textsf{scipy solver} to tackle the optimization and we verified numerically that it would achieve the global minimum.  In line \textsf{1} of  Algorithm \ref{alg:algoDM}, the \textsf{max} was computed by obtaining samples from $\nominalplant{k}{k-1}$ and evaluating the expression on the right-hand side for each sample. The largest value was then selected as maximum. \\

\begin{algorithm}[H]
\SetAlgoLined
\KwResult{Robot policy}
\KwIn{$\refplant{k}{k-1}$, $\refpolicy{k}{k-1}$, $\nominalplant{k}{k-1}$, $\radius{k}{k-1}$, $\statecost{k}$}
\KwOut{$\optimalpolicy{k}{k-1}$}
$\Mx{k}{k-1}\gets \max\limits_{\bv{x}_k\in\support{\nominalplant{k}{k-1}}}\ln\left(\frac{\nominalplant{k}{k-1}\exp{\statecost{k}}}{\refplant{k}{k-1}}\right)$;\\
$      {\Vxtilde{k}{k-1}} \gets \left\{ \begin{aligned}
 &\alpha \ln\E_{\nominalplant{k}{k-1}}\left[\left(\frac{\nominalplant{k}{k-1}  {\exp{\statecost{k}}}}{\shortrefplant{k}{k-1}}\right)^{\frac{1}{\alpha}}\right]+\alpha{\radius{k}{k-1}},& \alpha> 0 \\ 
&\Mx{k}{k-1}, & \alpha= 0; \\ 
 \end{aligned} \right. $\\
$\costuncertainty{k}{k-1} \gets \underset{\alpha \ge 0}{\min} \Vxtilde{k}{k-1}$;\\
$\optimalpolicy{k}{k-1} \gets \frac{\refpolicy{k}{k-1}\exp\left(-\radius{k}{k-1} - \costuncertainty{k}{k-1}\right)}{\int \refpolicy{k}{k-1}\exp\left(-\radius{k}{k-1} - \costuncertainty{k}{k-1}\right)d\bv{u}_{k}}$;
\caption{Pseudocode for \AlgoDM~resolution engine deployed on the robots}
\label{alg:algoDM}
\end{algorithm}

\begin{figure*}[!h]
	\centering
\noindent\includegraphics[width=\columnwidth]{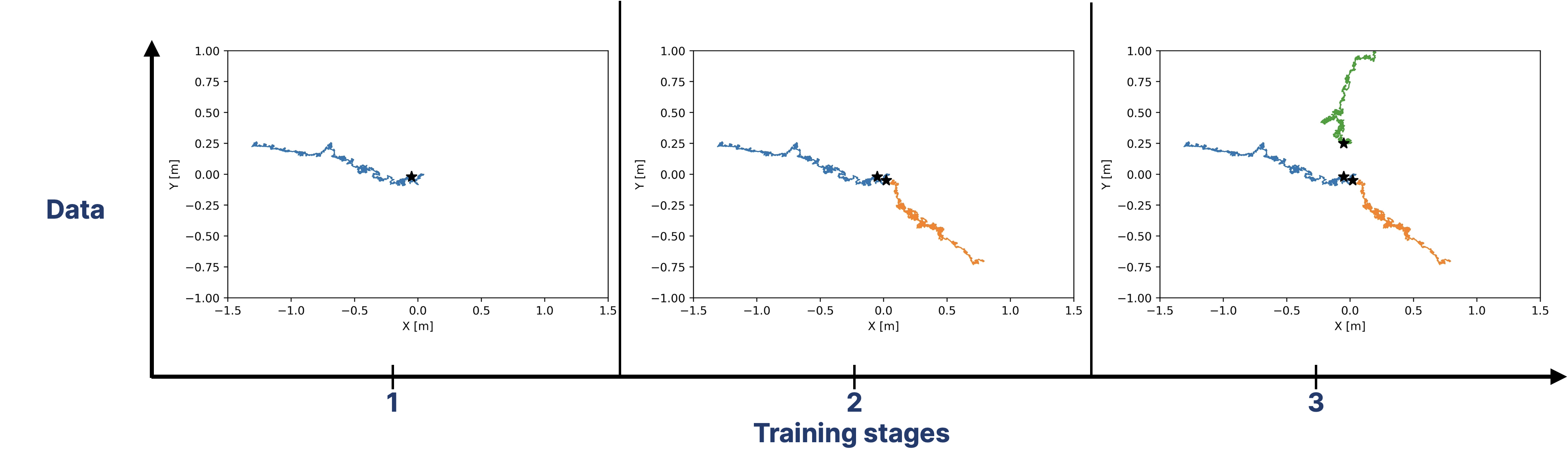}  
\caption{{Position d}ata collected for training at each training stage. The corresponding trained model is detailed in the Methods.} 
 \label{fig:SI-training}
\end{figure*}

Robot trajectories used to obtain the trained model at each phase of training are shown in Fig. \ref{fig:SI-training}. The trajectories were obtained by sampling actions from a uniform distribution and then injecting {in the position} a bias equal to $0.1\bv{x}_{k-1}$.  See Code Availability for the data used for training. \\
 
\noindent{\bf Ambiguity-unaware agent deployment.} The optimal policy from the literature\cite{EG_HJ_CDV_GR:24}  is given by 
$$
\frac{\exp\left(-\DKL{\nominalplant{k}{k-1}}{\refplant{k}{k-1}}-\E_{\nominalplant{k}{k-1}}\left[\statecost{k}\right]\right)}{\sum_{\bv{u}_k}\exp\left(-\DKL{\nominalplant{k}{k-1}}{\refplant{k}{k-1}}-\E_{\nominalplant{k}{k-1}}\left[\statecost{k}\right]\right)}.
$$
In the experiments,  the divergence was considerably higher than the expected cost, which accounted for the presence of obstacles. As a result,  in the policy,  the exponent was approximately equal to $\exp\left(-\DKL{\nominalplant{k}{k-1}}{\refplant{k}{k-1}}\right)$.  Since in the experiments $\refplant{k}{k-1}$ is a Gaussian centered in the goal destination, this policy would direct the robot along the shortest path towards the goal without taking into account the presence of the obstacles and thus explaining the behavior observed in the main text.  In the accompanying code, the policy is computed with a numerical method that prevents underflow due to large divergences.\\

\noindent{\bf Belief update.} The optimization problem solved to obtain the belief update results is
$$
\underset{\bv{v},\bv{w}}{\min}\sum_{k=1}^M\left( - \sum_{i=1}^{F} w_if_{i}\left(\hat{\bv{x}}_{k-1},\hat{\bv{u}}_k\right) + \ln\sum_{\bv{u}_k} \refpolicygiven{k}{k-1}\exp\left(\sum_{i=1}^{F} w_if_{i}\left(\hat{\bv{x}}_{k-1},{\bv{u}}_k\right) + \sum_{i=1}^Gv_{i}g_i\left({\bv{u}}_k\right)\right)\right).
$$
This problem -- derived by dropping the first term from the negative log-likelihood in the Methods -- is convex.  Data points used to reconstruct the cost and code available (see Code Availability).  {The code builds on the one from \cite{EG_HJ_CDV_GR:24}.} \\

\noindent{\bf Increasing ambiguity.} Fig.  3e gives a screenshot of the policy when the ambiguity radius is increased, without clipping, as detailed in the main text. When $\exp(-\radius{k}{k-1})$ leads to underflows, we replace \textsf{0} with \textsf{1e-10}.  The figure was obtained by discretizing the input space in a $50\times 50$ grid.  See Code Availability.\\

\begin{figure*}[!h]
	\centering
\noindent\includegraphics[width=0.75\columnwidth]{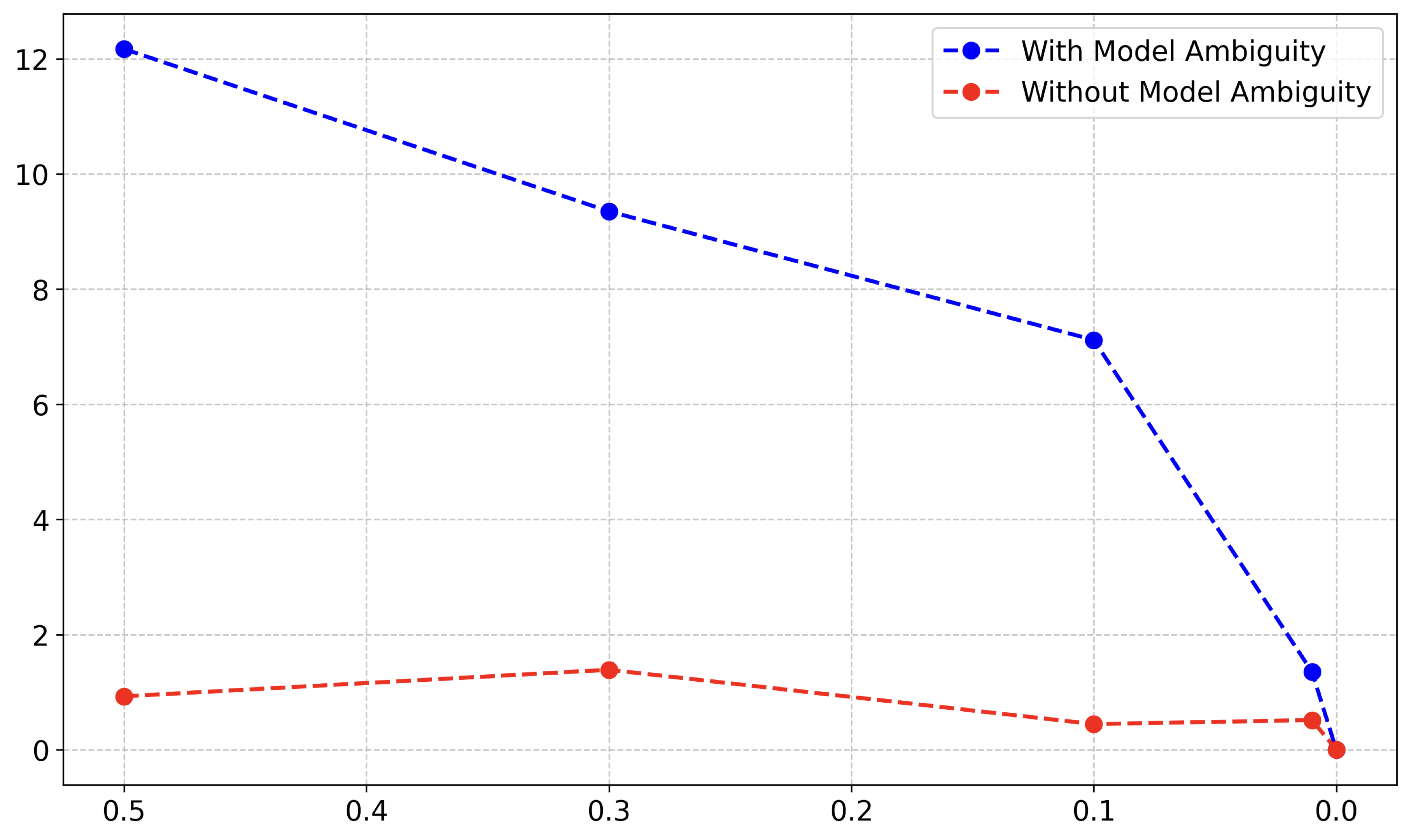}  
\caption{ {Mismatch between \AlgoDM~policy at different ambiguity levels and the optimal policy from the literature\cite{EG_HJ_CDV_GR:24}.  The mismatch is quantified using the KL divergence between the two policies.  Blue curve corresponds to Supplementary Fig.  5 and red curve to Supplementary Fig.  6, respectively.  The x-axis shows the scaling factor for the ambiguity radius and the y-axis the KL divergence. Values are for radii decreased by $50\%$, $70\%$, $90\%$, $99\%$ and for a radius of $0$.  The dashed lines connecting the points are interpolations.  When the radius is zero, the KL divergence is also zero -- the two policies are the same.}} 
 \label{fig:SI-ambiguity}
\end{figure*}

 {\noindent{\bf Decreasing ambiguity.} As in  
Fig.  3e,  Supplementary Fig.  5c and Supplementary Fig.  6b were obtained by discretizing the input space in a $50\times 50$ grid.  As described in the main paper,  when the ambiguity radius is zero (the ambiguity constraint is relaxed) \AlgoDM~policy can be conveniently computed by simply setting $\costuncertainty{k}{k-1}$ as in~\eqref{eqn:small_radius}.   Fig. \ref{fig:SI-ambiguity} confirms that, as the radius decreases,  \AlgoDM~policy approaches the optimal policy for the relaxed problem also  known in the literature\cite{EG_HJ_CDV_GR:24}. } \\

\noindent{{\bf MaxDiff settings.} We use the code provided in the original MaxDiff paper\cite{TB_AP_TM:24},  interfacing it with the Robotarium.  We created a wrapper around the Robotarium environment so that this could be compatible with the MaxDiff available code structure.  In the original code, the MaxDiff authors provide a \textsf{flag} to let MaxDiff use the real environment (reward and model) rather than learning it from data.  In the MaxDiff experiments of the main text and Supplementary Fig.   {7} (panel a) MaxDiff has access to the reward and model and this is achieved by properly setting the \textsf{flag}. For the experiments in the Supplementary Fig.   {7} (panel b) the \textsf{flag} value is changed so that MaxDiff learns reward and model. See Code Availability for the code to replicate the results.  Tab. \ref{tab:MaxDiff} contains the MaxDiff hyperparameters used to obtain the results described in the paper.}\\

\begin{figure*}[!h]
	\centering
\noindent\includegraphics[width=0.7\columnwidth]{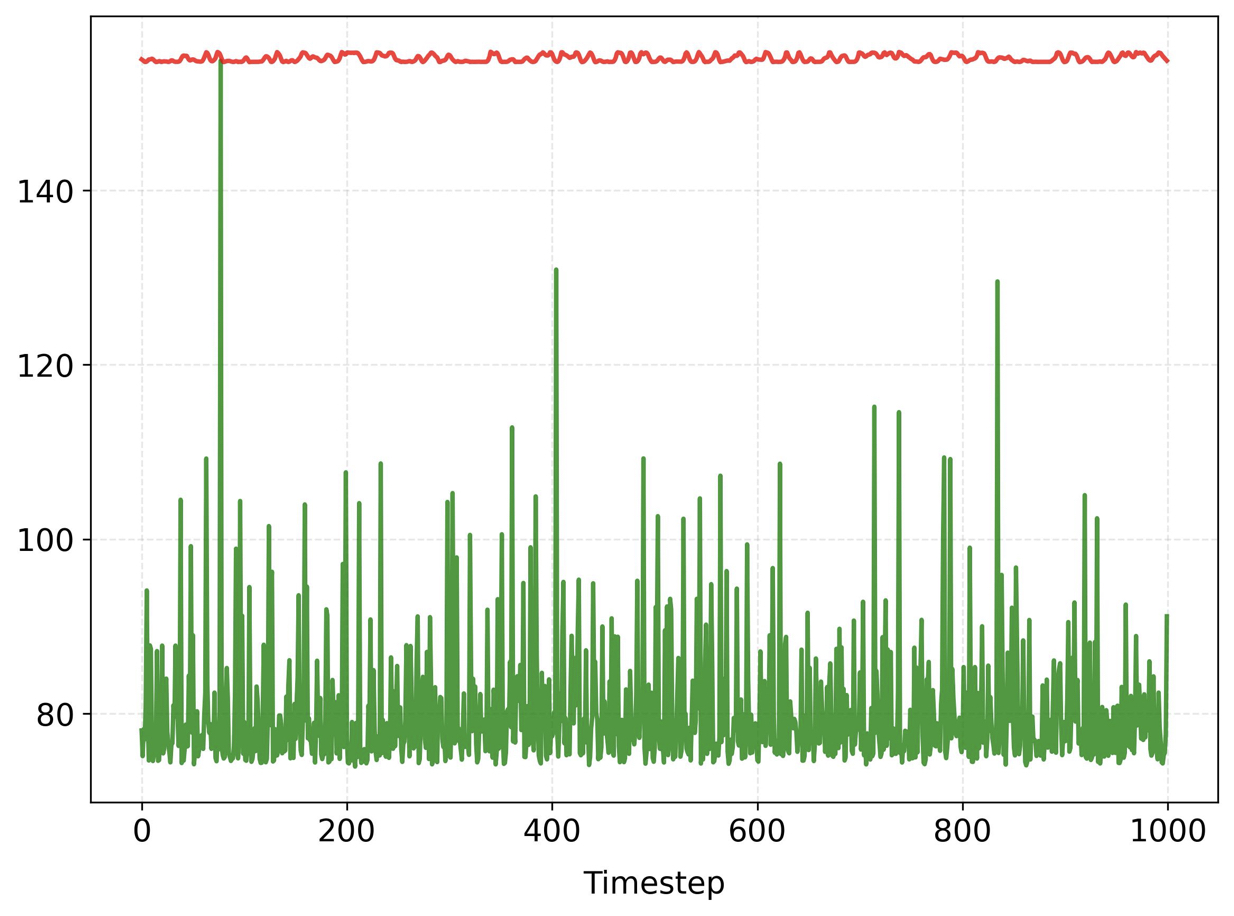}  
\caption{ {$\plant{k}{k-1}$ belongs to the ambiguity set $\ball{k}{k-1}$.  As in Supplementary Fig.  2, the plot confirms this by showing that $\radius{k}{k-1}$ -- in red --  is always bigger than $\DKL{\plant{k}{k-1}}{\nominalplant{k}{k-1}}$ -- in green.  The figure, obtained from one experiment, is representative for all the Ant experiments. }} 
 \label{fig:SI-ambiguity-ant}
\end{figure*}

 {\noindent{{\bf Ant experiments settings.} The task consists in moving the Ant forward, in the x-coordinate direction.   The covariance matrix for the generative model in \AlgoDM~is provided in our repository. For the MaxDiff and NN-MPPI implementations,  we use the code from the literature\cite{TB_AP_TM:24}.   In MaxDiff and NN-MPPI -- both having access to the environment reward -- the horizon is set to $2$ in accordance with the other MaxDiff experiments from the main text.  The other hyperparameters,  summarized in Tab. \ref{tab:Ant}, are taken as in the Ant experiments from the literature\cite{TB_AP_TM:24}.  The table also reports the architecture of the neural network providing mean and variance for $\nominalplant{k}{k-1}$. This is the same architecture used in the MaxDiff paper\cite{TB_AP_TM:24} for the Ant experiments.  The ambiguity radius,  capturing higher confidence as the Ant moves forward,  is the KL divergence between a Gaussian centered at a torso x-velocity of $1$ \textsf{m/s} and a Gaussian whose mean and variance are obtained by the neural network for the torso x-velocity.   Fig.  \ref{fig:SI-ambiguity-ant} shows that  $\plant{k}{k-1}$ is always inside the ambiguity set.   See repository  for details. }}

\section{Proving The  Statements}\label{sec:proofs}
We now give the detailed proofs for the formal statements in the \SI. {Proofs are presented for continuous variables;  the discrete case follows analogous arguments.} First, we introduce some instrumental results.

\subsection{Instrumental Results}

The following result, Bauer’s Maximum Principle, is adapted from  \cite[Corollary A.4.3]{CN_LP:06}. The result is instrumental to tackle concave programs that arise within the proof of our results when maximizing free energy across all possible environments in the ambiguity set 
\begin{lemma}\label{eq:Bauertheorem}
 Let: (i) $\sK$ be a nonempty compact set in a Hausdorff locally convex topological vector space $X$; (ii) $\partial  \sK$ be the frontier of $\sK$; (iii) $f:X\to\R$ be an upper-semi-continuous convex function. Then,  $\max\limits_{\sK} f=\max\limits_{\partial \sK}f$.
\end{lemma}

We also leverage the following Lagrange Duality result 
\cite[Section $8.6$, Theorem $1$]{DL:97}

\begin{theorem} \label{th:minmax}
 Let: (i) $X$ be a linear vector space; (ii) $\Omega$ be a convex
subset of $X$; (iii) $f$ be a real-valued convex 
 functional on $\Omega$; (iv) $G$ be a real convex mapping of $X$. Assume that:
 \begin{enumerate}
 \item   there exists some $\tilde x\in\Omega$ such that $G\left(\tilde x\right)<\eta$;
 \item $\mu_0\coloneq\left\{\min f(x),~{\rm s.t}~~G(x)\le\eta, x\in\Omega\right\}$  is finite, i.e., $\mu_0<\infty$.
\end{enumerate}
 Then:
 \begin{align*}
     \mathop {\min }\limits_{\scriptstyle G(x)\le\eta \hfill \atop 
  \scriptstyle x\in\Omega \hfill} \,f(x)=\max_{\alpha\ge0}\min_{x\in\Omega} [f(x)+\alpha (G(x)-\eta)].
 \end{align*}
\end{theorem}

\subsection{Finding The \AlgoDM~ Policy}\label{sec:policy_computation}

After characterizing the set $\ball{k}{k-1}$, we give a proof for Lemma \ref{lem:minimax_expectation_swap}, which shows that in \eqref{eq:robustproblem 2} the expectation and maximization can be swapped.
\begin{lemma}\label{lem:feasibility_domain}
The set $\ball{k}{k-1}$ in \eqref{eq:robustproblem 2} is convex and compact in $L^1(\sX)$ equipped with the usual norm.
\end{lemma}
\noindent{\bf Proof. }
We first show convexity. Pick any $\tilde{p}_1,\tilde{p}_2\in\ball{k}{k-1}$ and a constant, say $t\in[0,1]$.  Then, by convexity of the KL divergence we have:
$$
\DKL{t\tilde{p}_1+(1-t)\tilde{p}_2}{\shortnominalplant{k}{k-1}}\le t \DKL{\tilde{p}_1}{\shortnominalplant{k}{k-1}} + (1-t)\DKL{\tilde{p}_2}{\shortnominalplant{k}{k-1}} \le \radius{k}{k-1}.
$$
This, together with the fact that $\support \left(t\tilde{p}_1+(1-t)\tilde{p}_2\right) \subseteq\shortrefplant{k}{k-1}$ shows that  $\ball{k}{k-1}$ is convex. 

Now, we prove that the set is also compact and in what follows we let $\eta<+\infty$ be a positive number such that $\radius{k}{k-1}\le\eta$, $\forall \bv{x}_{k-1}, \bv{u}_k$. To this aim, we leverage Lemma \ref{lem:compactness}. By Radon-Nikodym Theorem \cite[Theorem 6.10, page 121]{WR:87} it follows that for each probability measure $\nu\in \mathcal{M}(\sX)$, there exists a unique pdf, say   $p_{\nu}$, such that $\nu(\sE)=\int_Ep_{\nu} d\lambda$ for all measurable set $\sE$ ($\lambda$ is the Lebesgue measure on $\sX$). In turn, by means of Lemma \ref{lem:compactness}, we have that the set $\tilde{\mathcal{B}}_{\eta}=\{\nu\in\mathcal{M}(\sX):~~\DKL{\nu}{\mu}\le\eta\}$ is compact. Let  $\shortnominalplant{k}{k-1}$ be the pdf
corresponding to $\mu$ and consider the functional $\Phi:\mathcal{M}(\sX)\to L^1(\sX)$, with $\Phi(\nu)=p_{k|k-1}^{x,v}$, ($L^1(\sX)$ is equipped with the usual norm). We now prove that $\Phi$ is  continuous. To this aim, pick the measure $\nu\in \mathcal{M}(\sX)$ given by $\nu_n\to\nu$, i.e., for all $\sE$, one has that $\int_{\sE}p_{k|k-1}^{x,v_n}d\lambda\to\int_{\sE}p_{k|k-1}^{x,v}d\lambda$ and hence $\Phi(\nu_n)\to\Phi(\nu)$ in the usual $L^1$ norm. This shows that $\Phi$ is in fact continuous. Moreover, continuous functions map compact sets into compact sets and, additionally, $\Phi(\tilde{\mathcal{B}}_{\eta})=\ball{k}{k-1}$. This yields the conclusion. \qed \\

Next, we prove Lemma \ref{lem:minimax_expectation_swap}, where we establish that in  \eqref{eq:robustproblem 2} the expectation and maximization can be swapped.\\

{\bf Proof of Lemma \ref{lem:minimax_expectation_swap}.}
First, note that (as shown in Lemma \ref{lem:feasibility_domain}) the set $\ball{k}{k-1}$ is convex and compact. This,  together with the continuity and convexity of the KL divergence in the decision variable implies that $\shortoptimalplant{k}{k-1}$ exists.  By definition, we have
\begin{align*}
& \max_{ \shortplant{k}{k-1}\in\ball{k}{k-1}}\DKL{\shortplant{k}{k-1}}{\shortrefplant{k}{k-1}} +\E_{\shortplant{k}{k-1}}\left[\costtotexpectation{k}\right] \\
& = \DKL{\shortoptimalplant{k}{k-1}}{\shortrefplant{k}{k-1}}+\E_{\shortoptimalplant{k}{k-1}}\left[\costtotexpectation{k}\right],
\end{align*}
and hence
\begin{equation}\label{eqn:proof_expectation_swap}
\begin{aligned}
 &       \E_{\shortpolicy{k}{k-1}}\left[ \max_{ \shortplant{k}{k-1}\in\ball{k}{k-1}}\DKL{\shortplant{k}{k-1}}{\shortrefplant{k}{k-1}}+\E_{\shortplant{k}{k-1}}\left[\costtotexpectation{k}\right]\right]  \\
 &= \E_{\shortpolicy{k}{k-1}}\left[\DKL{\shortoptimalplant{k}{k-1}}{\shortrefplant{k}{k-1}}+\E_{\shortoptimalplant{k}{k-1}}\left[\costtotexpectation{k}\right]\right]\\
        & \le\max_{ \shortplant{k}{k-1}\in\ball{k}{k-1}}\E_{\shortpolicy{k}{k-1}}\left[\DKL{\shortplant{k}{k-1}}{\shortrefplant{k}{k-1}}+\E_{\shortplant{k}{k-1}}\left[\costtotexpectation{k}\right]\right].
    \end{aligned}
    \end{equation}
Additionally, for each feasible $\shortplant{k}{k-1}$, 
    $$
    \DKL{\shortplant{k}{k-1}}{\shortrefplant{k}{k-1}}+\E_{\shortplant{k}{k-1}}\left[\costtotexpectation{k}\right]\le \max_{ \shortplant{k}{k-1}\in\ball{k}{k-1}}\DKL{\shortplant{k}{k-1}}{\shortrefplant{k}{k-1}}+\E_{\shortplant{k}{k-1}}\left[\costtotexpectation{k}\right].
    $$
    That is, for each feasible $\shortplant{k}{k-1}$,  
    \begin{align*}
    & \E_{\shortpolicy{k}{k-1}}\left[\DKL{\shortplant{k}{k-1}}{\shortrefplant{k}{k-1}}+\E_{\shortplant{k}{k-1}}\left[\costtotexpectation{k}\right]\right] \\
    & \le \E_{\shortpolicy{k}{k-1}}\left[\max_{ \shortplant{k}{k-1}\in\ball{k}{k-1}}\DKL{\shortplant{k}{k-1}}{\shortrefplant{k}{k-1}}+\E_{\shortplant{k}{k-1}}\left[\costtotexpectation{k}\right]\right].
    \end{align*}
    Hence we may continue the chain of inequalities in \eqref{eqn:proof_expectation_swap} with
\begin{equation}\label{eqn:eqn:proof_expectation_swap_continue}
    \begin{aligned}
    & \max_{ \shortplant{k}{k-1}\in\ball{k}{k-1}}\E_{\shortpolicy{k}{k-1}}\left[\DKL{\shortplant{k}{k-1}}{\shortrefplant{k}{k-1}}+\E_{\shortplant{k}{k-1}}\left[\costtotexpectation{k}\right]\right]\\
        &\le \E_{\shortpolicy{k}{k-1}}\left[\max_{ \shortplant{k}{k-1}\in\ball{k}{k-1}}\DKL{\shortplant{k}{k-1}}{\shortrefplant{k}{k-1}}+\E_{\shortplant{k}{k-1}}\left[\costtotexpectation{k}\right]\right],
    \end{aligned}        
    \end{equation}
Combining \eqref{eqn:proof_expectation_swap} and \eqref{eqn:eqn:proof_expectation_swap_continue} yields the desired conclusion. \qed\\

Next, we obtain the scalar reformulation enabling the computation of the cost of ambiguity in Fig.  2b.  Before proving this result, consider the likelihood ratio in \eqref{eqn:ratio_main_text}. Note that,  a.s., for every feasible $\shortplant{k}{k-1}$,  we have $\ratio{k}{k-1} \ge 0$ and $\E_{\shortnominalplant{k}{k-1}}\left[\ratio{k}{k-1}\right]=1$. This motivates the following:
\begin{definition}\label{def:ratio_domain}
The set of likelihood ratios generated by $\shortplant{k}{k-1}$ is
$\sR\coloneq\{\ratio{k}{k-1} \ : \  \E_{\shortnominalplant{k}{k-1}}\left[\ratio{k}{k-1}\right]=1, \ \ratio{k}{k-1}\ge0, ~{\rm a.s.}\}.$
\end{definition}
We can now prove Lemma \ref{lem:reformulation}.\\

{\bf Proof of Lemma \ref{lem:reformulation}.}
The maximization problem in green in Fig.  2b   can be recast as:
\begin{align}\label{eq:inner_problem_3}
 \begin{aligned}
&  & \underset{\ratio{k}{k-1}\in\sR}{\max} &\E_{\shortnominalplant{k}{k-1}}\left[\ratio{k}{k-1}\left(\ln\ratio{k}{k-1}+\ln\frac{\shortnominalplant{k}{k-1}}{\shortrefplant{k}{k-1}}+\costtotexpectation{k}\right)\right] \\
&   &  \st  &  \E_{\shortnominalplant{k}{k-1}}\left[\ratio{k}{k-1}\ln\ratio{k}{k-1}\right] \le \radius{k}{k-1}.
    \end{aligned} 
\end{align}
Also, from Lemma \ref{eq:Bauertheorem}, by setting:
\begin{itemize}
\item $f(\ratio{k}{k-1})\coloneq \E_{\shortnominalplant{k}{k-1}}\left[\ratio{k}{k-1}\ln \ratio{k}{k-1}+ \ratio{k}{k-1}\ln\frac{\shortnominalplant{k}{k-1}}{\shortrefplant{k}{k-1}}+\ratio{k}{k-1}\costtotexpectation{k}\right]$;
\item $\sK:=\{\ratio{k}{k-1}:~~\ratio{k}{k-1}\in\sR,~~\E_{\shortnominalplant{k}{k-1}}\left[\ratio{k}{k-1}\ln\ratio{k}{k-1}\right] \le \radius{k}{k-1}\}$, which is compact and convex. Compactness follows from Lemma \ref{lem:feasibility_domain}, while convexity in the decision variable follows from the fact that $\E_{\shortnominalplant{k}{k-1}}\left[\ratio{k}{k-1}\ln \ratio{k}{k-1}\right]$ is convex w.r.t.  $\ratio{k}{k-1}$;
\item $X  \coloneq L^1(\mathcal{X})$, which is a  topological vector space; 
\end{itemize}
we have that the optimal value of the problem in \eqref{eq:inner_problem_3} is given by:
\begin{align*}
    &\left\{ \begin{array}{l}
  \max\limits_{\ratio{k}{k-1}\in\sR}\E_{\shortnominalplant{k}{k-1}}\left[\ratio{k}{k-1}\ln \ratio{k}{k-1}+\ratio{k}{k-1}\ln\frac{\shortnominalplant{k}{k-1}}{\shortrefplant{k}{k-1}}+\ratio{k}{k-1}\costtotexpectation{k}\right]\\ 
 {\rm s.t}~ \E_{\shortnominalplant{k}{k-1}}\left[\ratio{k}{k-1}\ln \ratio{k}{k-1}\right]=\radius{k}{k-1} \\ 
 \end{array} \right.
 \\
 &=\radius{k}{k-1}+\left\{ \begin{array}{l}
  \max\limits_{\ratio{k}{k-1}\in\sR}\E_{\shortnominalplant{k}{k-1}}\left[\ratio{k}{k-1}\ln\frac{\shortnominalplant{k}{k-1}}{\shortrefplant{k}{k-1}}+\ratio{k}{k-1}\costtotexpectation{k}\right]\\ 
 {\rm s.t}~ \E_{\shortnominalplant{k}{k-1}}\left[\ratio{k}{k-1}\ln \ratio{k}{k-1}\right]=\radius{k}{k-1}. \\ 
 \end{array} \right.,
\end{align*}
which, by assumptions A1--A2, has a bounded optimal value. Also, we have:
\begin{align} \label{eq:inner_problem_4}
    &\left\{ \begin{array}{l} \max\limits_{\ratio{k}{k-1}\in\sR}\E_{\shortnominalplant{k}{k-1}}\left[\ratio{k}{k-1}\ln\frac{\shortnominalplant{k}{k-1}}{\shortrefplant{k}{k-1}}+\ratio{k}{k-1}\costtotexpectation{k}\right]\\ 
 {\rm s.t}~ \E_{\shortnominalplant{k}{k-1}}\left[\ratio{k}{k-1}\ln \ratio{k}{k-1}\right]=\radius{k}{k-1} \\ 
 \end{array} \right.=-\left\{ \begin{array}{l} \min\limits_{\ratio{k}{k-1}\in\sR}\E_{\shortnominalplant{k}{k-1}}\left[\ratio{k}{k-1}\ln\frac{\shortrefplant{k}{k-1}}{\shortnominalplant{k}{k-1}}-\ratio{k}{k-1}\costtotexpectation{k}\right]\\ 
 {\rm s.t}~ \E_{\shortnominalplant{k}{k-1}}\left[\ratio{k}{k-1}\ln\ratio{k}{k-1}\right]=\radius{k}{k-1} \\ 
 \end{array} \right..
\end{align}
The cost in the right-hand side of \eqref{eq:inner_problem_4} is linear  in {the decision variable}; thus, its optimal value is the same as
\begin{equation*}
\begin{array}{l} \min\limits_{\ratio{k}{k-1}\in\sR}\E_{\shortnominalplant{k}{k-1}}\left[\ratio{k}{k-1}\ln\frac{\shortrefplant{k}{k-1}}{\shortnominalplant{k}{k-1}}-\ratio{k}{k-1}\costtotexpectation{k}\right]\\ 
 {\rm s.t}~ \E_{\shortnominalplant{k}{k-1}}\left[\ratio{k}{k-1}\ln\ratio{k}{k-1}\right]\le\radius{k}{k-1} .
 \end{array}
\end{equation*}
Thus, we can leverage  Theorem \ref{th:minmax} to obtain a reformulation of the functional convex optimization problem in the right-hand side of \eqref{eq:inner_problem_4} having $\ratio{k}{k-1}$ as decision variable. To this aim, in Theorem  \ref{th:minmax} set: (i) $X\coloneq L^{1}(\sX)$; (ii) $\Omega \coloneq \sK$, with $\sK$ defined above; (iii) $G(\ratio{k}{k-1})=\E_{\shortnominalplant{k}{k-1}}[\ratio{k}{k-1}\ln \ratio{k}{k-1}]-\radius{k}{k-1}$, and $f(\ratio{k}{k-1})=\E_{\shortnominalplant{k}{k-1}}\left[\ratio{k}{k-1}\ln\frac{\shortrefplant{k}{k-1}}{\shortnominalplant{k}{k-1}}-\ratio{k}{k-1}\costtotexpectation{k}\right]$. Further, note that, with these definitions: (i) $\Omega$ is a convex subset of the linear vector space $X$; (ii) $f$ is real-valued and convex w.r.t. $\ratio{k}{k-1}$ on $\Omega$; (iii) $G$ is a real convex mapping of {$\ratio{k}{k-1}$}. The hypotheses of Theorem \ref{th:minmax} are all satisfied and this implies that
\begin{align*}
& \underset{\shortplant{k}{k-1}\in\ball{k}{k-1}}{\max}  \DKL{\shortplant{k}{k-1}}{\shortrefplant{k}{k-1}} +\E_{\shortplant{k}{k-1}}\left[\costtotexpectation{k}\right] 
= {\radius{k}{k-1}} -\max_{\alpha\ge0} \min_{\ratio{k}{k-1}\in\sR}\sL(\ratio{k}{k-1},\alpha),
    \end{align*}
with $\sL(\ratio{k}{k-1},\alpha)$ given in \eqref{eq:Lagrangian}. This yields the desired conclusion. \qed\\

Next, we give the proof for Theorem \ref{thm:optimal_value_max}, which yields the scalar, convex, optimization problem to compute $\costuncertainty{k}{k-1}$.\\

{\bf Proof of Theorem \ref{thm:optimal_value_max}.}
We discuss separately two cases for $\alpha >0$ and $\alpha = 0$.\\
\\
\noindent{\bf Case (i): $\alpha >0$.}
The problem in \eqref{eq:robustproblem 6} is a convex functional optimization problem with cost and constraints that are differentiable in the decision variable. We find the optimal solution by studying the variations of the Lagrangian  associated to the problem \eqref{eq:robustproblem 6}:
\begin{align*}
  \tilde{\sL}(\ratio{k}{k-1},\lambda) &=\E_{\shortnominalplant{k}{k-1}} \left[\ratio{k}{k-1}\ln \frac{\shortrefplant{k}{k-1}}{\shortnominalplant{k}{k-1}}-\ratio{k}{k-1}\costtotexpectation{k}+\alpha\left( \ratio{k}{k-1}\ln \ratio{k}{k-1}\right)\right]+\lambda\left(\E_{\shortnominalplant{k}{k-1}}\left[\ratio{k}{k-1}\right]-1\right)\\&
  =\E_{\shortnominalplant{k}{k-1}} \left[\ratio{k}{k-1}\ln \frac{\shortrefplant{k}{k-1}}{\shortnominalplant{k}{k-1}}-\ratio{k}{k-1}\costtotexpectation{k}+\alpha\left( \ratio{k}{k-1}\ln \ratio{k}{k-1}\right)+\lambda \ratio{k}{k-1}\right]-\lambda.
\end{align*}
In particular, by studying the variations of the Lagrangian, we find a stationary solution of the problem, which, since this is convex, is a global minimum. First, we consider the variation of the Lagrangian with respect to $\ratio{k}{k-1}$. This is given by:
\begin{align}
\delta_{r}\tilde{\sL}(\ratio{k}{k-1},\lambda)=\E_{\shortnominalplant{k}{k-1}}\left[\ln\frac{\shortrefplant{k}{k-1}}{\shortnominalplant{k}{k-1}}-\costtotexpectation{k}+\alpha\left(1+\ln \ratio{k}{k-1}\right)+\lambda\right].
\end{align}
Therefore, any candidate optimal solution, say $\feasibleratio{k}{k-1}$, must satisfy:
   \begin{align}\label{eq:problem2}
       \ln\frac{\shortrefplant{k}{k-1}}{\shortnominalplant{k}{k-1}}-\costtotexpectation{k}+\alpha\left(1+\ln \feasibleratio{k}{k-1}\right)+\lambda=0.
   \end{align}
That is,
\begin{align}\label{eq:optimal}
    \feasibleratio{k}{k-1}=\exp\left(\frac{-\lambda-\alpha}{\alpha}\right) \left(\frac{\shortnominalplant{k}{k-1}}{\shortrefplant{k}{k-1}}\right)^{\frac{1}{\alpha}}\exp\left(\frac{\costtot{k}}{\alpha}\right).
\end{align}
Now, the variation of the Lagrangian w.r.t. $\lambda$ is given by: $\E_{\shortnominalplant{k}{k-1}}\left[\ratio{k}{k-1}\right]-1 = 0$. This condition must hold for every feasible solution and, hence, in particular for $\feasibleratio{k}{k-1}$ given in \eqref{eq:optimal}. By imposing this stationarity condition on $\feasibleratio{k}{k-1}$ we obtain:
\begin{align}\label{eq:optimal lambda}
\exp\left(\frac{-\lambda-\alpha}{\alpha}\right) = \frac{1}{\E_{\shortnominalplant{k}{k-1}}\left[\left(\frac{\shortnominalplant{k}{k-1}}{\shortrefplant{k}{k-1}}\right)^{\frac{1}{\alpha}}\exp\left(\frac{\costtot{k}}{\alpha}\right)\right]}
\end{align}
Hence, from \eqref{eq:optimal lambda} and \eqref{eq:optimal} we get:
\begin{align}\label{eq:mainoptimal}
\optimalratio{k}{k-1}=\frac{\left(\frac{\shortnominalplant{k}{k-1}}{\shortrefplant{k}{k-1}}\right)^{\frac{1}{\alpha}}\exp\left(\frac{\costtot{k}}{\alpha}\right)}{\E_{\shortnominalplant{k}{k-1}}\left[\left(\frac{\shortnominalplant{k}{k-1}}{\shortrefplant{k}{k-1}}\right)^{\frac{1}{\alpha}}\exp\left(\frac{\costtot{k}}{\alpha}\right)\right]},
\end{align}
which is well defined $\forall \alpha >0$. Note that, by assumptions A1--A2, $\optimalratio{k}{k-1}$ is bounded. Thus, one implication is that $\support{\shortplant{k}{k-1}}\subset\support{\shortnominalplant{k}{k-1}}$. In turn, again from assumption A1 this implies that $\support{\shortplant{k}{k-1}}\subset\support{\shortrefplant{k}{k-1}}$ Next, we show that the optimal cost of the problem in \eqref{eq:robustproblem 6} is finite. To this aim, we note that the optimal cost of the problem equals the value of the Lagrangian evaluated at $\optimalratio{k}{k-1}$. In turn, we have:
 \begin{align*}
&        \tilde{\sL}\left( \optimalratio{k}{k-1},\lambda\right)=
        \E_{\shortnominalplant{k}{k-1}} \left[\optimalratio{k}{k-1}\ln \frac{\shortrefplant{k}{k-1}}{\shortnominalplant{k}{k-1}}-\optimalratio{k}{k-1}\costtotexpectation{k}+\alpha\left( \optimalratio{k}{k-1}\ln \optimalratio{k}{k-1}\right)\right]+\lambda\left(\E_{\shortnominalplant{k}{k-1}}\left[\optimalratio{k}{k-1}\right]-1\right)
        \\&     
        = \E_{\shortnominalplant{k}{k-1}}\left[\frac{\left(\frac{\shortnominalplant{k}{k-1}}{\shortrefplant{k}{k-1}}\right)^{\frac{1}{\alpha}}\exp\left(\frac{\costtot{k}}{\alpha}\right)}{\E_{\shortnominalplant{k}{k-1}}\left[\left(\frac{\shortnominalplant{k}{k-1}}{\shortrefplant{k}{k-1}}\right)^{\frac{1}{\alpha}}\exp\left(\frac{\costtot{k}}{\alpha}\right)\right]}\left(\ln\left(\frac{\shortrefplant{k}{k-1}}{\shortnominalplant{k}{k-1}}\frac{\left(\frac{\shortnominalplant{k}{k-1}}{\shortrefplant{k}{k-1}}\right)\exp\left(\costtot{k}\right)}{\E_{\shortnominalplant{k}{k-1}}\left[\left(\frac{\shortnominalplant{k}{k-1}}{\shortrefplant{k}{k-1}}\right)^{\frac{1}{\alpha}}\exp\left(\frac{\costtot{k}}{\alpha}\right)\right]^\alpha}\right)-\costtotexpectation{k}\right)\right] \\
        &= \E_{\shortnominalplant{k}{k-1}}\left[\frac{\left(\frac{\shortnominalplant{k}{k-1}}{\shortrefplant{k}{k-1}}\right)^{\frac{1}{\alpha}}\exp\left(\frac{\costtot{k}}{\alpha}\right)}{\E_{\shortnominalplant{k}{k-1}}\left[\left(\frac{\shortnominalplant{k}{k-1}}{\shortrefplant{k}{k-1}}\right)^{\frac{1}{\alpha}}\exp\left(\frac{\costtot{k}}{\alpha}\right)\right]} \left(\ln\left(\frac{\exp\left(\costtot{k}\right)}{\E_{\shortnominalplant{k}{k-1}}\left[\left(\frac{\shortnominalplant{k}{k-1}}{\shortrefplant{k}{k-1}}\right)^{\frac{1}{\alpha}}\exp\left(\frac{\costtot{k}}{\alpha}\right)\right]^\alpha}\right)-\costtotexpectation{k}\right)\right]\\
        & = -\alpha \ln\E_{\shortnominalplant{k}{k-1}}\left[\left(\frac{\shortnominalplant{k}{k-1}}{\shortrefplant{k}{k-1}}\right)^{\frac{1}{\alpha}}\exp\left(\frac{\costtot{k}}{\alpha}\right)\right],
    \end{align*}
which is finite by assumptions A1--A2. By using this last expression into \eqref{eq:problem_reformulated} yields the desired conclusion.\\

\noindent{\bf Case (ii): $\alpha =0$.} We note that  $\Mx{k}{k-1}$ is bounded (assumptions A1--A2) and, when $\alpha =0$, $\radius{k}{k-1}\alpha - \Wx{k}{k-1}$ becomes
\begin{align}\label{eqn:problem_case0}
    \max_{\ratio{k}{k-1}\in\sR}\E_{\shortnominalplant{k}{k-1}}\left[\ratio{k}{k-1}\ln\frac{\shortnominalplant{k}{k-1}}{\shortrefplant{k}{k-1}}{+\ratio{k}{k-1}\costtotexpectation{k}}\right] 
    & = \max_{\ratio{k}{k-1}\in\sR}\E_{\shortnominalplant{k}{k-1}}\left[\ratio{k}{k-1}\left(\ln\frac{\shortnominalplant{k}{k-1}}{\shortrefplant{k}{k-1}}{+\costtotexpectation{k}}\right)\right],
\end{align}
which we want to prove being equal to $\Mx{k}{k-1}$. To this aim, note that, by definition of limit superior,  there exists a sequence $\bv{x}_k^{(n)}\in\support{\shortnominalplant{k}{k-1}}$ such that (using the extended notation)
\begin{align}\label{eq:case0}
    \ln\frac{\nominalplantsequence{k}{k-1}}{\refplantsequence{k}{k-1}} {\exp\left(\costtotsequence{k}\right)}\to \Mx{k}{k-1}, \ \  \mbox{and} \ \ \plantsequence{k}{k-1}\to p, \ \ \mbox{as} \ \  n \to\infty,
\end{align}
Therefore, for the cost in \eqref{eqn:problem_case0}, we have that, in the feasibility domain of the problem:
\begin{align*}
&   \E_{\shortnominalplant{k}{k-1}}\left[\ratio{k}{k-1}\left(\ln\frac{\shortnominalplant{k}{k-1}}{\shortrefplant{k}{k-1}}{+\costtot{k}}\right)\right] \le \E_{\shortnominalplant{k}{k-1}} \left[\ratio{k}{k-1} \limsup \ln\left[\left(\frac{\shortnominalplant{k}{k-1}\exp(\costtot{k})}{\shortrefplant{k}{k-1}}\right)\right]\right]=\Mx{k}{k-1}.
\end{align*}
Next, we show that this is indeed the optimal value of the problem. Indeed, by contradiction, assume that there exists some $\tilde{M}$ such that
$$
\E_{\shortnominalplant{k}{k-1}}\left[\ratio{k}{k-1}\left(\ln\frac{\shortnominalplant{k}{k-1}}{\shortrefplant{k}{k-1}}{+\costtot{k}}\right)\right]  < \tilde{M} < \Mx{k}{k-1}.
$$
Then, by taking the limit superior of the above expression we would have $\Mx{k}{k-1} \le \tilde{M} < \Mx{k}{k-1}$, which is a contradiction. Finally, we prove that $\Mx{k}{k-1}$ is non-negative. Indeed, by contradiction, assume that $\Mx{k}{k-1}<0$. In turn, this means that there exists some $\beta<0$ such that 
\[\limsup\ln \left(\frac{\shortnominalplant{k}{k-1}\exp{\costtot{k}}}{\shortrefplant{k}{k-1}}\right)<\beta<0,\] so that, in particular \[\ln\frac{\shortnominalplant{k}{k-1}\exp{\costtot{k}}}{\shortrefplant{k}{k-1}}<\beta.\] In turn, this implies that  $\shortnominalplant{k}{k-1}\exp\costtot{k}<\shortrefplant{k}{k-1}\exp\beta$. By taking integration we get  
\[1=\int\shortnominalplant{k}{k-1}d\bv{x}_k\le \int \shortnominalplant{k}{k-1}\exp\costtot{k}d\bv{x}_k\le \exp\beta,\] 
where the first inequality follows from the fact that the cost is non-negative (following from assumption A2). The above chain of inequalities yields a contraction as we assumed that $\beta<0$. \qed\\

Finally, we prove Corollary \ref{cor:optimal_solution}, which ultimately states that the policy in Fig.  2b is in fact optimal and explains how $\costtot{k}$ is built.  {Since Corollary \ref{cor:optimal_solution} tackles the minimization problem in the policy space -- a problem consistent with prior literature\cite{EG_HJ_CDV_GR:24}, the proof leverages its key arguments. } \\

{\bf Proof of Corollary \ref{cor:optimal_solution}.}
Consider the cost in the robust free energy principle (main text). By the chain rule for the KL divergence (Lemma \ref{lem:splitting_property}) we can recast the robust free energy principle as the sum of the following sub-problems, see, e.g.,  \cite{EG_HJ_CDV_GR:24}:
\begin{equation}\label{eqn:main_problem_up_to_N-1}
    \begin{aligned}
& \underset{\left\{ \shortpolicy{k}{k-1}\right\}_{1:N-1}}{\min}  \underset{\left\{\shortplant{k}{k-1}\right\}_{1:N-1}}{\max}  \DKL{p_{0:N-1}}{q_{0:N-1}} + \sum_{k=1}^{N-1}\E_{p(\bv{x}_{k-1})}\left[\E_{\shortjointxu{k}{k-1}}\left[\statecostexpectation{k} + \actioncostexpectation{k}\right]\right] \\
  &   \st  \   \shortpolicy{k}{k-1}\in\sD, \ \ \shortplant{k}{k-1} \in\ball{k}{k-1}, \ \ \forall k\in 1:N-1,
    \end{aligned}
\end{equation}
and 
\begin{equation}\label{eqn:main_problem_N}
    \begin{aligned}
& \underset{\shortpolicy{N}{N-1}}{\min}  \underset{\shortplant{N}{N-1}}{\max}  \E_{p(\bv{x}_{N-1})}\left[\DKL{\jointxu{N}{N-1}}{\refjointxu{N}{N-1}} + \E_{\jointxu{N}{N-1}}\left[\costtotexpectation{N} + \actioncostexpectation{N}\right]\right] \\
  &   \st  \   \shortpolicy{N}{N-1}\in\sD, \ \ \shortplant{N}{N-1} \in\ball{N}{N-1},
    \end{aligned}
\end{equation}
where $\costtot{N} = \statecost{N}$. Now, following similar arguments to the ones used in Lemma \ref{lem:minimax_expectation_swap}, we have that the problem in \eqref{eqn:main_problem_N} can be solved by solving
\begin{equation}\label{eqn:main_problem_N_without_expectation}
    \begin{aligned}
& \underset{\shortpolicy{N}{N-1}}{\min}  \underset{\plant{N}{N-1}}{\max}  \DKL{\jointxu{N}{N-1}}{\refjointxu{N}{N-1}} + \E_{\jointxu{N}{N-1}}\left[\costtotexpectation{N} + \actioncostexpectation{N}\right] \\
  &   \st  \   \shortpolicy{N}{N-1}\in\sD, \ \ \shortplant{N}{N-1} \in\ball{N}{N-1},
    \end{aligned}
\end{equation}
and then by taking the expectation over $p(\bv{x}_{N-1})$. To tackle the problem in \eqref{eqn:main_problem_N_without_expectation} we note that the first term in the cost can be written as
$$
 \E_{\policy{N}{N-1}}\left[\DKL{\plant{N}{N-1}}{\refplant{N}{N-1}}\right] + \DKL{\policy{N}{N-1}}{\refpolicy{N}{N-1}},
$$
and also 
\begin{align*}
& \E_{\shortjointxu{N}{N-1}}\left[\costtotexpectation{N} + \actioncostexpectation{N}\right] = \\
& \E_{\policy{N}{N-1}}\left[\E_{\plant{N}{N-1}}\left[\costtotexpectation{N}\right]\right] + \E_{\policy{N}{N-1}}\left[\actioncostexpectation{N}\right].
\end{align*}
That is, the cost of the problem in \eqref{eqn:main_problem_N_without_expectation} can be written as:
\begin{equation}\label{eqn:cost_main_problem_N_without_expectation}
\E_{\shortpolicy{k}{k-1}}\left[\DKL{\shortplant{N}{N-1}}{\shortrefplant{N}{N-1}}+\E_{\shortplant{N}{N-1}}\left[\costtotexpectation{N}\right]\right] + \DKL{\shortpolicy{N}{N-1}}{\shortrefpolicy{N}{N-1}}+ \E_{\shortpolicy{N}{N-1}}\left[\actioncostexpectation{N}\right].
\end{equation}
This yields a problem of the form of \eqref{eq:robustproblem 2} and hence, by the above results:
\begin{itemize}
    \item the optimal solution of the problem in \eqref{eqn:cost_main_problem_N_without_expectation} is  
    \begin{equation}\label{eqn:optimal_policy_N}
    \shortoptimalpolicy{N}{N-1} = \frac{\shortrefpolicy{N}{N-1}\exp\left(-\radius{N}{N-1} - \costuncertainty{N}{N-1}-\actioncost{N}\right)}{\int \shortrefpolicy{N}{N-1}\exp\left(-\radius{N}{N-1}- \costuncertainty{N}{N-1}-\actioncost{N}\right)d\bv{u}_N},
    \end{equation}
    with
    \begin{align*}
\costuncertainty{N}{N-1} &= \min\left\{\limsup\limits_{\bv{x}_N\in\support{\shortnominalplant{N}{N-1}}}\ln\left(\frac{\shortnominalplant{N}{N-1}\exp\left(\costtot{N}\right)}{\shortrefplant{N}{N-1}}\right),\right.\\
&\left. \underset{\alpha>0}{\min} \ \radius{N}{N-1}\alpha + \alpha \ln\E_{\shortnominalplant{N}{N-1}}\left[\left(\frac{\shortnominalplant{k}{k-1}\exp{\costtot{N}}}{\shortrefplant{k}{k-1}}\right)^{\frac{1}{\alpha}}\right]\right\};
\end{align*}
\item the corresponding optimal cost is
$$
 \costtogo{N}{N-1} = - \ln \int \shortrefpolicy{N}{N-1}\exp\left(-\radius{k}{k-1} - \costuncertainty{N}{N-1}-\actioncost{N}\right)d\bv{u}_N,
 $$
 and therefore the cost for the problem in \eqref{eqn:main_problem_N} is 
 \begin{equation}\label{eqn:optimal_cost_N}
 \E_{p(\bv{x}_{N-1})}\left[\costtogoexpectation{N}{N-1}\right],
 \end{equation}
which is also bounded.
\end{itemize}
The policy and cost in \eqref{eqn:optimal_policy_N} and \eqref{eqn:optimal_cost_N} are, respectively, the optimal solution and cost given in the statement of the result at $k=N$. Therefore, by using \eqref{eqn:optimal_cost_N} and \eqref{eqn:main_problem_up_to_N-1} we have that the robust free energy principle can be written as 
\begin{equation}\label{eqn:main_problem_up_to_N-1_with_future_cost}
    \begin{aligned}
& \underset{\left\{ \shortpolicy{k}{k-1}\right\}_{1:N-1}}{\min}  \underset{\left\{\shortplant{k}{k-1}\right\}_{1:N-1}}{\max}  \DKL{p_{0:N-1}}{q_{0:N-1}} + \sum_{k=1}^{N-1}\E_{p(\bv{x}_{k-1})}\left[\E_{\shortjointxu{k}{k-1}}\left[\statecostexpectation{k} + \actioncostexpectation{k}\right]\right] \\
&  \ \ \ \ \ \ \ \ \ \ \ \ \ \ \ \ \ \ \ \ \ \ \ \ \ +  \E_{p(\bv{x}_{N-1})}\left[\costtogoexpectation{N}{N-1}\right]\\
  &   \st  \   \shortpolicy{k}{k-1}\in\sD, \ \ \shortplant{k}{k-1} \in\ball{k}{k-1}, \ \ \forall k\in 1:N-1.
    \end{aligned}
\end{equation}
On the other hand, we have that
\begin{align*}
    \E_{p(\bv{x}_{N-1})}\left[\costtogoexpectation{N}{N-1}\right] & = \E_{p(\bv{x}_{N-2})}\left[\E_{\shortjointxu{N-1}{N-2}}\left[\costtogoexpectation{N}{N-1}\right]\right],
\end{align*}
so that (using again the chain rule for the KL divergence) the cost in \eqref{eqn:main_problem_up_to_N-1_with_future_cost} can be written as:
\begin{align*}
& \DKL{p_{0:N-2}}{q_{0:N-2}} + \sum_{k=1}^{N-2}\E_{p(\bv{x}_{k-1})}\left[\E_{\shortjointxu{k}{k-1}}\left[\statecostexpectation{k} + \actioncostexpectation{k}\right]\right]  \\
& + \E_{p(\bv{x}_{N-2})}\left[\DKL{\shortjointxu{N-1}{N-2}}{\shortrefjointxu{N-1}{N-2}}+\E_{\shortjointxu{N-1}{N-2}}\left[\costtotexpectation{N-1} +\actioncostexpectation{N-1} \right]\right],
\end{align*}
with 
\begin{equation}\label{eqn:costtot_N-1}
\costtot{N-1}\coloneq \statecost{N-1} + \costtogo{N}{N-1}.
\end{equation}
The problem at $k=N-1$ can again be solved independently on the others and $\costtot{N-1}$ is bounded. Hence, by iterating the above steps one obtains that:
\begin{itemize}
    \item the optimal solution at $k=N-1$ is given by
    \begin{equation}
    \shortoptimalpolicy{N-1}{N-2} = \frac{\shortrefpolicy{N-1}{N-2}\exp\left(-\radius{N-1}{N-2} - \costuncertainty{N-1}{N-2}{-\actioncost{N-1}}\right)}{\int \shortrefpolicy{N-1}{N-2}\exp\left(-\radius{N-1}{N-2}- \costuncertainty{N-1}{N-2}{-\actioncost{N-1}}\right)d\bv{u}_{N-1}},
    \end{equation}
    with
\begin{align*}
\costuncertainty{N-1}{N-2} &= \min\left\{\limsup\limits_{\bv{x}_{N-1}\in\support{\shortnominalplant{N-1}{N-2}}}\ln\left(\frac{\shortnominalplant{N-1}{N-2}\exp\left(\costtot{N-1}\right)}{\shortrefplant{N-1}{N-2}}\right),\right.\\
&\left. \underset{\alpha>0}{\min} \ \radius{N-1}{N-2}\alpha + \alpha \ln\E_{\shortnominalplant{N-1}{N-2}}\left[\left(\frac{\shortnominalplant{k}{k-1}\exp{\costtot{N-1}}}{\shortrefplant{k}{k-1}}\right)^{\frac{1}{\alpha}}\right]\right\}.
\end{align*}
This is the optimal policy given in the statement at $k=N-1$
\item the optimal cost for the problem at $k=N-1$ is bounded and given by 
 \begin{equation}
 \E_{p(\bv{x}_{N-2})}\left[\costtogoexpectation{N-1}{N-2}\right].
 \end{equation}
\end{itemize}
The desired conclusions are then drawn by induction after noticing that, at each $k$, the problem in the robust free energy formulation can always be broken down into two sub-problems, with the problem at the last time-step given by:
\begin{equation}
    \begin{aligned}
& \underset{\shortpolicy{k}{k-1}}{\min}  \underset{\shortplant{k}{k-1}}{\max}  \E_{p(\bv{x}_{k-1})}\left[\DKL{\shortjointxu{k}{k-1}}{\shortrefjointxu{k}{k-1}} + \E_{\shortjointxu{k}{k-1}}\left[\costtotexpectation{k} + \actioncostexpectation{k}\right]\right] \\
  &   \st  \   \shortpolicy{k}{k-1}\in\sD, \ \ \shortplant{k}{k-1} \in\ball{k}{k-1},
    \end{aligned}
\end{equation}
with  $\costtot{k}$ bounded and given, at each $k$, by the recursion in \eqref{eqn:total_cost_statement_1} and \eqref{eqn:total_cost_statement_2}. \qed

\subsection{Computing The Cost Of Ambiguity}\label{sec:additional_properties}
Theorem \ref{thm:properties}  establishes several properties useful to compute $\costuncertainty{k}{k-1}$.  \\

\noindent{\bf Proof.} To prove part (1) it suffices to show that 
\[\ln\E_{\shortnominalplant{k}{k-1}}\left[\left(\frac{\shortnominalplant{k}{k-1}}{\shortrefplant{k}{k-1}}\right)^{\frac{1}{\alpha}}{\exp\left(\frac{\costtot{k}}{\alpha}\right)}\right]\ge0.\]
The proof is then by contradiction. Indeed, if we assumed that
\begin{align*}
    &\ln\E_{\shortnominalplant{k}{k-1}}\left[\left(\frac{\shortnominalplant{k}{k-1}}{\shortrefplant{k}{k-1}}\right)^{\frac{1}{\alpha}}{\exp\left(\frac{\costtot{k}}{\alpha}\right)}\right]<0,
\end{align*}
this would imply
 \begin{align*}   \int_{\sX}\left[\left(\frac{\shortnominalplant{k}{k-1}}{\shortrefplant{k}{k-1}}\right)^{\frac{1}{\alpha}}{\exp\left(\frac{\costtot{k}}{\alpha}\right)}\shortnominalplant{k}{k-1}-\shortnominalplant{k}{k-1}\right]d\bv{x}_k<0.
\end{align*}
In turn, this would mean that 
\begin{align*}
\left[\left(\frac{\shortnominalplant{k}{k-1}}{\shortrefplant{k}{k-1}}\right)^{\frac{1}{\alpha}}{\exp\left(\frac{\costtot{k}}{\alpha}\right)}\shortnominalplant{k}{k-1}-\shortnominalplant{k}{k-1}\right]<0,~~~{\rm a.e~in~\sX}
\end{align*}
and consequently, 
\begin{align}\label{eq:uperbound}
    \left(\frac{\shortnominalplant{k}{k-1}\exp\costtot{k}}{\shortrefplant{k}{k-1}}\right)^{\frac{1}{\alpha}}<1~~~{\rm a.e~in~\sX}
\end{align}
which by taking logarithm implies that   
\begin{align*}
    \ln\frac{\shortnominalplant{k}{k-1}}{\shortrefplant{k}{k-1}}< 0~~~{\rm a.e~in~\sX}.
\end{align*}
Finally, by multiplying both sides of the above inequality by $\shortnominalplant{k}{k-1}$ and taking the integral we would have:
\begin{align*}
    \DKL{\shortnominalplant{k}{k-1}}{\shortrefplant{k}{k-1}} \le 0,
\end{align*}
so that $\shortnominalplant{k}{k-1}=\shortrefplant{k}{k-1}$. However, this contradicts  \eqref{eq:uperbound} as by assumption A2 the state cost is positive.

Next we prove part (2) and we start with the case (ii). Note that, for a given function $f$, we have that if $f(\alpha)$ is strictly convex for $\alpha >0$, then $\alpha f(\frac{1}{\alpha})$ is also strictly convex in the same domain. In fact, pick some $\alpha,\beta>0$ and $t\in[0,1]$. If the function is strictly convex, we have 
\begin{align*}
    \left(t\alpha+(1-t)\beta\right)f\left(\frac{1}{t\alpha+(1-t)\beta}\right)&=(t\alpha+(1-t)\beta)\left[f\left(\frac{t\alpha}{t\alpha+(1-t)\beta}.\frac{1}{\alpha}+\frac{(1-t)\beta}{t\alpha+(1-t)\beta}.\frac{1}{\beta}\right)\right]\\&
    <(t\alpha+(1-t)\beta)\left[\frac{t\alpha}{t\alpha+(1-t)\beta}f\left(\frac{1}{\alpha}\right)+\frac{(1-t)\beta}{t\alpha+(1-t)\beta}f\left(\frac{1}{\beta}\right)\right]\\&
    =t\alpha f\left(\frac{1}{\alpha}\right)+(1-t)\beta f\left(\frac{1}{\beta}\right),
\end{align*}
which shows the desired statement. Let 
$$
y(\alpha):=\int\left(\frac{\shortnominalplant{k}{k-1}\exp\costtot{k}}{\shortrefplant{k}{k-1}}\right)^\alpha\shortnominalplant{k}{k-1}d\bv{x}_k
$$ and
$f(\alpha):=\ln\int\left(\frac{\shortnominalplant{k}{k-1}\exp\costtot{k}}{\shortrefplant{k}{k-1}}\right)^{\alpha}\shortnominalplant{k}{k-1}d\bv{x}_k+\radius{k}{k-1}=\ln y(\alpha)+\radius{k}{k-1}$. Then, $\Vx{k}{k-1}=\alpha f(\frac{1}{\alpha})$ and we prove the result by showing that $f(\alpha)$ is strictly convex. 
Now, the function $f$ is differentiable and its second derivative w.r.t. $\alpha$ is:
\begin{align*}
    \frac{d^2}{d\alpha}f(\alpha)=\frac{y^{''}(\alpha)y(\alpha)-(y^{'}(\alpha))^2}{y^2(\alpha)},
\end{align*}
where we the shorthand notation $y^{'}(\alpha)$ and $y^{''}(\alpha)$ to denote the first and second derivative of $y$ w.r.t. $\alpha$. Now, a direct calculation yields: 
\begin{equation}\label{eq:Cauchy}
\begin{aligned}
   &y^{''}(\alpha)y(\alpha)-(y^{'}(\alpha))^2=\int\left(\frac{\shortnominalplant{k}{k-1}\exp\costtot{k}}{\shortrefplant{k}{k-1}}\right)^{\alpha} \left(\ln\frac{\shortnominalplant{k}{k-1}\exp\costtot{k}}{\shortrefplant{k}{k-1}}\right)^2\shortnominalplant{k}{k-1} d\bv{x}_k\\
   & \cdot\int\left(\frac{\shortnominalplant{k}{k-1}\exp\costtot{k}}{\shortrefplant{k}{k-1}}\right)^\alpha\shortnominalplant{k}{k-1}d\bv{x}_k\\ &
   -\left(\int\left(\frac{\shortnominalplant{k}{k-1}\exp\costtot{k}}{\shortrefplant{k}{k-1}}\right)^{\alpha}\left(\ln\frac{\shortnominalplant{k}{k-1}\exp\costtot{k}}{\shortrefplant{k}{k-1}}\right)\shortnominalplant{k}{k-1}d\bv{x}_k\right)^2.
\end{aligned}
\end{equation}
Moreover, note that:
\begin{align*}
   &\left(\int\left(\frac{\shortnominalplant{k}{k-1}\exp\costtot{k}}{\shortrefplant{k}{k-1}}\right)^{\alpha}\left(\ln\frac{\shortnominalplant{k}{k-1}\exp\costtot{k}}{\shortrefplant{k}{k-1}}\right)\shortnominalplant{k}{k-1}d\bv{x}_k\right)^2\\&
   = \left(\int\left(\frac{\shortnominalplant{k}{k-1}\exp\costtot{k}}{\shortrefplant{k}{k-1}}\right)^{\frac{\alpha}{2}}\left(\ln\frac{\shortnominalplant{k}{k-1}\exp\costtot{k}}{\shortrefplant{k}{k-1}}\right)(\shortnominalplant{k}{k-1})^{\frac{1}{2}}. \left(\frac{\shortnominalplant{k}{k-1}\exp\costtot{k}}{\shortrefplant{k}{k-1}}\right)^{\frac{\alpha}{2}}(\shortnominalplant{k}{k-1})^{\frac{1}{2}}d\bv{x}_k\right)^2\\&
   <\int\left(\left(\frac{\shortnominalplant{k}{k-1}\exp\costtot{k}}{\shortrefplant{k}{k-1}}\right)^{\frac{\alpha}{2}}\left(\ln\frac{\shortnominalplant{k}{k-1}\exp\costtot{k}}{\shortrefplant{k}{k-1}}\right)\left(\shortnominalplant{k}{k-1}\right)^{\frac{1}{2}}\right)^2d\bv{x}_k.\int \left(\left(\frac{\shortnominalplant{k}{k-1}\exp\costtot{k}}{\shortrefplant{k}{k-1}}\right)^{\frac{\alpha}{2}}\left(\shortnominalplant{k}{k-1}\right)^{\frac{1}{2}}
    \right)^2d\bv{x}_k\\&
    = \int\left(\frac{\shortnominalplant{k}{k-1}\exp\costtot{k}}{\shortrefplant{k}{k-1}}\right)^{\alpha} \left(\ln\frac{\shortnominalplant{k}{k-1}\exp\costtot{k}}{\shortrefplant{k}{k-1}}\right)^2\shortnominalplant{k}{k-1} d\bv{x}_k\int\left(\frac{\shortnominalplant{k}{k-1}\exp\costtot{k}}{\shortrefplant{k}{k-1}}\right)^\alpha\shortnominalplant{k}{k-1}d\bv{x}_k,
\end{align*}
where we used Cauchy inequality (which is strict in this case). By combining the above expression with  \eqref{eq:Cauchy} we get that $y^{''}(\alpha)y(\alpha)-(y^{'}(\alpha))^2 >0$ and hence $\frac{d^2}{d\alpha^2}f(\alpha)>0$. That is, $\Vx{k}{k-1}$ is strictly convex. To prove case (i) it suffices to know that, when $\ln\frac{\shortnominalplant{k}{k-1}\exp{\costtot{k}}}{\shortrefplant{k}{k-1}} = \bar{c}$, Cauchy inequality holds with equality.\\

We now prove part (3) and start with noticing that  $\limsup_{\alpha\to0} \Vx{k}{k-1} \le \Mx{k}{k-1}$. In fact, by definition of $\Mx{k}{k-1}$ we get that 
\[\Vx{k}{k-1}\le\ln\left( \int\exp\left(\frac{\Mx{k}{k-1}}{\alpha}\right)\shortnominalplant{k}{k-1}dx_k\right)^{\alpha}+\alpha\radius{k}{k-1},\]
which means that $\limsup_{\alpha\to0}\Vx{k}{k-1}\le \Mx{k}{k-1}$. Next, we  show that $\liminf_{\alpha\to0} \Vx{k}{k-1}\ge \Mx{k}{k-1}$, and from here we will draw the desired conclusions. To this aim, pick any $\varepsilon >0$ and define the set 
$$
\sX_{\varepsilon} \coloneq \left\{\bv{x}_k\in\sX \ : \frac{\shortnominalplant{k}{k-1}\exp\costtot{k}}{\shortrefplant{k}{k-1}}\ge \exp(\Mx{k}{k-1}-\epsilon) \ \right\},
$$
which is non-empty by definition of $\Mx{k}{k-1}$. Then, note that
\begin{align*}
    \Vx{k}{k-1}&=\alpha \ln\E_{\shortnominalplant{k}{k-1}}\left[\left(\frac{\shortnominalplant{k}{k-1}}{\shortrefplant{k}{k-1}}\right)^{\frac{1}{\alpha}}\exp\left(\frac{\costtot{k}}{\alpha}\right)\right]+\radius{k}{k-1}\alpha \\
     &    \ge \ln\left(
     \int_{\sX_{\varepsilon}} \left(\frac{\shortnominalplant{k}{k-1}\exp\costtot{k}}{\shortrefplant{k}{k-1}}\right)^{\frac{1}{\alpha}}\shortnominalplant{k}{k-1}d\bv{x}_k\right)^{\alpha}+\alpha\radius{k}{k-1}\\ 
     &
    \ge \left(\Mx{k}{k-1}-\epsilon\right)+\ln\left(\int_{\sX_{\varepsilon}}\shortnominalplant{k}{k-1}d\bv{x}_k\right)^{\alpha}+\alpha\radius{k}{k-1},
\end{align*}
where to obtain the first inequality we used the fact that 
$$
\int_{\sX} \left(\frac{\shortnominalplant{k}{k-1}\exp\costtot{k}}{\shortrefplant{k}{k-1}}\right)^{\frac{1}{\alpha}}\shortnominalplant{k}{k-1}d\bv{x}_k \ge \int_{\sX_{\varepsilon}} \left(\frac{\shortnominalplant{k}{k-1}\exp\costtot{k}}{\shortrefplant{k}{k-1}}\right)^{\frac{1}{\alpha}}\shortnominalplant{k}{k-1}d\bv{x}_k.
$$
The above chain of inequalities shows that  $$\liminf_{\alpha\to0} \Vx{k}{k-1}\ge \Mx{k}{k-1}-\epsilon.$$ Hence, we have that:
\[\Mx{k}{k-1}-\epsilon\le \liminf_{\alpha\to0}\Vx{k}{k-1}\le\limsup_{\alpha\to0}\Vx{k}{k-1}\le \Mx{k}{k-1}.\] 
Moreover, since $\epsilon$ is arbitrary, by taking the limit $\varepsilon \to 0$ we get $\liminf_{\alpha\to0}\Vx{k}{k-1}=\limsup_{\alpha\to0}\Vx{k}{k-1}=\Mx{k}{k-1}$. In turn, this means that $\lim_{\alpha\to0} \Vx{k}{k-1}=\Mx{k}{k-1}$.

The proof of part (4) is by contradiction. Define the following set 
\begin{align*}
  \sS = \left\{ {\alpha  \ge0 ;\,\,\,{\Vxtilde{k}{k-1}} \le \Mx{k}{k-1}} \right\}.
\end{align*}
We note that $\sS$ is: (i) non-empty (by definition of $\Mx{k}{k-1}$); (ii) closed, due to (right) continuity of the function; (iii) bounded (indeed, if $S$ was unbounded, then it would have been possible to find an unbounded sequence, $\alpha_n\to\infty$, such that $\alpha_n\in S$ and therefore $\Vxtilde{k}{k-1}\le \Mx{k}{k-1}$ and this contradicts part (1) of the statement). Now, since $\Vxtilde{k}{k-1}$ is continuous at $0$ (by part (3)), then it has a  minimum over the compact and bounded set $S$, i.e., there exists $\alpha^\star\in S$ such that $\min_{\alpha\in S}\Vxtilde{k}{k-1}=\tilde V_{\alpha^\star}(\bv{x}_{k-1},\bv{u}_k)$. We want to show that this is indeed a global minimum over the set $[0,\infty)$. To prove this, assume by contradiction that there exists some $\beta >0$ that does not belong to $\sS$ and such that $\tilde V_{\beta}(\bv{x}_{k-1},\bv{u}_k)<\tilde V_{\alpha^\star}(\bv{x}_{k-1},\bv{u}_k)$. In turn, this would imply that $\tilde V_{\beta}(\bv{x}_{k-1},\bv{u}_k)< \tilde V_{\alpha^\star}(\bv{x}_{k-1},\bv{u}_k)\le \Mx{k}{k-1}$. That is, $\beta\in S$. However, this is a contradiction because $\beta$ was assumed to not belong to $\sS$. \qed

\subsection{Determining The Role Of Ambiguity}
Theorem \ref{thm:small_radius} establishes how $\costuncertainty{k}{k-1}$ changes when the radius of ambiguity shrinks. This result is instrumental to determine if an agent affected by {ambiguity} can outperform an {ideal}, ambiguity-free, agent.\\

\noindent{\bf Proof of Theorem \ref{thm:small_radius}.}
We use the formulation given in the problem in \eqref{eq:inner_problem_4} and let  
$$\phi(\theta)\coloneq\ln\E_{\shortnominalplant{k}{k-1}}\left[\left(\frac{\shortnominalplant{k}{k-1}\exp\costtotexpectation{k}}{\shortrefplant{k}{k-1}}\right)^{\theta}\right].$$
In what follows we use the shorthand notation $\phi^{(k)}(\theta)$ to denote the $k$-th order derivative of $\phi(\theta)$.
By definition of the optimal solution of the problem in \eqref{eq:inner_problem_4}, we have the following chain of identities:

\begin{align}
    \eta &\nonumber
    =\E_{\shortnominalplant{k}{k-1}}\left[\frac{\left(\frac{\shortnominalplant{k}{k-1}\exp\costtotexpectation{k}}{\shortrefplant{k}{k-1}}\right)^{\frac{1}{\alpha^{\star}}}}{\E_{\shortnominalplant{k}{k-1}}\left[\left(\frac{\shortnominalplant{k}{k-1}\exp\costtotexpectation{k}}{\shortrefplant{k}{k-1}}\right)^{\frac{1}{\alpha^{\star}}}\right]}\ln\frac{\left(\frac{\shortnominalplant{k}{k-1}\exp\costtotexpectation{k}}{\shortrefplant{k}{k-1}}\right)^{\frac{1}{\alpha^{\star}}}}{\E_{\shortnominalplant{k}{k-1}}\left[\left(\frac{\shortnominalplant{k}{k-1}\exp\costtotexpectation{k}}{\shortrefplant{k}{k-1}}\right)^{\frac{1}{\alpha^{\star}}}\right]}\right]\\&\nonumber
    =\frac{\E_{\shortnominalplant{k}{k-1}}\left[\left(\frac{\shortnominalplant{k}{k-1}\exp\costtotexpectation{k}}{\shortrefplant{k}{k-1}}\right)^{\frac{1}{\alpha^{\star}}}\ln \left(\frac{\shortnominalplant{k}{k-1}\exp\costtotexpectation{k}}{\shortrefplant{k}{k-1}}\right)\right]}{\alpha^{\star}\E_{\shortnominalplant{k}{k-1}}\left[\left(\frac{\shortnominalplant{k}{k-1}\exp\costtotexpectation{k}}{\shortrefplant{k}{k-1}}\right)^{\frac{1}{\alpha^{\star}}}\right]}-\ln\E_{\shortnominalplant{k}{k-1}}\left[\left(\frac{\shortnominalplant{k}{k-1}\exp\costtotexpectation{k}}{\shortrefplant{k}{k-1}}\right)^{\frac{1}{\alpha^{\star}}}\right].
\end{align}    
Hence, by letting $\theta =\frac{1}{\alpha^{\star}}$, this yields
\begin{align}\label{eq:theta star}
   \eta  =\theta\phi^{(1)}(\theta)-\phi(\theta),
\end{align}
Also, for the optimal value of the problem in \eqref{eq:inner_problem_4} we have  
    \begin{align*}
       \costuncertainty{k}{k-1}& =\E_{\shortnominalplant{k}{k-1}}\left[\frac{\left(\frac{\shortnominalplant{k}{k-1}\exp\costtotexpectation{k}}{\shortrefplant{k}{k-1}}\right)^{\frac{1}{\alpha^{\star}}}}{\E_{\shortnominalplant{k}{k-1}}\left[\left(\frac{\shortnominalplant{k}{k-1}\exp\costtotexpectation{k}}{\shortrefplant{k}{k-1}}\right)^{\frac{1}{\alpha^{\star}}}\right]}\ln\left(\frac{\shortnominalplant{k}{k-1}\exp\costtotexpectation{k}}{\shortrefplant{k}{k-1}}\right)\right]
       \\&=\frac{
       \E_{\shortnominalplant{k}{k-1}}\left[\left(\frac{\shortnominalplant{k}{k-1}\exp\costtotexpectation{k}}{\shortrefplant{k}{k-1}}\right)^{\frac{1}{\alpha^{\star}}}\ln\left(\frac{\shortnominalplant{k}{k-1}\exp\costtotexpectation{k}}{\shortrefplant{k}{k-1}}\right)\right]}{\E_{\shortnominalplant{k}{k-1}}\left[\left(\frac{\shortnominalplant{k}{k-1}\exp\costtotexpectation{k}}{\shortrefplant{k}{k-1}}\right)^{\frac{1}{\alpha^{\star}}}\right]}=\phi^{(1)}(\theta).
\end{align*}
Next, we prove the result by: (i) showing that the equation in \eqref{eq:theta star} has a solution, say $\theta^{\star}$; (ii) obtaining an expression for $\costuncertainty{k}{k-1}$ by computing $\phi^{(1)}(\theta^\star)$ and expressing it as a function of $\eta$. \\

\noindent{\bf Finding the root of \eqref{eq:theta star}.} First, we show that a solution exists and is unique. To this sim, since $\phi$ is continuous and differentiable, we Taylor expand the right-hand side in \eqref{eq:theta star} around $\theta = 0$. This yields:
\begin{align*}
    \theta\phi^{(1)}(\theta)-\phi(\theta)&=\sum_{m=0}^{\infty}\frac{1}{m!}\phi^{(m+1)}(0)\theta^{m+1}-\sum_{m=0}^{\infty}\frac{1}{m!}\phi^{(m)}(0)\theta^{m}=\sum_{m=0}^{\infty}\frac{1}{m!}\nu_{m+1}\theta^{m+1}-\sum_{m=0}^{\infty}\frac{1}{m!}\nu_m\theta^{m}\\&= 
    {\sum_{m=1}^{\infty}\left[\frac{1}{(m-1)!}-\frac{1}{m!}\right]\nu_m\theta^{m}
    =\sum_{m=2}^{\infty}\frac{1}{m(m-2)!}\nu_m\theta^m=\frac{1}{2}\nu_2\theta^2+\frac{1}{3}\nu_3\theta^3+\mathcal{O}\left(\theta^4\right)},
\end{align*}
where we used the shorthand notation $\nu_m$ to denote $\phi^{(m)}(0)$. Moreover, following similar arguments as those used to show \eqref{eq:Cauchy}, we have that $\nu_2=\phi^{(2)}(0)>0$. Thus, for small enough $\eta$, there exists a root for \eqref{eq:theta star} and, since {$\phi$ is strictly convex by assumption}, the root $\theta^{\star}$ is unique. Moreover, the function $\theta\phi^{(1)}(\theta)-\phi(\theta)$ is strictly increasing ($\frac{d}{d\theta}[\theta\phi^{(1)}(\theta)-\phi(\theta)]=\theta\phi^{(2)}(\theta)>0$ for $\theta>0$) and hence we can invert \eqref{eq:theta star}. In particular, we have:
\begin{align*}
    \eta&=\frac{1}{2}\nu_2{\theta^\star}^{2}+\frac{1}{3}\nu_3{\theta^\star}^{3}+\frac{1}{8}\nu_4{\theta^\star}^{4}+\mathcal{O}\left({\theta^\star}^{5}\right) = \frac{1}{2}{\theta^\star}^{2}\nu_2\left[1+\frac{2\nu_3}{3\nu_2}{\theta^\star}+\frac{\nu_4}{\nu_2}{\theta^\star}^{2}+\mathcal{O}\left({\theta^\star}^{3}\right)\right],
\end{align*}
so that
\begin{align}\label{eq:aprox1}
\theta^\star=\sqrt{\frac{2\eta}{\nu_2}}\left[1+\left(\frac{2\nu_3}{3\nu_2}\theta^\star+\frac{\nu_4}{\nu_2}{\theta^{\star}}^{2}+\mathcal{O}\left({\theta^\star}^{3}\right)\right)\right]^{-\frac{1}{2}}=\sqrt{\frac{2\eta}{\nu_2}}\left[1-\frac{\nu_3}{3\nu_2}\theta^\star+\mathcal{O}\left({\theta^\star}^{2}\right)\right],
\end{align}
where we used the binomial series expansion for a negative fractional power.  Here, we used the  identity 
$$
\left(1+(a_1x+a_2x^2+\mathcal{O}(x^3))\right)^{-\frac{1}{2}}=1-\frac{1}{2}a_1x+\mathcal{O}(x^2).
$$
Next, disregarding the higher order terms in $\theta^\star$ we have
\begin{align}\label{eq:aprox2}
    \theta^*\approx\sqrt{\frac{2\eta}{\nu_2}},
\end{align}
which, together with  \eqref{eq:aprox1}, yields
\begin{align}
    \theta^\star&=\sqrt{\frac{2\eta}{\nu_2}}\left[1-\frac{\nu_3}{3\nu_2}\theta^\star+\mathcal{O}\left({\theta^{\star}}^2\right)\right]
    =\sqrt{\frac{2}{\nu_2}}\eta^{\frac{1}{2}}-\frac{2\nu_3}{3\nu^2_2}\eta+\mathcal{O}(\eta^{\frac{3}{2}}).
\end{align}
We can now compute $\costuncertainty{k}{k-1}$.\\

\noindent{\bf Computing $\costuncertainty{k}{k-1}$.} By using the above expression, we have, by letting $A(x)\coloneq \ln\frac{\shortnominalplant{k}{k-1}\exp\costtot{k}}{\shortrefplant{k}{k-1}}$: 
\begin{align*}
    \costuncertainty{k}{k-1}=\phi^{(1)}(\theta^*) &=\sum_{k=0}^{\infty}\frac{\nu_{k+1}}{k!}\theta^{*k}=\nu_1+\nu_2\theta^*+\frac{\nu_3}{2}\theta^{*2}+\mathcal{O}(\theta^{*3})\\&=\nu_1+\nu_2\left(\sqrt{\frac{2}{\nu_2}}\eta^{\frac{1}{2}}-\frac{2\nu_3}{3\nu_2^2}\eta+\mathcal{O}(\eta^{\frac{3}{2}})\right)+\frac{\nu_3}{2}\left(\frac{2}{\nu_2}\eta+\mathcal{O}(\eta^{\frac{3}{2}})\right)+\mathcal{O}(\eta^{\frac{3}{2}})
\\&=\nu_1+\sqrt{2\nu_2}\eta^{\frac{1}{2}}+\frac{\nu_3}{3\nu_2}\eta+\mathcal{O}(\eta^{\frac{3}{2}})\\
& = \E_{\shortnominalplant{k}{k-1}}[A(x)] +\mathcal{O}(\eta^{\frac{1}{2}}),
\end{align*}
where we used the definition of $\nu_1$. The above expression gives the desired conclusion, as 
$$\lim_{\eta\to0}\costuncertainty{k}{k-1}=\E_{\shortnominalplant{k}{k-1}}[A(x)]=\DKL{\shortnominalplant{k}{k-1}}{\shortrefplant{k}{k-1}} + \E_{\shortnominalplant{k}{k-1}}\left[\costtotexpectation{k}\right].$$\qed\\

We can now give the proof of Lemma \ref{lem:worst}, which establishes that, for a free energy minimizing agent, ambiguity cannot be exploited to achieve better performance.\\

\noindent{\bf Proof of Lemma \ref{lem:worst}.} By breaking down \eqref{eq:worst}, we need to show that
\begin{align*}
    \DKL{\shortnominalplant{k}{k-1}}{\shortrefplant{k}{k-1}}+\E_{\shortnominalplant{k}{k-1}}[\costtotexpectation{k}]<\radius{k}{k-1}+\costuncertainty{k}{k-1}.
\end{align*}
That is, by exploiting the definition of the KL divergence
\begin{align}
    \E_{\shortnominalplant{k}{k-1}}\left[ {\ln}\frac{\shortnominalplant{k}{k-1}\exp\statecost{k}}{\shortrefplant{k}{k-1}}\right]<\radius{k}{k-1}+\costuncertainty{k}{k-1}.
\end{align}
Assuming on the contrary that the above equation does not hold, we then have $ \radius{k}{k-1}+\costuncertainty{k}{k-1}\le\int_{\sX}\ln\frac{\shortnominalplant{k}{k-1}\exp\statecost{k}}{\shortrefplant{k}{k-1}}\shortnominalplant{k}{k-1}d\bv{x}_k$,
this leads to
\begin{align}\label{eq:costinequality}
  \costuncertainty{k}{k-1}<\int_{\sX}\ln\frac{\shortnominalplant{k}{k-1}\exp\statecost{k}}{\shortrefplant{k}{k-1}}\shortnominalplant{k}{k-1}d\bv{x}_k.  
\end{align}
Next, we will analyze two cases,\\

Case (1): if $\costuncertainty{k}{k-1}=\Mx{k}{k-1}$, by definition of $\Mx{k}{k-1}$,  we get  
\begin{align*}
     \costuncertainty{k}{k-1}<\int_{\sX}\ln\frac{\shortnominalplant{k}{k-1}\exp\statecost{k}}{\shortrefplant{k}{k-1}}\shortnominalplant{k}{k-1}d\bv{x}_k\le\int_{\sX}\Mx{k}{k-1}\shortnominalplant{k}{k-1}d\bv{x}_k=\Mx{k}{k-1}
\end{align*}
which is a contradiction. \\

Case (2): in this case, we have that there exists some $\alpha^\star >0$ such that
\begin{align*}
    \costuncertainty{k}{k-1}=\radius{k}{k-1}\alpha^\star + \alpha^\star \ln\E_{\shortnominalplant{k}{k-1}}\left[\left(\frac{\shortnominalplant{k}{k-1}\exp{\costtot{k}}}{\shortrefplant{k}{k-1}}\right)^{\frac{1}{\alpha^\star}}\right].
\end{align*}
Then, employing Jensen inequality, it follows that:
\begin{align*}
    \costuncertainty{k}{k-1} \ge \radius{k}{k-1}\alpha^\star + \E_{\shortnominalplant{k}{k-1}}\left[\ln\left(\frac{\shortnominalplant{k}{k-1}\exp{\costtot{k}}}{\shortrefplant{k}{k-1}}\right)\right].
\end{align*}
This  however contradicts \eqref{eq:costinequality} since $\alpha^\star >0$ and $\radius{k}{k-1} >0$.\qed

{\section{Concluding Interdisciplinary Remarks And Potential Implications across Psychology, Economics and Neuroscience}\label{sec:ambiguity}
Ambiguity is a key theme in both psychology and economics\cite{Pearson2014,Hsu2005,Zak2004}. The large literature in this area speaks to a key dialectic: on the one hand, psychological studies suggest that people prefer certainty over uncertainty. On the other hand, the opportunity to resolve uncertainty underwrites cognitive flexibility information and novelty-seeking behaviour\cite{Berlyne1950,Corcoran2020,Kiverstein2019,Krebs2009,Schwartenbeck2013,Wittmann2008,Barto2013}.   In active inference, this is often described in terms of intrinsic motivation or epistemic affordance\cite{Corcoran2020,Oudeyer2007,Schwartenbeck2019}. The paradigms used to study ambiguity generally leverage the exploration-exploitation dilemma\cite{Cohen2007, Daw2006, Ishii2002,Schwartenbeck2013, Schwartenbeck2019}: e.g., using the two-step maze task or multiarmed bandit problems.  An empirical evaluation of active inference formulations is available in the literature\cite{Markovic2021}. In behavioural economics, ambiguity refers to uncertain outcomes with uncertain probabilities in contrast with risk that refers to uncertain outcomes with certain probabilities. Similarly, ambiguity in this work refers to uncertainty about the environment (i.e., world) model -- as opposed to the risk entailed by uncertainty about outcomes under a given model.}

{The neural correlates of ambiguity and risk -- as assessed using EEG and multiarmed bandit tasks\cite{Zhang2025} -- appear to segregate; with activity in the frontal, central and parietal regions reflecting ambiguity; with activity in frontal and central brain regions reflecting risk. In economic decision making using fMRI, ambiguous cues elicit activity in the frontal and parietal cortex during outcome anticipation\cite{Bach2009}. The authors {suggest that these regions subserve a general function of contextual analysis} that reflects situational or ambiguity awareness.}

{In this broad context,  \AlgoDM~can potentially offer a formal model of ambiguity awareness that can be leveraged using computational phenotyping\cite{Markovic2022, Schwartenbeck2016}. In principle, it is possible to estimate ambiguity awareness -- e.g., in terms of the ambiguity radius -- by estimating the radius that best explains a given subject’s responses. For example, computational phenotyping of this sort has been used to characterise psychiatric cohorts in terms of greater decision uncertainty during approach-avoidance conflict\cite{Smith2021}.}

\begin{landscape}
\begin{table}[h!]
\centering
{\begin{tabular}{|l|c|c|c|c|c|c|}
\hline
\textbf{Hyperparameter} & \textbf{Figure 4.c} & \textbf{SF 4.a top} & \textbf{SF 4.a bottom} & \textbf{SF 4.b top} & \textbf{SF 4.b bottom} \\
\hline
Model layer   & N/A     & N/A     & N/A     & 128$\times$128 & 128$\times$128 \\
\hline
Discount factor ($\gamma$)     & 0.95    & 0.95    & 0.95    & 0.95           & 0.95           \\
\hline
Learning rate & N/A     & N/A     & N/A     & 0.0005         & 0.0005         \\
\hline
Batch size    & N/A     & N/A     & N/A     & 128            & 128            \\
\hline
$\alpha$      & 0.1     & 0.1     & 0.1     & 0.1            & 0.1            \\
\hline
$\lambda$         & 0.5     & 0.5     & 0.5     & 0.5            & 0.5          \\
\hline
Horizon       & 2       & 20      & 50      & 2              & 20             \\
\hline
Samples       & 50      & 50      & 100     & 50             & 50             \\
\hline
\end{tabular}}
\caption{%
{MaxDiff Hyperparameters across experiments.  SF stands for Supplementary Figure. 
\textbf{Model layer} refers to the architecture of the neural network model used to approximate dynamics and reward.   {The notation $x\times y$ means that there are two layers -- the width of the first layer is of $x$ neurons and the width of the second layer is of $y$ neurons.}
\textbf{Discount} ($\gamma$ in the text) tunes the importance of future rewards in planning. 
\textbf{Learning rate} is used in gradient-based optimization for updating the dynamics/reward model parameters.
\textbf{Batch size} determines the number of training data points used per update step in dynamics/reward model training. 
The parameter \textbf{$\alpha$} in the text is the {\bf temperature-like parameter} in MaxDiff's objective.  The parameter \textbf{$\lambda$} is used to scale the softmax-like weighting of trajectory costs. It controls how sharply or smoothly the algorithm differentiates between high-cost and low-cost trajectories during action selection.
\textbf{Horizon} sets the planning horizon for policy roll-outs. 
\textbf{Samples} indicates the number of trajectory samples used for expectation estimation during policy roll-outs.  We refer to the original MaxDiff code for a more detailed explanation of each of the hyperparameters. The values used for the experiments of this paper are set in accordance with the code from the Maxdiff repository.}}
\label{tab:MaxDiff}
\end{table}

\begin{table}[h!]
\centering

{\begin{tabular}{|l|l|}
\hline
\textbf{Hyperparameter} & \textbf{Value} \\
\hline
Discount factor & 0.99 \\
\hline
Stopping criterion & $10^{-5}$ \\
\hline
Maximum iteration & 2000 \\
\hline
Gradient stopping criterion & 0.001 \\
\hline
Learning rate & $\frac{1}{1+k}$ \\
\hline
\end{tabular}}
\caption{
{
Hyperparameter settings for the belief update benchmark. 
\textbf{Discount factor} determines the trade-off between immediate and long-term returns in soft-value iteration. 
\textbf{Stopping Criterion} specifies the convergence threshold for soft-value iteration. 
\textbf{Maximum iteration} is the maximum number of iterations allowed in the soft-value iteration optimisation loop. 
\textbf{Gradient stopping criterion} is the stopping criterion for gradient descent optimisation. 
\textbf{Learning rate} of the gradient descent decays linearly with rate from $1$ ($k$ is the learning step).  All parameters are defined in our repository.}}
\label{tab:Inverse}
\end{table}
\begin{table}[h]
\centering
{\begin{tabular}{|l|c|}
\hline
\textbf{Method} & \textbf{Average Computation time (sec)} \\
\hline
DR-FREE  & 0.22 (763 steps) \\
Ambiguity Unaware  & 0.04 (556 steps) \\
MaxDiff  & 0.02 (637 steps) \\
\hline
\end{tabular}}
\caption{{average time required to output an action at each $k$ for all the policy computation methods considered in the paper. The measurements were obtained as described in the Methods ({Experiments settings} section).  The recordings were obtained from the Robotarium hardware experiments -- in parentheses the number of steps for each experiment ({Ambiguity Unaware} has a lower number of steps because the robot encountered an obstacle before reaching the destination).  In \AlgoDM~and ambiguity unaware agent $q_{0:N}$ is the one from the first section in the Results -- also, the number of samples used in line $2$ of Algorithm \ref{alg:algoDM} was set to $50$, consistently with the MaxDiff settings.  In fact,  for the MaxDiff agent, the planning horizon was set to $20$, and the number of sampled trajectories was also set to $50$. This was a configuration where MaxDiff would consistently complete the task.  }}
\label{tab:avg_time_per_action}
\end{table}

\begin{table}[h!]
\centering {
{\begin{tabular}{|l|c|c|}
\hline
\textbf{Hyperparameter} & \textbf{MaxDiff} & \textbf{NN-MPPI}  \\
\hline
Model layer   & {$512\times 512 \times 512$}     & {$512\times 512 \times 512$}   \\
\hline
Discount factor ($\gamma$)     & 0.95    & 0.95          \\
\hline
$\alpha$      & 5    & N/A   \\
\hline
$\lambda$         & 0.5     & 0.5       \\
\hline
Samples       & 1000      & 1000  \\
\hline
\end{tabular}}}
\caption{%
 {Relevant MaxDiff and NN-MPPI hyperparameters for the Ant experiments.   Definitions are as in Tab. \ref{tab:MaxDiff} and parameters are as in the literature\cite{TB_AP_TM:24}.  The network,  used in all our Ant experiments, is from the original MaxDiff repository: it outputs mean and variance of $\nominalplant{k}{k-1}$. }
}
\label{tab:Ant}
\end{table}

\end{landscape}

\renewcommand{\refname}{Supplementary References}